\documentclass{article} 
\usepackage[accepted]{icml2024}

\usepackage{amsmath,amssymb,amsfonts}
\usepackage{graphicx}
\usepackage{textcomp}
\def\BibTeX{{\rm B\kern-.05em{\sc i\kern-.025em b}\kern-.08em
    T\kern-.1667em\lower.7ex\hbox{E}\kern-.125emX}}
\usepackage{booktabs}
\usepackage{tabularx}
\usepackage{amsfonts}
\usepackage{pgfplots}
\usepackage{pgfplotstable}
\usepgfplotslibrary{fillbetween}
\usepackage{filecontents}
\usepackage{numprint}
\usepackage{multirow}
\renewcommand{\algorithmicforall}{\textbf{for each}}

\newcommand\Tau{\mathcal{T}}
\usepackage{xr}
\usepackage{xcolor}
\usepackage{caption}
\usepackage{subcaption}
\usepackage{tikz}
\tikzstyle{roccurvestyle}=[
    mark=none,
]

\pgfplotsset{compat=1.17}
\usepackage{etoolbox} 
\usepackage{lipsum} 
\AtBeginEnvironment{tabular}{\footnotesize}

\usepackage{hyperref}
\usepackage{adjustbox}

\usepackage{wrapfig}

\usepackage{bbm}
\usepackage[makeroom]{cancel}
\usepackage{array}

\usepackage{pdflscape}
\usepackage[capitalize]{cleveref}

\usepackage{changepage}

\usepackage{subcaption}
\usepackage{booktabs} 
\usepackage{natbib}

\usepackage{amsmath,amsfonts,bm}

\def\eqref#1{equation~\ref{#1}}

\def\1{\bm{1}}

\DeclareMathAlphabet{\mathsfit}{\encodingdefault}{\sfdefault}{m}{sl}
\SetMathAlphabet{\mathsfit}{bold}{\encodingdefault}{\sfdefault}{bx}{n}

\usepackage{hyperref}
\usepackage{url}

\usepackage{amsmath}
\usepackage{amssymb}
\usepackage{mathtools}
\usepackage{amsthm}

\theoremstyle{plain}
\newtheorem{theorem}{Theorem}[section]

\theoremstyle{definition}
\newtheorem{definition}[theorem]{Definition}

\theoremstyle{remark}

\usepackage[textsize=tiny]{todonotes}

\newcommand{\colorRelative}{blue}
\newcommand{\colorRelativeBis}{green}

\newcommand{\colorRelativeDirect}{\colorRelative!20!\colorRelativeBis}

\newcommand{\colorOnline}{red}

\newcommand{\colorLiRAOffline}{purple}

\newcommand{\colorRelativeOffline}{cyan}

\newcommand{\colorPopulation}{olive}
\newcommand{\colorReference}{purple!50!yellow}

\newcommand{\colorRandom}{gray}

\newcommand{\ROCthickness}{ultra thick}

\newcommand{\styleAugOurs}{solid}
\newcommand{\styleAugOthers}{solid} 

\newcommand{\TPRatFPR}{TPR@FPR}

\newcommand{\hundredth}{0.01\%}
\newcommand{\zero}{0.0\%}

\newcommand{\relative}{RMIA }
\newcommand{\relativens}{RMIA}

\newcommand{\relativebayes}{RMIA-Bayes }
\newcommand{\relativedirect}{RMIA-direct }

\newcommand{\lira}{LiRA }
\newcommand{\lirans}{LiRA}
\newcommand{\population}{Attack-P }
\newcommand{\populationns}{Attack-P}
\newcommand{\reference}{Attack-R }
\newcommand{\referencens}{Attack-R}
\newcommand{\offline}{(Offline) }
\newcommand{\online}{(Online) }
\newcommand{\offlinens}{(Offline)}
\newcommand{\quantile}{Quantile-Reg. }

\newcommand{\liraAuthors}{~\cite{Carlini2022Membership} }
\newcommand{\referenceAuthors}{~\cite{Ye2022Enhanced} }

\newcommand{\random}{Random Guess }

\newcommand{\referenceWithAuthors}{Attack-R, \referenceAuthors}
\newcommand{\populationWithAuthors}{Attack-P, \referenceAuthors}
\newcommand{\liraWithAuthors}{\lirans, \liraAuthors}
\newcommand{\liraOfflineWithAuthors}{\lira \offlinens, \liraAuthors}

\newcommand{\AUC}{AUC}

\newcommand{\ROCscaleFactor}{0.82} 
\newcommand{\MainROCscaleFactor}{0.69}
\newcommand{\AUCscaleFactor}{0.75}
\newcommand{\SensiscaleFactor}{0.8}
\newcommand{\ROCMultiscaleFactor}{0.7}
\newcommand{\AUCMainscaleFactor}{0.85}

\newcommand{\setvalue}[2]{
    \ifdefined #1
        \renewcommand{#1}{#2}
    \else
        \newcommand{#1}{#2}
    \fi
}

\newcommand{\parbf}[1]{\paragraph{\textbf{#1.}}}

\def\compileFigures{0}

\newcounter{figureNumber}

\if\compileFigures1
\usetikzlibrary{external}
\fi

\icmltitlerunning{Low-Cost High-Power Membership Inference Attacks}

\begin{document}

\twocolumn[
\icmltitle{Low-Cost High-Power Membership Inference Attacks}

\begin{icmlauthorlist}
\icmlauthor{Sajjad Zarifzadeh}{nus}
\icmlauthor{Philippe Liu}{nus}
\icmlauthor{Reza Shokri}{nus}
\end{icmlauthorlist}

\icmlaffiliation{nus}{National University of Singapore (NUS), CS Department}

\icmlcorrespondingauthor{Sajjad Zarifzadeh}{s.zarif@nus.edu.sg}
\icmlcorrespondingauthor{Reza Shokri}{reza@comp.nus.edu.sg}

\icmlkeywords{Privacy Auditing, Information Leakage, Membership Inference Attacks, Reference Models}

\vskip 0.3in
]

\printAffiliationsAndNotice{}

\begin{abstract}
Membership inference attacks aim to detect if a particular data point was used in training a model. We design a novel statistical test to perform robust membership inference attacks (\textrm{RMIA}) with low computational overhead. We achieve this by a fine-grained modeling of the null hypothesis in our likelihood ratio tests, and effectively leveraging both reference models and reference population data samples. RMIA has superior test power compared with prior methods, \textit{throughout the TPR-FPR curve} (even at extremely low FPR, as~low~as~$0$). Under computational constraints, where only a limited number of pre-trained reference models (as~few~as~$1$) are available, and also when we vary other elements of the attack (e.g., data distribution), our method performs exceptionally well, unlike prior attacks that approach random guessing. RMIA lays the groundwork for practical yet accurate data privacy risk assessment in machine learning.
\end{abstract}

\section{Introduction}

Membership inference attacks (MIA) are used to quantify the information leakage of machine learning algorithms about their training data~\citep{Shokri2017Membership}. Membership inference attacks originated within the realm of summary statistics on high-dimensional data~\citep{Homer2008Resolving}. In this context, different hypothesis testing methods were designed to optimize the trade-off between test power and its error~\citep{Sankararaman2009Genomic, visscher2009limits, dwork2015robust, Murakonda2021Quantifying}. For deep learning algorithms, these tests evolved from using ML itself to perform MIA~\citep{Shokri2017Membership} to using various approximations of the original statistical tests~\citep{Sablayrolles2019White, Ye2022Enhanced, Carlini2022Membership, Watson2021Importance, Bertran2023Scalable}. Attacks also vary based on the threat models and the computation needed to tailor the attacks to specific data points and models (e.g., global attacks~\citep{Shokri2017Membership, Yeom2018Privacy} versus per-sample tailored attacks~\citep{Ye2022Enhanced, Carlini2022Membership, Sablayrolles2019White, Watson2021Importance}) which all necessitate training a \textit{large} number of reference models. 

Although there have been improvements in the effectiveness of attacks, their \textbf{computation cost} renders them useless for practical privacy auditing. Also, as it is shown in the prior work~\citep{Carlini2022Membership, Ye2022Enhanced} different strong attacks exhibit mutual dominance \textit{depending on the test scenarios}! Under a practical computation budget, \citep{Carlini2022Membership} verges on \textbf{random guessing}, and in the abundance of computation budget,~\citep{Ye2022Enhanced} shows low power at low FPR. Through extensive empirical analysis, we observe further \textbf{performance instabilities} in the prior attacks across different settings, where we investigate the impact of varying the number of reference models, the number of required inference queries, the similarity of reference models to the target model, the distribution shift in target data versus population data, and the performance on out-of-distribution (OOD) data. The limitations of existing MIA tests in these scenarios calls for \textit{robust and efficient} membership inference attacks.

Membership inference attack is a hypothesis testing problem, and attacks are evaluated based on their TPR-FPR trade-off curve. We design a \textbf{novel statistical test} for MIA by enumerating the fine-grained plausible worlds associated with the null hypothesis, in which the \textit{target data point could have been replaced with any random sample from the population}. We perform the attack by composing the likelihood ratio (LR) tests of these cases. Our test enhances the differentiation between member and non-member data points, enabling a more precise estimation of test statistics. The computation we propose to compute the LR test statistics is also extremely efficient (as it requires very few reference models) and remains powerful under uncertainties about the data distribution. \textbf{Our robust attack method RMIA dominates prior work in all test scenarios, and consistently achieves a high TPR \textit{across all} FPR (even as low as~$0$), given \textit{any} computation budget.} Another significant aspect of our framework is that many \textit{prior attacks can be framed as simplifications of ours}, shedding light on the causes of their instability and low performance.

\setvalue{\tmplabel}{roc_1_cifar_10_and_100-}
\setcounter{figureNumber}{0}
{\tikzset{external/figure name/.add={}{\tmplabel}}
\begin{figure}[t]
    \centering
    \if\compileFigures1
        \begin{subfigure}{\linewidth}
        \centering
        \scalebox{\ROCscaleFactor}{
        \begin{tikzpicture}
            \pgfplotstableread[col sep=comma]{data/cifar100_num_ref_comparison_4_true_aug_18_roc.csv}\roc
            \pgfplotstableread[col sep=comma]{data/cifar100_num_ref_comparison_4_true_aug_18_auc.csv}\auc
            \begin{axis} [scale=1.0,width=0.8\linewidth,height=0.8\linewidth,
                ylabel={True Positive Rate},ymin=10e-6,ymax=1,xmin=10e-6,xmax=1,
                xlabel={False Positive Rate},
                xmode=log,ymode=log,
                xtick align = outside,
                ytick align = outside,
                ylabel near ticks,ylabel style={align=center, text width=4cm, font=\small},
                legend pos=south east,
                legend columns=1,
                legend style={nodes={scale=0.7, transform shape}, fill opacity=0.6, draw opacity=1,text opacity=1},
                every axis plot/.append style={\ROCthickness},
                legend cell align={left},
                reverse legend=true,
                ]
                \addplot [\colorRandom,dashed] table[x=random,y=random, col sep=comma] \roc;
                \addlegendentry {\random}

                \addplot [\colorPopulation,no marks] table[x=random,y=population, col sep=comma] \roc;
                \addlegendentry {\protect\NoHyper\population\protect\endNoHyper (\AUC=\pgfplotstablegetelem{0}{population}\of\auc\pgfplotsretval)}

                \addplot [\colorReference,no marks] table[x=random,y={reference 2}, col sep=comma] \roc;
                \addlegendentry {\protect\NoHyper\reference\protect\endNoHyper (\AUC=\pgfplotstablegetelem{0}{{reference 2}}\of\auc\pgfplotsretval)}

                \addplot [\colorOnline,no marks,\styleAugOthers] table[x=random,y={liraoffline 2}, col sep=comma] \roc;
                \addlegendentry {\protect\NoHyper\lira\protect\endNoHyper (\AUC=\pgfplotstablegetelem{0}{liraoffline 2}\of\auc\pgfplotsretval)}

                \addplot [\colorRelative,no marks,\styleAugOurs] table[x=random,y={relativeofflineln 2}, col sep=comma] \roc;
                \addlegendentry {\relative (\AUC=\pgfplotstablegetelem{0}{{relativeofflineln 2}}\of\auc\pgfplotsretval)}


            \end{axis}
        \end{tikzpicture}\hspace*{2em}
        \begin{tikzpicture}
            \pgfplotstableread[col sep=comma]{data/cifar100_num_ref_comparison_4_true_aug_18_roc.csv}\roc
            \pgfplotstableread[col sep=comma]{data/cifar100_num_ref_comparison_4_true_aug_18_auc.csv}\auc
            \begin{axis} [scale=1.0,width=0.4\linewidth,height=0.4\linewidth,
                ylabel={True Positive Rate},ymin=10e-6,ymax=1,xmin=10e-6,xmax=1,
                xlabel={False Positive Rate},axis y line*=left,axis x line*=bottom,
                xtick align = outside,
                ytick align = outside,
                ylabel near ticks,ylabel style={align=center, text width=4cm, font=\small},
                legend pos=south east,
                legend columns=1,
                legend style={nodes={scale=0.7, transform shape}, fill opacity=0.6, draw opacity=1,text opacity=1},
                every axis plot/.append style={\ROCthickness},
                legend cell align={left},
                reverse legend=true,
                ]
                \addplot [\colorRandom,dashed] table[x=random,y=random, col sep=comma] \roc;
                \addlegendentry {\random}

                \addplot [\colorPopulation,no marks] table[x=random,y=population, col sep=comma] \roc;
                \addlegendentry {\protect\NoHyper\population\protect\endNoHyper (\AUC=\pgfplotstablegetelem{0}{population}\of\auc\pgfplotsretval)}

                \addplot [\colorOnline,no marks,\styleAugOthers] table[x=random,y={liraoffline 2}, col sep=comma] \roc;
                \addlegendentry {\protect\NoHyper\lira\protect\endNoHyper (\AUC=\pgfplotstablegetelem{0}{liraoffline 2}\of\auc\pgfplotsretval)}

                \addplot [\colorReference,no marks] table[x=random,y={reference 2}, col sep=comma] \roc;
                \addlegendentry {\protect\NoHyper\reference\protect\endNoHyper (\AUC=\pgfplotstablegetelem{0}{{reference 2}}\of\auc\pgfplotsretval)}

                \addplot [\colorRelative,no marks,\styleAugOurs] table[x=random,y={relativeofflineln 2}, col sep=comma] \roc;
                \addlegendentry {\relative (\AUC=\pgfplotstablegetelem{0}{{relativeofflineln 2}}\of\auc\pgfplotsretval)}

            \end{axis}
        \end{tikzpicture}
        }
        \caption{CIFAR-100}
        \label{fig:roc_1_cifar_100}
    \end{subfigure}
    \else
    \centering
    \scalebox{\ROCscaleFactor}{
    \includegraphics{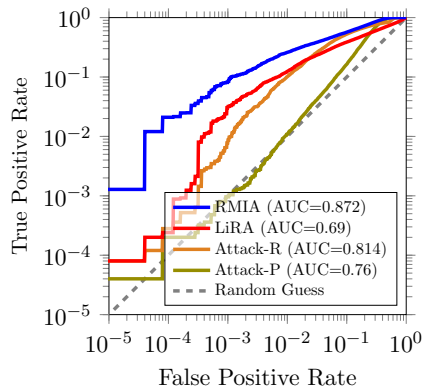}\stepcounter{figureNumber}\stepcounter{figureNumber}
    }
    \fi
    \caption{RMIA versus the prior attacks, Attack-P and Attack-R \citep{Ye2022Enhanced} and also LiRA \citep{Carlini2022Membership}, on CIFAR-100 models, with the restriction of using only \textbf{$1$ reference model} (in an offline setting). RMIA outperforms other attacks throughout the TPR-FPR trade-off curve (e.g. by at least 25\% higher AUC and an order of magnitude better TPR at zero FPR, compared with LiRA). 
    }
    \label{fig:roc_1_cifar_100}
\end{figure}
}

We show that RMIA outperforms prior attacks\footnote{We focus on \citet{Ye2022Enhanced, Carlini2022Membership, Bertran2023Scalable} that represent prior strong MIA methods.} across benchmark datasets\footnote{CIFAR10/100, CINIC10, ImageNet, and Purchase100}, by achieving a significantly higher AUC, i.e., TPR throughout all FPR values, and $2\times$ to $4\times$ higher TPR at low FPRs, when using only~$1$ or $2$ reference models. See Figure~\ref{fig:roc_1_cifar_100}. RMIA's gain is particularly obvious where the adversary exclusively uses pre-trained reference models (i.e., trained independently from the target data). 

We also test how shifting the distribution of training data and population data (by using noisy and OOD data), and modifying model architectures, can impact the attack performance. When used as an oracle in reconstruction attacks~\cite{Carlini2021Extracting}, MIA needs to perform accurately under low FPR regime to filter out the astronomically large number of non-members for discovering members in a high-dimensional space. The vast majority of tested non-members in this application are OOD data. Thus, the advantage of having high TPR at a low FPR primarily comes into play when the attack is evaluated using a large number of non-member (potentially OOD) data for reconstruction attacks. Also, in this setting, \textit{online}\footnote{We do analyze online attacks in this paper, but we can consider them exclusively as proof-of-concept attacks.} MIA methods are useless in practice, as they require training a large number of models per MIA query. Thus, MIA methods need to be both efficient and robust to OOD samples. 

We perform extensive tests to analyze the \textit{robustness} of MIAs, and even considering worst-case scenarios, \textbf{RMIA consistently outperforms other attacks in all settings}.

\newcommand{\LR}{\mathrm{LR}_{\theta}(x, z)}
\newcommand{\MIA}{\mathrm{MIA}(x; \theta)}
\newcommand{\ScoreMIA}{\mathrm{Score}_\mathrm{MIA}(x; \theta)}
\newcommand*\diff{\mathop{}\!\mathrm{d}}

\section{Performing Membership Inference Attacks}
\label{sec:framework}

Membership inference attacks (MIA) determine whether a specific data point~$x$ was used in the training of a given machine learning model~$\theta$. MIA is defined by an indistinguishability game between a challenger and adversary (i.e., privacy auditor). See~\citep{Ye2022Enhanced} for a comprehensive presentation of MIA games. We use the widely-used game and attack template~\citep{Homer2008Resolving, Sankararaman2009Genomic, Shokri2017Membership, Ye2022Enhanced, Carlini2022Membership, Bertran2023Scalable}. The game models random experiments related to two worlds/hypotheses. $\textsc{H}_{in}$: the model~$\theta$ was trained on~$x$, and $\textsc{H}_{out}$: $x$ was not in~$\theta$'s training set (the null hypothesis). The adversary is randomly placed in one of these two worlds and tasked with inferring which world he is in, using only data point~$x$, the trained model~$\theta$, and his background knowledge about the training algorithm and population data distribution.

\begin{definition}[\textbf{Membership Inference Game}]
\label{def:mia_game}
Let~$\pi$ be the data distribution, and let $\mathcal{A}$ be the training algorithm.

\begin{adjustwidth}{0.2cm}{0cm}
\textbf{i} -- The challenger samples a training dataset $S \sim \pi$, and trains a model $\theta \sim \mathcal{A}(S)$.

\textbf{ii} -- The challenger flips a fair coin $b$. If $b = 1$, it randomly samples a data point $x$ from $S$. Otherwise, it samples $x \sim \pi$, such that $x \notin S$. The challenger sends the target model~$\theta$ and the target data point~$x$ to the adversary.

\textbf{iii} -- The adversary, having access to the distribution over the population data $\pi$, computes $\ScoreMIA$ and uses it to output a membership prediction bit ${\hat{b} \leftarrow \MIA}$.
\end{adjustwidth}
\end{definition}

A membership inference attack assigns a membership score $\ScoreMIA$ to every pair of $(x, \theta)$, and performs the hypothesis testing by outputting a membership bit through comparing the score with a threshold~$\beta$:
\begin{equation}
\label{eq:MIA_test}
\MIA = \mathbbm{1}_{\ScoreMIA \ge \beta}
\end{equation}
The adversary's \textbf{power} (true positive rate) and \textbf{error} (false positive rate) are quantified over numerous repetitions of the MIA game experiment. The threshold~$\beta$ controls the false-positive error the adversary is willing to tolerate~\citep{Sankararaman2009Genomic, Murakonda2021Quantifying, Ye2022Enhanced, Bertran2023Scalable}.

The $\ScoreMIA$ and the test \eqref{eq:MIA_test} are designed to maximize the \textbf{MIA test performance} as its power (TPR) for any FPR. The (lower-bound for the) \textit{leakage} of the ML algorithm is defined as the power-error trade-off curve (the ROC curve), which is derived from the outcome of the game experiments across all values of~$\beta$. We primarily compare attacks based on their TPR-FPR curves, but also analyze their computational \textbf{efficiency} and their \textbf{stability} (i.e., how much their power changes when we vary the data distribution and attacker's computational budget).

\section{Designing RMIA}
\label{sec:method}
\label{sec:game}

We propose a novel statistical test for membership inference attacks. We model the null hypothesis (where $x$ is not a member of the training set of $\theta$) as the \textit{composition} of worlds in which the target data point $x$ is replaced by a random data point $z$ sampled from the population. We then compose many \textbf{pairwise likelihood ratio tests} each testing the membership of a data point~$x$ \textit{relative} to another data point~$z$. To reject the null hypothesis, we need to collect substantial evidence (i.e., a large fraction of population data~$z$) that the probability of observing $\theta$ under the hypothesis that $x$ is in its training set is larger than the probability of observing $\theta$ when, instead of $x$, a random $z$ is in the training set. This approach provides a much more fine-grained analysis of leakage, and differentiates between the worlds in which $x$ is not a member (as opposed to relying on the average likelihood of the null hypothesis). We define the likelihood ratio corresponding to the pair of $x$ and $z$ as:
\begin{equation}
\label{eq:likelihood_ratio}
    \LR = \frac{\Pr(\theta | x)}{\Pr(\theta | z)},
\end{equation}
where $\Pr(\theta | . )$ is computed over the randomness of the training algorithm (e.g., SGD). The term $\Pr(\theta | x )$ is the probability that the algorithm produces the model $\theta$ given that $x$ was in the training set, while the rest of the training set is randomly sampled from the population distribution~$\pi$.

\parbf{Computing the Pairwise Likelihood Ratio}
\label{sec:computing_likelihood_ratio}

To efficiently compute the pair-wise LR values in the black-box setting (where the adversary can observe the model output), we apply the Bayes rule to compute~\eqref{eq:likelihood_ratio}:\footnote{$\Pr(\theta)$ is canceled from the numerator and denominator.}
\begin{align}
\label{eq:LR_computation}
    \LR 
    &= \left(\frac{\Pr(x | \theta)}{\Pr(x)}\right)\cdot\left(\frac{\Pr(z | \theta)}{\Pr(z)}\right)^{-1}
\end{align}
Here, $\Pr(x | \theta)$ is the likelihood function of model $\theta$ evaluated on data point $x$. In the case of classification models, and black-box MIA setting, $\Pr(x | \theta)$ is the prediction score (SoftMax) of output of the model $f_\theta(x_{\text{features}})$ for class $x_{\text{label}}$~\citep{mackay2003information, blundell2015weight}.\footnote{See Appendix~\ref{app:calculate_p_x_theta} for alternatives for computing $\Pr(x | \theta)$.}

It is important to note that~$\Pr(x)$ is not the same as~$\pi(x)$, which is rather the prior distribution over~$x$. The term~$\Pr(x)$ is the normalizing constant in the Bayes rule, and is computed by integrating over all models~$\theta'$ with the same structure and training data distribution as~$\theta$.
\begin{align}
\label{eq:px}
    \Pr(x) &= \sum_{\theta'} \Pr(x | \theta') \Pr(\theta') \nonumber \\
    &= \sum_{D, \theta'} \Pr(x | \theta') \Pr(\theta' | D) \Pr(D) 
\end{align}
In practice, we compute $\Pr(x)$ as the empirical mean of $\Pr(x|\theta')$ by sampling \textit{reference models}~$\theta'$, each trained on random datasets $D$ drawn from the population distribution~$\pi$. As we use only a small number of $D, \theta'$ pairs, we need to make sure the reference models are sampled in an \textit{unbiased} way, in particular, with respect to whether $x$ is part of their training data~$D$. Thus, $x$ should be included in the training set of half the reference models (IN models) and be excluded from the training set of the other half (OUT models). This (online attack) is computationally expensive, as customized reference models need to be trained for each MIA query. To avoid this cost, our offline algorithm only computes $\Pr_{OUT}(x)$ by averaging $\Pr(x|\theta')$ over OUT models where $x\not\in D$. To approximate the other half, $\Pr_{IN}(x)$, we scale up $\Pr_{OUT}(x)$, as the inclusion of a data point typically increases its probability. See Appendix~\ref{app:computing_p(x)}, for the details of computing unbiased $\Pr(x)$ from reference models in both online and offline attack settings. The same computation process applies to $\Pr(z)$.

\parbf{Constructing RMIA by Composing Pair-Wise LRs}
\label{sec:mia}

Given $\LR$, we formulate the hypothesis test for our novel membership inference attack RMIA, as follows:
\begin{equation}
\label{eq:MIA_score}
\ScoreMIA = \Pr_{z \sim \pi} \big( \LR \ge \gamma \big)
\end{equation}
We measure the probability that $x$ can $\gamma$-\textit{dominate} a random sample $z$ from the population, for threshold~$\gamma \ge 1$. The threshold~$\gamma \ge 1$ enables us to adjust how much larger the probability of learning~$\theta$ with $x$ as a training data should be \textit{relative} to a random alternative point $z$ to pass the test.\footnote{Figure~\ref{fig:stats_vs_gamma} shows that the MIA test is not very sensitive to small variations of~$\gamma$.} For the simplest setting of $\gamma=1$, the MIA score reflects the quantile corresponding to ${\Pr(x | \theta)}/{\Pr(x)}$ in the distribution of ${\Pr(z | \theta)}/{\Pr(z)}$ over random $z$ samples. 

We reject the null hypothesis if we find enough fraction of $z$ samples for which the probability of $x$ on target model versus its probability over reference models has a larger gap than that of reference population $z$ (which are not in the training set of $\theta$). The following presents our attack procedure (we provide a detailed pseudo-code in Appendix~\ref{app:rmia_pseudocode}).

\begin{definition}[\textbf{Robust Membership Inference Attack}] Let $\theta$ be the target model, and let~$x$ be the target data point. Let~$\gamma$ and $\beta$ be the MIA test parameters. RMIA determines if $x$ was in the training set of $\theta$, by following these steps:
\label{def:attack}
\begin{adjustwidth}{0.2cm}{0cm}
\textbf{i} -- Sample many $z \sim \pi$, and compute $\ScoreMIA$ as the fraction of $z$ samples that pass the pair-wise membership inference likelihood ratio test $\LR \ge \gamma$. See~\eqref{eq:MIA_score}.

\textbf{ii} -- Return \textsc{member} if $\ScoreMIA \ge \beta$, and \textsc{non-member} otherwise. See~\eqref{eq:MIA_test}.
\end{adjustwidth}
\end{definition}


By performing the test over all possible values of $\beta \in [0, 1]$, we can compute the ROC power-error trade-off curve. As Figure~\ref{fig:gamma_beta} shows, our test is calibrated in a sense that when~$\gamma$ is set to $1$, the expected FPR of the attack is $1-\beta$. The ability to adjust the attack to achieve a specific FPR is a significant advantage when conducting practical audits to assess the privacy risk of models. 

\begin{table*}[t!]
    \centering
    \renewcommand{\arraystretch}{1.5}
    \resizebox{\textwidth}{!}{
        \begin{tabular}{|l|l|l|l|l|l|}
        \hline
            \textbf{Method}
            &  RMIA
            & LiRA
            & Attack-R
            & Attack-P
            & Global
            \\
            &  (this paper)
            & \citep{Carlini2022Membership}
            & \citep{Ye2022Enhanced}
            & \citep{Ye2022Enhanced}
            & \citep{Yeom2018Privacy}
            \\ \hline
            \textbf{MIA Score}
            & $\Pr_z \big( \frac{\Pr(\theta | x)}{\Pr(\theta | z)} \ge \gamma \big)$
            & $\frac{\Pr(\theta | x)}{\Pr(\theta | \bar{x})}$
            & $\Pr_{\theta'} \big( \frac{\Pr(x | \theta)}{\Pr(x | \theta')} \ge 1 \big)$
            & $\Pr_z \big( \frac{\Pr(x | \theta)}{\Pr(z | \theta)} \ge 1 \big)$
            &  $\Pr(x | \theta)$\\ \hline
        \end{tabular}
    }
    \caption{Computation of $\ScoreMIA$ in different membership inference attacks, where the notation $\bar{x}$ (for LiRA) represents the case where $x$ is not in the training set. The attack is $\MIA = \mathbbm{1}_{\ScoreMIA \ge \beta}$ based on Definition~\ref{def:mia_game}. }
    \label{tab:mia_scores}
    \renewcommand{\arraystretch}{1}
\end{table*}

\section{Why is RMIA a More Powerful Test Compared with Prior Attacks?}

Membership inference attacks, framed as hypothesis tests, essentially compute the \textit{relative} likelihood of observing $\theta$ given $x$'s membership in the training set of $\theta$ versus observing $\theta$ under $x$'s non-membership (null hypothesis). The key to a powerful test is accounting for \textit{all information sources} that distinguish these possible worlds. Membership inference attacks use \textit{references} from the hypothesis worlds, comparing the pair $(x, \theta)$ against them. Effectively designing the test involves leveraging all possible informative references, which could be either population data or models trained on them. MIA methods predominantly focus on using reference models. The \textit{way} that such reference models are used matters a lot. As we show in our empirical evaluation, prior attacks~\citep{Carlini2022Membership, Ye2022Enhanced} exhibit different behavior depending on the reference models (i.e., in different scenarios, they \textit{dominate each other in opposing ways}).  Also, even though they outperform attacks that are based on population data by a large margin, they do not strictly dominate them on all membership inference queries~\citep{Ye2022Enhanced}. They, thus, fall short due to overlooking some type of distinguishing signals.

Table~\ref{tab:mia_scores} summarizes the MIA scores of various attacks. Our method offers a novel perspective on the problem. This approach leverages both population data and reference models, enhancing attack power and robustness against changes in adversary's background knowledge (e.g. about the distribution of training and population data as well as the structure of the models). Our likelihood ratio test, as defined in~\eqref{eq:MIA_score} and~\eqref{eq:LR_computation}, effectively measures the distinguishability between~$x$ and any $z$ based on the shifts in their probabilities when conditioned on~$\theta$, through contrasting ${\Pr(x | \theta)}/{\Pr(x)}$ versus ${\Pr(z | \theta)}/{\Pr(z)}$. The prior work could be seen as \textit{average-case} and \textit{uncalibrated} versions of our test. \population~\citep{Ye2022Enhanced} and related methods~\citep{Shokri2017Membership, chang2021privacy, Bertran2023Scalable} rely primarily on how the likelihood of the target model on $x$ and $z$ change, and neglect or fail at accurately capturing the ${\Pr(x)}/{\Pr(z)}$ term of our test (i.e., they implicitly assume ${\Pr(x)}={\Pr(z)}$). The strength of our test lies in its ability to detect subtle differences in ${\Pr(x | \theta)}/{\Pr(z | \theta)}$ and ${\Pr(z)}/{\Pr(x)}$ LRs, which reflect the noticeable change in LR due to the inclusion of $x$ in the target training set. In our test, a reference data point $z$ would vote for the membership of $x$ only if the ratio ${\Pr(x)}/{\Pr(z)}$ is enlarged when point probabilities are computed on the target model~$\theta$ (which indicates that $\theta$ fits $x$ better). 

Stronger prior attacks utilizing reference models, especially as seen in~\citep{Ye2022Enhanced} and similar attacks~\citep{Watson2022Difficulty}, neglect the ${\Pr(z | \theta)}/{\Pr(z)}$ component of our test. Calibration by~$z$ would tell us if the magnitude of ${\Pr(x | \theta)}/{\Pr(x)}$ is significant (compared to non-members), without which the attacks would under-perform throughout the TPR-FPR curve. \lira~\citep{Carlini2022Membership} falters in the same way, missing the essential calibration of their test with population data.\footnote{By applying the Bayes rule on the numerator of \lira LR and considering that the denominator is the same as $\Pr(\theta)$ when the OUT reference models are sampled from the population (and not a small finite set), the \lira $\ScoreMIA$ is ${\Pr(x | \theta)}/{\Pr(x)}$.} But, the unreliability of \lira is not only because of this. To better explain our advantage, we present an alternative method for computing our LR~\eqref{eq:likelihood_ratio}, which shows LiRA is an average-case of RMIA. 

In the black-box setting, the divergence between the output distributions of a model trained on $x$ and its corresponding leave-one-out model (which is not trained on $x$), when the distribution is computed over the randomness of the training algorithm, is maximum \textit{when} the models are queried on the differing point~$x$~\citep{ye2023leave}. So, a good approximation for the likelihood ratio in the black-box setting is to evaluate the probability of $f_\theta(x_{\text{features}})$ and $f_\theta(z_{\text{features}})$, where~$f_{\theta}(.)$ is a classification model with parameters $\theta$. A \textbf{direct} way to compute LR in~\eqref{eq:likelihood_ratio} is the following:
\begin{align}
\label{eq:LR_expansion_direct}
    \LR = \frac{\Pr(\theta | x)}{\Pr(\theta | z)} \approx \frac{\Pr(f_\theta(x), f_\theta(z) | x)}{\Pr(f_\theta(x), f_\theta(z) | z)},
\end{align}
where the terms can be computed as in Appendix~\ref{app:direct_likelihood_computation} and~\citep{Carlini2022Membership}. LiRA (online) computes the average of this LR over all $z$, which reduces its test power. In addition, the direct computation of LR requires a large number of reference models for constructing a stable test. We provide an empirical comparison between attack performance of our main computation (\eqref{eq:LR_computation} using Bayes rule) and direct computations of the likelihood ratio in Appendix~\ref{app:direct_likelihood_computation}. The results show that our construction of \relative using the Bayes rule \eqref{eq:LR_computation} is very robust and dominates the direct computation of LR \eqref{eq:LR_expansion_direct} when \textit{a few} reference models are used (Figure~\ref{fig:roc_2_combined}), and they match when we use a large number of reference models (Figure~\ref{fig:roc_32_combined}). Thus, our attack strictly dominates~\lira throughout the power-error (TPR-FPR) curve, and the gap increases significantly when we limit the budget for reference models (See Figure~\ref{fig:aucs_line_graph}). The combination of a pairwise LR and its computation using the Bayesian approach results in our robust, high-power, and low-cost attack, which only requires training of OUT models in offline setting. 

\section{Empirical Evaluation}

In this section, we present a comprehensive empirical evaluation of \relative  and compare its performance with the prior state-of-the-art attacks. Our goal is to analyze:

\begin{enumerate}
    \item Performance of attacks under \textit{limited computation resources} for training reference models. This includes limiting the attack to the offline mode where reference models are pre-trained.
    \item \textit{Ultimate power of attacks} when unlimited number of reference models could be trained (in online mode).
    \item The strength of the attacks in distinguishing members from non-members when target data points are \textit{out-of-distribution}. This reflects the \textit{robustness and usefulness} of attacks for acting as an oracle for partitioning the entire data space into members and non-members.
    \item Impact of \textit{adversary's knowledge} on the performance of attacks (in particular data distribution shift about the population data, and mismatch of network architecture between target and reference models).
\end{enumerate}

\parbf{Setup}
We perform our experiments on CIFAR-10, CIFAR-100, CINIC-10, ImageNet and Purchase-100, which are benchmark datasets commonly used for MIA evaluations. We use the standard metrics, notably the FPR versus TPR curve, and the area under the ROC curve (AUC), for analyzing attacks. The setup of our experiments, including the description of datasets, models, hyper-parameters, and metrics, is presented in Appendix~\ref{app:experiment_setup}.

\parbf{Attack Modes (offline vs. online)}
Reference models are trained on population samples~\citep{Homer2008Resolving, Sankararaman2009Genomic, Shokri2017Membership, Ye2022Enhanced, Watson2022Difficulty}. Adversaries can also simulate the leave-one-out scenario~\citep{ye2023leave}, and perform online attacks by training some reference models on the target data (MIA query)~\citep{Carlini2022Membership}. We consider online attacks as proof-of-concept attacks, due to their high cost in practical scenarios (e.g., reconstruction attacks). We refer to models trained on the target data as IN models, and to those that are not trained on it as OUT models. In the offline mode, all reference models are OUT models.

\parbf{Baseline Attacks}
We mainly compare the performance of \relative with the state-of-the-art attacks~\citep{Ye2022Enhanced, Carlini2022Membership} which had been shown to outperform their prior methods~\citep{Watson2022Difficulty, Shokri2017Membership, Song2021Systematic, Sablayrolles2019White, Long2020Pragmatic, Yeom2018Privacy, jayaraman2020revisiting}. In summary, we use \population~\citep{Ye2022Enhanced} as a baseline attack that does not use any reference models. Note that \population is equivalent to the LOSS attack~\cite{Yeom2018Privacy} which operates by setting threshold on the loss signal from the target model. Both attacks test the same statistic (loss or probability), and the threshold for obtaining a test for a given FPR needs to be set using the population data. We also compare our results with \reference~\citep{Ye2022Enhanced} which uses reference models in an offline mode, and \lira~\citep{Carlini2022Membership} which is an inference attack in online and offline modes. We also compare the results with Quantile Regression~\citep{Bertran2023Scalable} (which improves \population attack in a similar way as~\citep{chang2021privacy} does) outperforming~\cite{Carlini2022Membership} in some scenarios on ImageNet, due to the high uncertainty of the test signal used by~\citep{Carlini2022Membership} on large models. However, \citep{Bertran2023Scalable}[Table 4] shows relative weakness of the attack on other benchmark datasets, even when baseline~\citep{Carlini2022Membership} uses a few reference models. 

\parbf{Reproducing the Attack Results}

To reproduce the results for the prior work, we exclusively use the attacks' implementation as provided by the authors.\footnote{\url{https://github.com/privacytrustlab/ml\_privacy\_meter/tree/master/research/} \cite{Ye2022Enhanced} and RMIA} \footnote{\url{https://github.com/tensorflow/privacy/tree/master/research/mi\_lira\_2021} \cite{Carlini2022Membership}} \footnote{\url{https://github.com/amazon-science/quantile-mia} \citep{Bertran2023Scalable}} We would like to highlight the discrepancy in the empirical results we have obtained for the offline version of \lira (using the authors' code) which is also shown in the prior work~\citep{Ye2022Enhanced, Wen2023Canary}. The obtained attack power is much lower than what is presented in~\citep{Carlini2022Membership}. 

Our attack \relative operates in both offline and online modes (See Appendix~\ref{app:computing_p(x)}). The reader can reproduce our results using our source code.\footnotemark[8] 

\begin{table*}[t!]
  \centering
  \pgfplotstableread[col sep=comma]{data/cifar10_aug_18_num_ref_comparison_4_true.csv}\tab
  \pgfplotstableread[col sep=comma]{data/cifar100_aug_18_num_ref_comparison_4_true.csv}\cifartab
  \pgfplotstableread[col sep=comma]{data/cinic10_num_ref_comparison_4_report_18_aug.csv}\cinictab
  \pgfplotstableread[col sep=comma]{data/cinic10_aug_18_comparison.csv}\cinicpoptab
  \pgfplotstableread[col sep=comma]{data/imagenet_num_ref_comparison_4_true.csv}\imagenettab

      \resizebox{0.96\textwidth}{!}{
        \begin{tabular}{lllllllllllllllll}
          \toprule
          \multirow{3}{*}{\# Ref} &
            \multirow{3}{*}{Attack} &
            \multicolumn{3}{c}{CIFAR-10} &
             &
            \multicolumn{3}{c}{CIFAR-100} &
             &
            \multicolumn{3}{c}{CINIC-10} &
             &
            \multicolumn{3}{c}{ImageNet} \\ \cmidrule{3-5} \cmidrule{7-9} \cmidrule{11-13} \cmidrule{15-17}
           &
             &
            \multicolumn{1}{c}{\multirow{2}{*}{AUC}} &
            \multicolumn{2}{c}{\TPRatFPR} &
             &
            \multicolumn{1}{c}{\multirow{2}{*}{AUC}} &
            \multicolumn{2}{c}{\TPRatFPR} &
            &
            \multicolumn{1}{c}{\multirow{2}{*}{AUC}} &
            \multicolumn{2}{c}{\TPRatFPR} &
            &
            \multicolumn{1}{c}{\multirow{2}{*}{AUC}} &
            \multicolumn{2}{c}{\TPRatFPR} \\ \cmidrule{4-5} \cmidrule{8-9} \cmidrule{12-13} \cmidrule{16-17}
           &
             &
            \multicolumn{1}{c}{} &
            \hundredth &
            \zero &
             &
            \multicolumn{1}{c}{} &
            \hundredth &
            \zero &
             &
            \multicolumn{1}{c}{} &
            \hundredth &
            \zero &
             &
            \multicolumn{1}{c}{} &
            \hundredth &
            \zero \\
\bottomrule
       \end{tabular}
}
  \caption{Performance of attacks where a \textbf{few reference models} are used. All attacks, except LiRA, are \textit{offline} (do not need new reference models to be trained per query). The Quantile Regression attack trains regression-based attack models (instead of reference models) to compute the attack threshold for a certain FPR value. For tuning of hyper-parameters in this attack, \citet{Bertran2023Scalable} suggested training several models which has a considerable overhead compared to training a reference model.  Results are averaged over 10 random target models (Results with standard deviations are shown in Table~\ref{tab:cifar_10_100_cinic_models_1_2_4} in Appendix~\ref{app:fewer_reference_models}). The result of Purchase-100 is illustrated in Table~\ref{tab:purchase_models_2_4_new} in Appendix~\ref{app:purchase-100}.}
  \label{tab:cifar_cinic_imagenet_models_1_2_4}
\end{table*}

\subsection{Inference Attack under Low Computation Budget}

\parbf{Using a Few Reference Models}
Figure~\ref{fig:roc_1_cifar_100} shows the TPR versus FPR tradoff curves for all attacks when the number of reference models is limited to 1.  \relative outperforms \reference~\cite{Ye2022Enhanced} and \lira~\cite{Carlini2022Membership} across all FPR values. Table~\ref{tab:cifar_cinic_imagenet_models_1_2_4} compares the result of attacks in a low-cost scenario where the adversary has access to only a limited number of reference models. Our focus is on attacks in offline mode, where a fixed number of reference models are pre-trained and used to perform the inference on all queries. We also include \lira\online as a benchmark, despite its high computation cost (half reference models need to be trained on the target data). 

The table also incorporates results from the Quantile-Regression attack by~\citet{Bertran2023Scalable}. Although this method does not rely on a reference model, it instead involves training several regression-based models to determine the attack threshold for each sample (to reach a certain FPR value) and optimizing its hyper-parameters through fine-tuning. The attack can be considered a computationally expensive method as it requires many models to be trained to fine-tune its hyper-parameters, and in the end one attack model needs to be constructed for each FPR. 

Our proposed \relative demonstrates its strict dominance over the prior work across all datasets. For instance, with only 2 CIFAR-10 reference models, it achieves around 10\% higher AUC than \reference and \lira and still gains a better TPR at zero FPR. It is important to highlight that, \relative (in offline mode) also dominates \lira (Online). For example, with 4 CIFAR-10 models, it has at least 6\% higher AUC and a significant $3\times$ improvement for TPR at zero FPR, compared to \lira (Online). In the extreme case of using one single reference model, \relative shows at least 24\% higher AUC and also a better TPR at low FPRs than \lira over all datasets. In this case, our attack outperforms the Quantile-Regression attack with respect to both AUC and TPR at low FPR. Moreover, it is important to highlight that we can further improve the performance of \relative by adding more reference models, but there is no room for such an improvement in the Quantile-Regression attack. \reference outperforms \lira (Offline) and results in a relatively high AUC, but it does not perform as well at lower FPR regions. Also, \population does not use any reference models, yet it outperforms \lira in the offline mode. 

\parbf{Using Population Data}
When the number of reference models is low, all attacks must use information from population data to operate effectively. Specifically, \lira necessitates computing the mean and variance of rescaled-logits across reference models. In scenarios with a limited number of reference models (i.e., below 64 models as indicated by \citet{Carlini2022Membership}[Figure 9]), employing a global variance over the population/test data proves advantageous. Similarly, \reference employs smoothing techniques to approximate the CDF for the loss distribution across population/test data to compute percentiles~\cite{Ye2022Enhanced}. The Quantile-Regression attack~\citep{Bertran2023Scalable} also uses a large set of population data to train the attack model. Consequently, there is no significant difference between attacks regarding their demand for additional population data. 

\parbf{Attacking Larger Models with Large Training Sets} 
\setvalue{\tmplabel}{roc_imagenet_1_model-}
\setcounter{figureNumber}{0}
{\tikzset{external/figure name/.add={}{\tmplabel}}
\begin{figure}[!t]
    \centering
    \if\compileFigures1
    \scalebox{\ROCscaleFactor}{
    \begin{tikzpicture}
        \pgfplotstableread[col sep=comma]{data/rocs/imagenet_roc.csv}\roc
        \pgfplotstableread[col sep=comma]{data/rocs/imagenet_auc.csv}\auc
        \begin{axis} [
            scale=0.31,
            width=1.0\linewidth,height=1.0\linewidth,
            ylabel={True Positive Rate},
            ymin=10e-6,ymax=1,xmin=10e-6,xmax=1, 
            xlabel={False Positive Rate},
            xmode=log,ymode=log,
            xtick align = outside,
            ytick align = outside,
            ylabel near ticks,ylabel style={align=center, text width=4cm, font=\small},
            legend pos=south east,
            legend columns=1,
            legend style={nodes={scale=0.7, transform shape}, fill opacity=0.6, draw opacity=1,text opacity=1},
            legend cell align={left},
            every axis plot/.append style={\ROCthickness},
            every axis label/.append style={font=\small},
            reverse legend=true,
            ]
            \addplot [\colorRandom,dashed] table[x=random,y=random, col sep=comma] \roc;
            \addlegendentry {\random}

            \addplot [\colorOnline,no marks,\styleAugOthers] table[x=random,y={LiRA}, col sep=comma] \roc;
            \addlegendentry {\lira}

            \addplot [\colorPopulation,no marks] table[x=random,y={Attack-P}, col sep=comma] \roc;
            \addlegendentry {\population}

            \addplot [\colorReference,no marks] table[x=random,y={Attack-R}, col sep=comma] \roc;
            \addlegendentry {\reference}

            \addplot [pink,no marks,\styleAugOurs] table[x=q-fpr,y={q-tpr}, col sep=comma] \roc;
            \addlegendentry {Quantile-Reg.}

            \addplot [\colorRelative,no marks,\styleAugOurs] table[x=random,y={RMIA}, col sep=comma] \roc;
            \addlegendentry {\relative}


        \end{axis}
    \end{tikzpicture}\hspace*{2em}
    \begin{tikzpicture}
        \pgfplotstableread[col sep=comma]{data/rocs/imagenet_roc.csv}\roc
        \pgfplotstableread[col sep=comma]{data/rocs/imagenet_auc.csv}\auc
        \begin{axis} [
            scale=0.31,
            width=1.0\linewidth,height=1.0\linewidth,
            ylabel={True Positive Rate},
            ymin=10e-6,ymax=1,xmin=10e-6,xmax=1, 
            xlabel={False Positive Rate},
            xtick align = outside,
            ytick align = outside,
            ylabel near ticks,ylabel style={align=center, text width=4cm, font=\small},
            legend pos=south east,
            legend columns=1,
            legend style={nodes={scale=0.7, transform shape}, fill opacity=0.6, draw opacity=1,text opacity=1},
            legend cell align={left},
            every axis plot/.append style={\ROCthickness},
            every axis label/.append style={font=\small},
            reverse legend=true,
            ]
            \addplot [\colorRandom,dashed] table[x=random,y=random, col sep=comma] \roc;
            \addlegendentry {\random}

            \addplot [\colorOnline,no marks,\styleAugOthers] table[x=random,y={LiRA}, col sep=comma] \roc;
            \addlegendentry {\lira (\AUC=\pgfplotstablegetelem{0}{LiRA}\of\auc\pgfplotsretval)}

            \addplot [\colorPopulation,no marks] table[x=random,y={Attack-P}, col sep=comma] \roc;
            \addlegendentry {\population (\AUC=\pgfplotstablegetelem{0}{{Attack-P}}\of\auc\pgfplotsretval)}

            \addplot [\colorReference,no marks] table[x=random,y={Attack-R}, col sep=comma] \roc;
            \addlegendentry {\reference (\AUC=\pgfplotstablegetelem{0}{{Attack-R}}\of\auc\pgfplotsretval)}

            \addplot [pink,no marks,\styleAugOurs] table[x=q-fpr,y={q-tpr}, col sep=comma] \roc;
            \addlegendentry {Quantile-Reg. (\AUC=\pgfplotstablegetelem{0}{{Quantile}}\of\auc\pgfplotsretval)}

            \addplot [\colorRelative,no marks,\styleAugOurs] table[x=random,y={RMIA}, col sep=comma] \roc;
            \addlegendentry {\relative (\AUC=\pgfplotstablegetelem{0}{{RMIA}}\of\auc\pgfplotsretval)}

        \end{axis}
    \end{tikzpicture}
    }
    \else
    \scalebox{\ROCscaleFactor}{
    \includegraphics{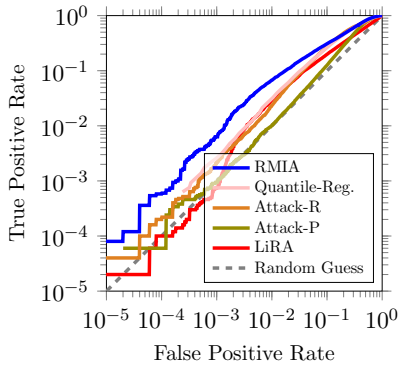}\stepcounter{figureNumber}}
    \fi
    \caption{ROC of attacks against \textbf{ImageNet} models. The result is obtained on one random target model. We use \textbf{1 reference model} (OUT). Table~\ref{tab:cifar_cinic_imagenet_models_1_2_4} reports AUC of attacks. }    \label{fig:roc_imagenet_1_model}
\end{figure}
}

As denoted by~\citet{Bertran2023Scalable}, membership inference attacks might experience performance deterioration against models with larger training set. To study this, Table~\ref{tab:cifar_cinic_imagenet_models_1_2_4} also presents the outcomes of attacks against models trained on the ImageNet dataset, which is 20 times larger than CIFAR datasets. The Quantile-Regression attack outperforms \lirans, even under online \lira conditions with a couple of reference models. However, the efficacy of offline \relative with only one reference model surpasses all other attacks in terms of both AUC and TPR at low FPRs, e.g. showing approximately 2\% higher AUC compared to the Quantile-Regression attack. Furthermore, the performance gap between our attack and the Quantile-Regression attack widens considerably with the inclusion of additional reference models. Figure~\ref{fig:roc_imagenet_1_model} shows the ROC of attacks on ImageNet models when we use 1 reference model. \relative consistently obtains superior TPR across all FPR values.

\begin{table*}[t]
    \centering  
    \pgfplotstableread[col sep=comma]{data/cifar10_augmentation_comparison_new.csv}\tab 
        \pgfplotstableread[col sep=comma]{data/cifar100_18_aug_comparison.csv}\cifartab
        \pgfplotstableread[col sep=comma]{data/cinic10_aug_18_comparison.csv}\cinictab
        \pgfplotstableread[col sep=comma]{data/cifar10_augmentation_comparison_offline_new.csv}\offlinetab
        \resizebox{0.98\textwidth}{!}{
          \begin{tabular}{lllllllllllll}
    \toprule
    \multicolumn{1}{c}{\multirow{3}{*}{Attack}} &
      \multicolumn{3}{c}{CIFAR-10} &
       &
      \multicolumn{3}{c}{CIFAR-100} &
      \multirow{3}{*}{} &
      \multicolumn{3}{c}{CINIC-10} \\ \cmidrule{2-4} \cmidrule{6-8} \cmidrule{10-12}
    \multicolumn{1}{c}{} &
      \multicolumn{1}{c}{\multirow{2}{*}{AUC}} &
      \multicolumn{2}{c}{\TPRatFPR} &
       &
      \multicolumn{1}{c}{\multirow{2}{*}{AUC}} &
      \multicolumn{2}{c}{\TPRatFPR} &
       &
      \multicolumn{1}{c}{\multirow{2}{*}{AUC}} &
      \multicolumn{2}{c}{\TPRatFPR} \\ \cmidrule{3-4} \cmidrule{7-8} \cmidrule{11-12}
    \multicolumn{1}{c}{} &
      \multicolumn{1}{c}{} &
      \hundredth &
      \zero &
       &
      \multicolumn{1}{c}{} &
      \hundredth &
      \zero &
       &
      \multicolumn{1}{c}{} &
      \hundredth &
      \zero \\
\bottomrule
\end{tabular}
                }
    \caption{Performance of attacks using a \textit{large number of reference models}, averaged over 10 random target models. The top part (the first four rows) represents offline attacks, using 127 OUT reference models (except Attack-P that uses no reference models). The second part (the bottom two rows) represents online attacks, using 127 OUT and 127 IN models.}
    \label{tab:other_datasets_std}
\end{table*}


\parbf{Using More Reference Models in the Offline Mode}
Table~\ref{tab:other_datasets_std} compares the performance of offline attacks when using a larger number of OUT models (127 models). The \relative consistently outperforms other offline attacks across all datasets. \relative has a much larger power than the \reference (which is the strongest offline attack in the prior work). \reference may wrongly reject a typical member as a non-member solely because it has a higher probability in reference models. As a result, it yields 5\%-10\% lower AUC than \relative across all datasets. \relative is designed to overcome these limitations by considering both the characteristics of the target sample within reference models and its relative probability among other population records.

Comparing Table~\ref{tab:cifar_cinic_imagenet_models_1_2_4} and Table~\ref{tab:other_datasets_std} shows that the AUC of \relative when using a few reference models is almost the same as that of using a large number (127) of models. This reveals that the overall power of the attack is not significantly dependent on having many reference models. However, when we increase the number of reference models the attack TPR in low FPR regions increases.

To better analyze the difference between the attacks, we compare the variation in MIA scores of member and non-member samples across all attacks in Appendix~\ref{app:mia_test_score_comparison}.

\setvalue{\tmplabel}{aucs_single_line_graph_cifar10-}
\setcounter{figureNumber}{0}
{\tikzset{external/figure name/.add={}{\tmplabel}}
\begin{figure}[t]
    \centering
    \if\compileFigures1    \scalebox{\AUCMainscaleFactor}{
\begin{tikzpicture}
    \pgfplotstableread[col sep=comma]{data/aucs/cifar10_online_fig1.csv}\onlinetab
    \pgfplotstableread[col sep=comma]{data/aucs/cifar10_offline_fig1.csv}\offlinetab
    \begin{axis}[
        legend pos=south east,
        legend columns=2,
        scale=0.9,
        ylabel={AUC},
        xlabel={Number of Reference Models},
        ymin=44,ymax=75,
        xmode=log,
        log basis x={2},
        legend style={nodes={scale=0.8, transform shape}, fill opacity=0.6, draw opacity=1,text opacity=1},
        reverse legend=true,
        every axis plot/.append style={thick},
        ]


        \addplot [\colorLiRAOffline, mark=square] table[x=nums, y={liraoffline 18}, col sep=comma] \offlinetab;
        \addlegendentry {\lira \offline}

        \addplot[\colorPopulation, dashed] coordinates {(1,58.19) (254,58.19)};
        \addlegendentry {\population}

        \addplot [\colorReference, mark=o] table[x=nums, y={reference 0}, col sep=comma] \offlinetab;
        \addlegendentry {\reference}


        \addplot [\colorOnline, mark=x] table[x=nums, y={liraonline 18}, col sep=comma] \onlinetab;
        \addlegendentry {\lira \online}
        
        \addplot [\colorRelativeOffline, mark=triangle] table[x=nums, y={relativeofflineln 18}, col sep=comma] \offlinetab;
        \addlegendentry {\relative \offline}
        
        \addplot [\colorRelative, mark=diamond] table[x=nums, y={relativeonline 18}, col sep=comma] \onlinetab;
        \addlegendentry {\relative \online}

    \end{axis}
\end{tikzpicture}
}
    \else
    \scalebox{\AUCMainscaleFactor}{
    \includegraphics{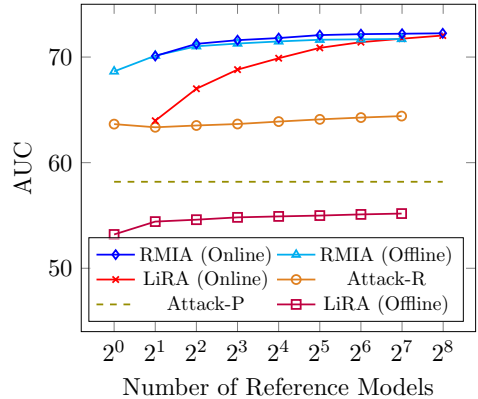}\stepcounter{figureNumber}}
    \fi
    \caption{Number of reference models versus the AUC of the attacks on CIFAR-10. In online attacks, half of reference models need to be trained per each MIA query. }
    \label{fig:aucs_line_graph_cifar10}
\end{figure}
}

\subsection{Ultimate Power of (Online) Inference Attacks}

Table~\ref{tab:other_datasets_std} presents the performance of all attacks on models trained with different datasets where we have enough resources to train many (254 IN and OUT) reference models. In this setting, \relative\online performs better than \lira\online (which gains significantly from a large number of reference models) with respect to AUC. In this case, even minor AUC improvements are particularly significant when approaching the maximum leakage of the training algorithm in the corresponding MIA game. What is, however, more important to note is that \textit{\relative (Online)'s TPR at zero FPR is significantly better than that of \lira\online (by up to 50\%)}. It is important to note that our offline attack achieves performance comparable to online attacks, which is quite remarkable, considering the large gap between the cost associated with offline and online attacks. 

We can further improve attacks by querying the target model with multiple \textit{augmentations} of input query, obtained via simple mirror and shift operations on image data.  In the multi-query setting, we use
majority voting on our hypothesis test in~\eqref{eq:likelihood_ratio}: query $x$ is considered to dominate population record $z$ if more than half of all augmentations of $x$ dominate $z$. See Appendix~\ref{app:augmented_queries} for the details. \relative can leverage this technique to a significantly greater extent, and can achieve \textit{a $4\times$ improvement in TPR at zero FPR and about 4.6\% higher AUC} compared to \lirans. Our results are based on applying 18 augmented queries.

\subsection{Dependency on Availability of Reference Models}

Inference attacks must be robust to changes in their assumptions about the prior knowledge of adversary, for example population data available to train reference models. Low-cost MIA methods should also be less dependent on availability of a large number of reference models.

Figure~\ref{fig:aucs_line_graph_cifar10} presents the number of reference models needed to achieve an overall performance for attacks (computed using their AUC). Note that online attacks need at least 2 reference models (1 IN, 1 OUT). As opposed to other strong attacks, \relative obtains stable results that do not change significantly when reducing the number of reference models. \relative does gain from increasing the number of reference models, but even with a small number of models, it is very close to its maximal overall performance. However, \lira\online displays a great sensitivity to the changes in number of reference models (with a huge AUC gap of about 8\% in CIFAR-10 models when comparing 2 models versus 254 models), underscoring the necessity of a large number of models for this attack to function effectively. Also it is remarkable to note that the \relative (Offline) is stronger than \lira\online attacks unless when we use hundreds of reference models. Appendix~\ref{app:roc_datasets} presents additional results obtained with various number of models trained on other datasets, showing a comparable superiority among attacks similar to what was observed with CIFAR-10.

\subsection{Robustness of Attacks against OOD Non-members}

We challenge membership inference attacks by testing them with non-member out-of-distribution (OOD) data. A strong MIA should be able to rule out \textit{all non-members} regardless of whether they are from the same distribution as its training data or not. This is of a great importance also in scenarios where we might use MIA oracles in applications such as data extraction attacks~\citep{Carlini2021Extracting}. In such cases, it is essential for the attack to remain accurate (high TPR for all FPR) on out-of-distribution (OOD) non-member data, while detecting members. Note that naive filtering techniques cannot be used to solve this problem. While OOD samples generally exhibit lower confidence levels compared to in-distribution samples, filtering them just based on confidence leads to a low TPR by rejecting hard member samples.

To examine the robustness and effectiveness of attacks in presence of OOD samples, we train our models with CIFAR-10 and use samples from a different dataset (CINIC-10) to generate OOD non-member test queries. The setting of the MIA game described here diverges from the game outlined in Definition~\ref{def:mia_game}, due to $x \not\sim \pi$. Consequently, the outcomes cannot be directly compared with those of the original game. We focus on offline attacks, as it is practically infeasible to train IN models for each and every individual OOD test sample, especially when dealing with a large volume of queries. 

Figure~\ref{fig:ood_roc} illustrates the ROC curves. \relative has a substantial performance advantage (at least 21\% higher AUC) over \referencens, with \lira not performing much better than random guessing. \population demonstrates strong AUC performance by capitalizing on the disparity between confidence/loss of OOD and in-distribution samples. However, as explained above, it struggles at achieving good TPR at low FPRs. It is also not a powerful method to detect in-distribution non-members, as shown in the results in previous figures and tables. Conducting our hypothesis test by assessing the pair-wise LR between the target sample $x$ and several \mbox{\textit{in-distribution}} $z$ samples enables us to capture significant differences between $\frac{\Pr(x|\theta)}{\Pr(x)}$ and $\frac{\Pr(z|\theta)}{\Pr(z)}$ in \eqref{eq:LR_computation} to effectively distinguish OOD queries from members. We also present the results of using pure noise as non-member test queries in Figure~\ref{fig:ood_noise_roc} (Appendix~\ref{app:robustness_ood}), revealing a wider disparity between \relative and other methods compared to OOD samples

\setvalue{\tmplabel}{rocs_ood_pop_cifar10-}
\setcounter{figureNumber}{0}
{\tikzset{external/figure name/.add={}{\tmplabel}}
\begin{figure}[t]
    \centering
    \scalebox{\AUCMainscaleFactor}{
    \includegraphics{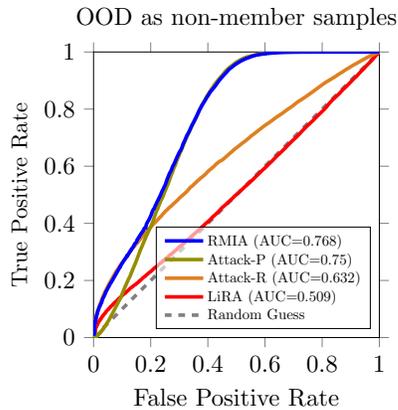}\stepcounter{figureNumber}}
    \caption{ROC of offline attacks using models trained on CIFAR-10, while non-member test queries are OOD samples from CINIC-10. We use 127 reference models.}
    \label{fig:ood_roc}
\end{figure}
}

\subsection{Robustness to Model and Data Distribution Shift}

When assessing membership inference attacks, it is crucial to examine how the attack's performance is influenced when the adversary lacks precise knowledge about the distribution of training data and also the structure of the target model. Therefore, we compare the result of attacks when the reference models are trained on different datasets than the target models. More specifically, the target models are trained on CIFAR-10, while the reference models are trained on CINIC-10.  The performance of all attacks is affected when there is a data distribution shift between the training set of the target model and the reference models. However, compared with other attacks, \relative always obtains a higher AUC (e.g. by up to 25\% in comparison with \lirans, using 2 reference models) and a better TPR at low FPRs. This confirms that our attack remains effective  when pre-trained models on different datasets are used as reference models, requiring zero training cost. Table~\ref{tab:cifar_cinic_models_2_4_new} in Appendix~\ref{app:data_shift} presents the detailed result of this experiment.

We also study the impact of network architecture change between the target model and the reference models. While the optimal performance of all attacks is noted when both the target and reference models share a similar architecture, the superiority of our attack becomes more pronounced in the presence of architecture shifts (we observe up to 3\% increase in the AUC gap between our attack and others).
See Appendix~\ref{app:architecture_variation} for the details of the empirical results.

\subsection{Analyzing RMIA Parameters}
\label{app:reference_records}

\parbf{Number of $z$ samples} Our attack evaluates the likelihood ratio of the target model on a target data $x$ versus other population samples $z$. Computing the LR versus the population samples enables us to compute reliable test statistics for our attack. Table~\ref{tab:z_fraction} shows how the result of \relative changes, as we use different number of reference samples. The AUC increases when we increase the number of $z$ samples, but it is noteworthy that using 2500 population samples, equivalent to 10\% of the size of the models' training set, yields results comparable to those obtained with a 10 fold larger population set. Moreover, the TPR at low FPRs is still high even when we consider just 250 reference samples. The trend of results remains consistent when using fewer reference models (See Table~\ref{tab:z_fraction_1_ref} in Appendix~\ref{app:z_samples}). In the default setting, when performing an attack on a target model with sample $x$, we use all non-members of the target model as the set of $z$ samples. We exclude the query $x$ itself from being used as $z$.

\begin{table}[t]
\centering
\pgfplotstableread[col sep=comma]{data/relative_ratio_comparison_table_updated.csv}\relativetab
    \resizebox{0.45\textwidth}{!}{
    \begin{tabular}{llllll}
    \toprule
    \multirow{2}{*}{\# population samples $z$} & \multirow{2}{*}{AUC}   & \multicolumn{2}{c}{\TPRatFPR}  \\ \cmidrule{3-4}
                                &                        & \hundredth & \zero             \\ \midrule
    25 samples                                                             & \pgfplotstablegetelem{0}{{z 25}}\of\relativetab\pgfplotsretval$\,\pm\,$\pgfplotstablegetelem{1}{{z 25}}\of\relativetab\pgfplotsretval                                      & \pgfplotstablegetelem{4}{{z 25}}\of\relativetab\pgfplotsretval$\,\pm\,$\pgfplotstablegetelem{5}{{z 25}}\of\relativetab\pgfplotsretval                                        & \pgfplotstablegetelem{6}{{z 25}}\of\relativetab\pgfplotsretval$\,\pm\,$\pgfplotstablegetelem{7}{{z 25}}\of\relativetab\pgfplotsretval                                      \\
    250 samples                                                             & \pgfplotstablegetelem{0}{{z 250}}\of\relativetab\pgfplotsretval$\,\pm\,$\pgfplotstablegetelem{1}{{z 250}}\of\relativetab\pgfplotsretval                                      & \pgfplotstablegetelem{4}{{z 250}}\of\relativetab\pgfplotsretval$\,\pm\,$\pgfplotstablegetelem{5}{{z 250}}\of\relativetab\pgfplotsretval                                        & \pgfplotstablegetelem{6}{{z 250}}\of\relativetab\pgfplotsretval$\,\pm\,$\pgfplotstablegetelem{7}{{z 250}}\of\relativetab\pgfplotsretval                                      \\
\bottomrule
    \end{tabular}
  }
\caption{Performance of \relative\online using different number of random $z$ samples. The number of reference models, trained on CIFAR-10, is 254 (127 IN, 127 OUT). Here, we do not use augmented queries. Results are averaged over 10 random target models. 
}
\label{tab:z_fraction}
\end{table}

\parbf{Sensitivity to Pair-Wise LR Test Threshold $\gamma$} The result of our experiments, presented in Appendix~\ref{app:relation_gamma_beta}, shows that our attack’s performance is consistent against changes in the value of $\gamma$. Both AUC and FPR-TPR curve remain relatively stable with small changes to $\gamma$ (except for a considerably high value of $\gamma$). In fact, by adjusting the value of the threshold $\beta$, we achieve roughly the same result across different $\gamma$ values.

\subsection{Applying \relative to Other ML Algorithms}

To assess how attacks perform against alternative ML algorithms, we investigate the privacy risks of Gradient Boosting Decision Tree (GBDT) algorithms. In this case, the TPR obtained by our attack consistently outperforms all other attacks, particularly by an order of magnitude at zero FPR, as demonstrated in Figure~\ref{fig:gradient_boosted_decision_trees} (Appendix~\ref{app:gradient_boosting_decision_tree}).

\section{Conclusions}

We argue that MIA tests, as privacy auditing tools, must be stress-tested with low computation budget, few available reference models, and changes to data distribution and models. A strong test is the one that can outperform others in these scenarios, and not only in typical scenarios. We present a novel statistical test for MIA, and a series of evaluation scenarios to compare MIAs based on their efficiency and robustness. \textbf{RMIA can be reliably used to audit privacy risks in ML under realistic practical assumptions.} 

\newpage
\clearpage

\section*{Acknowledgements}

The authors would like to thank Vincent Bindschaedler, Jiayuan Ye, and Hongyan Chang for their comments on the earlier versions of this work. The authors would also like to thank the reviewers for their valuable comments and for a very engaging discussion during the rebuttal process. The work of Reza Shokri was supported by the Asian Young Scientist Fellowship 2023, and the Ministry of Education, Singapore, Academic Research Fund (AcRF) Tier 1.


\newpage
\clearpage

\onecolumn
\tableofcontents
\clearpage

\appendix
\section{Experimental Setup}
\label{app:experiment_setup}

To conduct attacks, we must first train models. We adopt the same training setup as in \cite{Carlini2022Membership}, wherein, for a given dataset, we train our models on randomly selected training sets, with each set containing half of the dataset. Moreover, each sample of the dataset is included in exactly half of the reference models' training set. We pick our target models from the set of trained models at random. It is worth noting that in this setting, the training set of a target model can overlap by 50\% with each reference model.

\parbf{Datasets} We report the attack results on models trained on five different datasets, traditionally used for membership inference attack evaluations. For CIFAR-10 \cite{He2016Deep} (a traditional image classification dataset), we train a Wide ResNets (with depth 28 and width 2) for 100 epochs on half of the dataset chosen at random. For CIFAR-100 and CINIC-10 (as other image datasets), we follow the same process as for CIFAR-10 and train a wide ResNet on half of the dataset\footnote{For CINIC-10, we randomly choose 50k samples (out of 270k samples) for training models. In Appendix~\ref{app:larger_datasets}, we show the result of attacks when training models with larger subsets of CINIC-10.} We set the batch size to 256. We assess the impact of attacks on larger datasets by examining the ImageNet dataset, comprising approximately 1.2 million images with 1000 class labels. We train the ResNet-50 on half of the dataset for 100 epochs, with a batch size of 256, a learning rate of 0.1, and a weight decay of 1e-4. We also include the result of attacks on Purchase-100 dataset (a tabular dataset of shopping records)~\cite{Shokri2017Membership}, where models are 4-layer MLP with layer units=[512, 256, 128, 64], trained on 25k samples for 50 epochs. Table~\ref{tab:model_accuracy} displays the accuracy of models trained on various datasets.

\begin{table}[htbp]
    \centering
    \begin{tabular}{ccc}
        \hline
        Dataset & Train Accuracy & Test Accuracy \\
        \hline
        CIFAR-10 & 99.9\% & 92.4\% \\
        CIFAR-100 & 99.9\% & 67.5\% \\
        CINIC-10 & 99.5\% & 77.2\% \\
        ImageNet & 90.2\% & 58.6\% \\
        Purchase-100 & 100\% & 83.4\% \\
        \hline
    \end{tabular}
    \caption{Average accuracy of models trained on different datasets.}
    \label{tab:model_accuracy}
\end{table}
\parbf{Evaluation Metrics} We measure the performance of each attack using two underlying metrics: its true positive rate (TPR), and its false positive rate (FPR), over all member and non-member records of random target models. Then, we use the ROC curve to reflect the trade-off between the TPR and FPR of an attack, as we sweep over all possible values of threshold $\beta$ to build different FPR tolerance. The AUC (area under the ROC curve) score gives us the average success across all target samples and measures the overall strength of an attack. Inspired from previous discussions in \cite{Carlini2022Membership}, we also consider TPR at very low FPRs. More precisely, we focus on TPR at 0\% FPR, a metric that has seen limited usage in the literature. When attacking a target model, all samples in the population data are used as input queries. Hence, for each target model, half of queries are members and the other half are non-members. 

\section{Details of RMIA and its Evaluation}
\subsection{Pseudo-code of RMIA}
\label{app:rmia_pseudocode}
Membership inference attacks require training reference (or shadow) models in order to distinguish members from non-members of a given target model. We train $k$ reference models on a set of samples randomly drawn from the population data. 
Concerning the training of models, we have two versions of RMIA. The \relative\online (similar to~\citet{Carlini2022Membership}) trains reference models separately for each target data (MIA test query). Specifically, upon receiving a MIA test query $x$, we train $k$ IN models that include $x$ in their training set. However, such a training is costly and impractical in many real-world scenarios due to the significant resource and time requirements. On the other hand, the offline version uses only $k$ pre-trained reference models on randomly sampled datasets, avoiding any training on test queries (as it exclusively uses OUT models). The two versions differ in the way they compute the normalizing term $\Pr(x)$ in~\eqref{eq:LR_computation}. The online algorithm computes the terms by averaging prediction probabilities over all IN and OUT models in an unbiased manner (\eqref{eq:p_online}), while the offline algorithm computes the terms solely with OUT models based on~\eqref{eq:p_in_approximation}. In the offline mode, we approximate what $\Pr(x)$ would have been if we had access to IN models (see Appendix~\ref{app:computing_p(x)} for more explanations).

Algorithm~\ref{psuedocode:rmia} outlines the pseudo-code for RMIA. The input to this algorithm includes the target model $\theta$, the target sample~$x$, the threshold $\gamma$, and the set of reference models denoted by $\Theta$, which are trained prior to computing the MIA score of~$x$. Additionally, it takes an $\textit{online}$ flag indicating if we run the algorithm in online mode. If it we are in offline mode, a scaling factor $a$ is also obtained as described in Appendix~\ref{app:computing_p(x)}. The output of the algorithm is $\ScoreMIA$, which is then used according to~\eqref{eq:MIA_test} to infer the membership of $x$. As described in Algorithm~\ref{psuedocode:rmia}, we first select our reference population samples (i.e., the set $Z$) from the population dataset. If we are operating in the online mode, we train $k$ IN models using target data~$x$ and a randomly selected dataset form the population data (line~5-9). Depending on the attack mode, we query the reference models to obtain the average prediction probabilities over reference models as $\Pr(x)$ (line~10 and line~12). In the case of offline RMIA, line~13 approximates $\Pr(x)$ using \eqref{eq:p_in_approximation}.  Then, we calculate the ratio of prediction probabilities between the target model and the reference models for sample $x$ as $Ratio_x$ (line~15). The same routine is applied to obtain the ratio for each reference sample $z \in Z$ (line~16-18).

According to \eqref{eq:LR_computation}, the division of the computed ratio for sample $x$ by the obtained ratio for sample $z$ determines $\LR$ to assess if $z$ is $\gamma$-dominated by $x$ (line~19). The fraction of $\gamma$-dominated reference samples establishes the MIA score of the target sample $x$ in RMIA (line~23).

\begin{algorithm}[t]
\caption{\textbf{MIA Score Computation with RMIA.} The input to this algorithm is $k$ reference models $\Theta$, the target model~$\theta$, target (test) sample $x$, parameter $\gamma$, and a scaling factor $a$ as described in Appendix~\ref{app:computing_p(x)}. We assume the reference models $\Theta$ are pre-trained on random samples from a population dataset available to adversary; each sample from the population dataset is included in training of half of reference models. The algorithm also takes an $\textit{online}$ flag which indicates whether we intend to run MIA in the online mode.}
\label{psuedocode:rmia}
\begin{algorithmic}[1]
    \STATE Randomly choose a subset $Z$ from the population dataset 
    \STATE $C \gets 0$ 
    \IF {$\textit{online}$}
        \STATE $\Theta_{in} \gets \emptyset$ 
        \FOR {$k$ times}
            \STATE $D_i \gets$ randomly sample a dataset from population data $\pi$
            \STATE $\theta_{x} \gets \Tau(D_i \cup x)$
            \STATE $\Theta_{in} \gets \Theta_{in} \cup \{\theta_{x}\}$
        \ENDFOR
    \STATE $\Pr(x) \gets \frac{1}{2k}\big(\sum_{\theta' \in \Theta}{\Pr(x|\theta')} + \sum_{\theta' \in \Theta_{in}}{\Pr(x|\theta')}\big)$  \qquad {\color{lightgray}(See \eqref{eq:p_online})}
    \ELSE
    \STATE $\Pr(x)_{OUT} \gets \frac{1}{k}\sum_{\theta' \in \Theta}{\Pr(x|\theta')}$
    \STATE $\Pr(x) \gets \frac{1}{2}\big((1+a).\Pr(x)_{OUT} + (1-a)\big)$ \qquad {\color{lightgray}(See Appendix~\ref{app:computing_p(x)})}
    \ENDIF
    \STATE $Ratio_x \gets \frac{\Pr(x|\theta)}{\Pr(x)}$ 
    \FOR {each sample $z$ in $Z$}
        \STATE $\Pr(z) \gets \frac{1}{k}\sum_{\theta' \in \Theta}{\Pr(z|\theta')}$
        \STATE $Ratio_z \gets \frac{\Pr(z|\theta)}{\Pr(z)}$
        \IF {$(Ratio_x/Ratio_z) > \gamma$}
        \STATE $C \gets C + 1$
        \ENDIF
    \ENDFOR
    \STATE $\ScoreMIA \gets C/|Z|$ \qquad {\color{lightgray}(See \eqref{eq:MIA_score})}
\end{algorithmic}
\end{algorithm}

\subsection{Likelihood Ratio Computation in \eqref{eq:LR_computation}}
\label{app:calculate_pr}

\subsubsection{Computing $\Pr(x | \theta)$}
\label{app:calculate_p_x_theta}

Let $\theta$ be a classification (neural network) model that maps each input data $x$ to a probability distribution across $d$ classes. Assume data point $x$ is in class $y$. Let ${c(x)=\langle c_1, \cdots, c_d \rangle}$ be the output vector (logits) of the neural network for input $x$, before applying the final normalization. We denote the normalized prediction probability of class $y$ for the input $x$ as $f_{\theta}(x)_y$.  For the Softmax function, the probability is given by $$f_{\theta}(x)_y=\frac{e^\frac{c_y}{T}}{\sum_{i=1}^{d}e^\frac{c_i}{T}},$$ where $T$ is a temperature constant. We can use this to estimate $\Pr(x | \theta)$. However, there are many alternatives to Softmax probability proposed in the literature to improve the accuracy of estimating $\Pr(x | \theta)$, using the Taylor expansion of the exponential function, and using heuristics to refine the relation between the probabilities across different classes~\cite{Brebisson2016Exploration, Liu2016Large, Liang2017Soft, Banerjee2021Exploring}. A more recent method, proposed by~\citet{Banerjee2021Exploring}, combines the Taylor expansion with the soft-margin technique to compute the confidence as:

\begin{equation}
\label{eq:softmargin_taylor_softmax}
f_{\theta}(x)_y=\frac{apx(c_y-m)}{apx(c_y-m)+\sum_{i \neq y}apx(c_i)},
\end{equation}
where $apx(a)=\sum_{i=0}^{n}{\frac{{a}^i}{i!}}$ is the $n$th order Taylor approximation of $e^{a}$, and $m$ is a hyper-parameter that controls the separation between probability of different classes. 

In Table~\ref{tab:prob_functions}, we compare the result of RMIA obtained with four different confidence functions, i.e. Softmax, Taylor-Softmax \cite{Brebisson2016Exploration}, Soft-Margin Softmax (SM-Softmax)~\cite{Liang2017Soft}, and the combination of last two functions (SM-Taylor-Softmax)~\cite{Banerjee2021Exploring}, formulated in \eqref{eq:softmargin_taylor_softmax}. Based on our empirical results, we set the soft-margin $m$ and the order $n$ in Taylor-based functions to 0.6 and 4, respectively. The temperature ($T$) is set to 2 for CIFAR-10 models. While the performance of all functions is closely comparable, SM-Taylor-Softmax stands out by achieving a slightly higher AUC and a rather better TPR at low FPRs. Therefore, we use this function for CIFAR and CINIC-10 datasets. In Figure~\ref{fig:auc_vs_hyperparams}, we present the performance sensitivity of our attack in terms of AUC concerning three hyper-parameters in this function: order $n$, soft-margin $m$, and temperature $T$. The AUC of RMIA appears to be robust to variations in $n$, $m$ and $T$. For $n \geq 3$, the results are consistent, but employing lower orders leads to a poor Taylor-based approximation for Softmax, as reported by \citet{Banerjee2021Exploring}. We use Softmax (with temperature of 1) for our ImageNet and Purchase-100 datasets. We do not apply confidence functions to other attacks, as they do not use confidence as their signal. For instance, LiRA~\cite{Carlini2022Membership} operates with rescaled logit, while Attack-P and Attack-R~\cite{Ye2022Enhanced} use the loss signal. In fact, we follow the original algorithms and codes to reproduce their results.

\begin{table}[t]
    \centering
    \pgfplotstableread[col sep=comma]{data/signal_comparison_table_updated.csv}\tab 
    	\resizebox{0.55\textwidth}{!}{
    		\begin{tabular}{lllll}
    		\toprule
    		 \multicolumn{1}{c}{\multirow{2}{*}{Confidence Function}} & \multicolumn{1}{c}{\multirow{2}{*}{AUC}}                                                                                                                        & \multicolumn{2}{c}{\TPRatFPR}                                                                                                                                                                                                                                                                                                     \\ \cmidrule{3-4}
    										  \multicolumn{1}{c}{}                                               & \multicolumn{1}{c}{}                                                                                                                                                                                                                                                                               & \multicolumn{1}{c}{\hundredth}                                                                                                                                   & \multicolumn{1}{c}{\zero}                                                                                                                                   \\ \midrule
    															   Softmax                                                            & \pgfplotstablegetelem{0}{{softmax}}\of\tab\pgfplotsretval$\,\pm\,$\pgfplotstablegetelem{1}{{softmax}}\of\tab\pgfplotsretval                                                                             & \pgfplotstablegetelem{4}{{softmax}}\of\tab\pgfplotsretval$\,\pm\,$\pgfplotstablegetelem{5}{{softmax}}\of\tab\pgfplotsretval                                        & \pgfplotstablegetelem{6}{{softmax}}\of\tab\pgfplotsretval$\,\pm\,$\pgfplotstablegetelem{7}{{softmax}}\of\tab\pgfplotsretval                                      \\
\bottomrule

    		\end{tabular}
    	}
    \caption{ Performance of \relative obtained with various confidence functions. The number of reference models, trained with CIFAR-10, is 254. The soft-margin $m$ and the order $n$ in Taylor-based functions are 0.6 and 4, respectively.}
    \label{tab:prob_functions}
\end{table}

\setvalue{\tmplabel}{auc_vs_hyperparams-}
\setcounter{figureNumber}{0}
{\tikzset{external/figure name/.add={}{\tmplabel}}
\begin{figure*}[h]
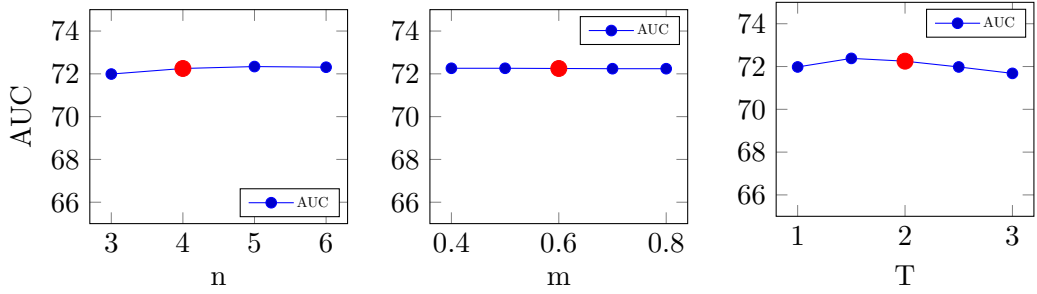

    \centering
    \if\compileFigures1
    \begin{tikzpicture}
    \centering
    \pgfplotstableread[col sep=comma]{data/stats_and_ns.csv}\table
    \begin{axis}[
        legend pos=south east,
        ymin=65,ymax=75,
        scale=0.5,
        ylabel={AUC},
        xlabel={n},
        legend style={nodes={scale=0.5, transform shape}},
        ]
        \addplot table[x=x, y=0] \table;
        \addlegendentry {AUC}
    \end{axis}
\end{tikzpicture}\hspace*{2em}
\begin{tikzpicture}
    \centering
    \pgfplotstableread[col sep=comma]{data/stats_and_ms.csv}\table
    \begin{axis}[
        legend pos=north east,
        ymin=65,ymax=75,
        scale=0.5,
        xlabel={m},
        legend style={nodes={scale=0.5, transform shape}},
        ]
        \addplot table[x=x, y=0] \table;
        \addlegendentry {AUC}
    \end{axis}
\end{tikzpicture}\hspace*{2em}
\begin{tikzpicture}
    \centering
    \pgfplotstableread[col sep=comma]{data/stats_and_temperature.csv}\table
    \begin{axis}[
        legend pos=north east,
        ymin=65,ymax=75,
        scale=0.5,
        xlabel={T},
        legend style={nodes={scale=0.5, transform shape}},
        ]
        \addplot table[x=x, y=0] \table;
        \addlegendentry {AUC}
    \end{axis}
\end{tikzpicture}
    \else
    \includegraphics{fig/figure-\tmplabel\thefigureNumber.pdf}\stepcounter{figureNumber}\hspace*{1em}
    \includegraphics{fig/figure-\tmplabel\thefigureNumber.pdf}\stepcounter{figureNumber}\hspace*{1em}
    \includegraphics{fig/figure-\tmplabel\thefigureNumber.pdf}\stepcounter{figureNumber}
    \fi
    \caption{AUC of our attack (RMIA) obtained by using different values of $n$ (order in Taylor function), $m$ (soft-margin) and $T$ (temperature) in SM-Taylor-Softmax function. When modifying one parameter, we hold the values of the other two parameters constant at their optimal values. Here, we use 254 reference models trained on CIFAR-10. Results are averaged over 10 target models. The red points indicate the default values used in our experiments.}
    \label{fig:auc_vs_hyperparams}
\end{figure*}
}

\subsubsection{Computing $\Pr(x)$}
\label{app:computing_p(x)}

Recall that $\Pr(x)$ is the normalizing constant for the Bayes rule in computing our LR. See \eqref{eq:px}. We compute it as the empirical average of $\Pr(x | \theta')$ on reference models $\theta'$ trained on random datasets $D$ drawn from the population distribution~$\pi$. This is then used in pairwise likelihood ratio \eqref{eq:LR_computation}. 

\setvalue{\tmplabel}{auc_vs_offline_params-}
\setcounter{figureNumber}{0}
{\tikzset{external/figure name/.add={}{\tmplabel}}
\begin{figure}[h]
    \centering
    \if\compileFigures1
    \begin{tikzpicture}
    \centering
    \pgfplotstableread[col sep=comma]{data/stats_and_offline_as.csv}\table
    \begin{axis}[
        legend pos=north east,
        ymin=65,ymax=75,
        scale=0.8,
        ylabel={AUC},
        xlabel={a},
        legend style={nodes={scale=0.5, transform shape}},
        ]
        \addplot table[x=x, y=0] \table;
        \addlegendentry {AUC}
    \end{axis}
\end{tikzpicture}
    \else
    \scalebox{\SensiscaleFactor}{\includegraphics{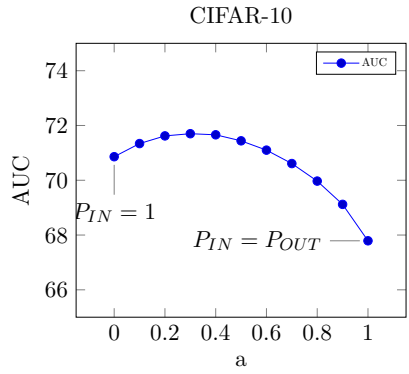}\stepcounter{figureNumber}}
    \fi
    \caption{AUC of offline RMIA obtained by using different values of $a$ in the linear approximation function. Here, we use 127 reference models (OUT) trained on CIFAR-10. }
    \label{fig:auc_vs_offline_params}
\end{figure}
}

In order to compute $\Pr(x)$, we need to train reference models~$\theta'$. Note that the reference models must be sampled in an unbiased way with respect to whether $x$ is part of their training data. This is because the summation in~\eqref{eq:px} is over all $\theta'$, which can be partitioned to the set of models trained on $x$ (IN models), and the set of models that are not trained on x (OUT models). Let $\bar{x}$ denote that a training set does not include~$x$. Let $\theta'_{x}$ denote an IN model trained on dataset $D_x$ ($x \in D_x$) and $\theta'_{\Bar{x}}$ be an OUT model trained on dataset $D_{\Bar{x}}$ ($x \notin D_{\Bar{x}}$). Then, from~\eqref{eq:px}, we have:
\begin{align}
    \begin{split}
        \Pr(x) &= \sum_{\theta', D} \Pr(x | \theta') \Pr(\theta' | D) \Pr(D) \\
        &= \sum_{\theta'_{x},D_{x}} \Pr(x | \theta'_{x})  \Pr(\theta'_{x} | D_x) \Pr(D_x) + \sum_{\theta'_{\Bar{x}},D_{\Bar{x}}} \Pr(x | \theta'_{\Bar{x}})  \Pr(\theta' | D_{\Bar{x}}) \Pr(D_{\Bar{x}})
    \end{split}
\end{align}
The two sums on the right-hand side of the above equation can be computed empirically using sampling methods. Instead of integrating over all possible datasets and models, we sample datasets~$D_x$ and~$D_{\bar{x}}$ and models $\theta'_x$ and $\theta'_{\bar{x}}$, and compute the empirical average of $\Pr(x | \theta'_{x})$ and $\Pr(x | \theta'_{\bar{x}})$ given the sampled models. We sample $D_{\bar{x}}$ from the probability distribution $\Pr(D_{\bar{x}})$, which is the underlying data distribution $\pi$.  For sampling $D_x$, we sample a dataset from $\pi$ and add $x$ to the dataset. We sample $\theta'_x$ and $\theta'_{\bar{x}}$ by training models on $D_x$ and $D_{\bar{x}}$, respectively.  

In the online setting for MIA, we can empirically estimate $\Pr(x)$ by computing the average $\Pr(x | \theta')$ over 50\% IN models and 50\% OUT models (using $2k$ models), i.e.:
\begin{align}
\label{eq:p_online}
    \begin{split}
        \Pr(x) \approx
        \frac{1}{2}\big(\underbrace{\frac{1}{k}\sum_{\theta'_{x}} \Pr(x | \theta'_{x})}_{\Pr(x)_{IN}} + \underbrace{\frac{1}{k}\sum_{\theta'_{\Bar{x}}} \Pr(x | \theta'_{\Bar{x}})}_{\Pr(x)_{OUT}}\big)
    \end{split}
\end{align}

However, in the offline setting, we do not have access to IN models. Thus, we need to exclusively use $\Pr(x | \theta'_{\Bar{x}})$. We now introduce a simple heuristic to obtain a less biased estimate of $\Pr(x)$ without having access to IN models, through an offline pre-computation to approximate the shift of probability between IN and OUT models. Essentially, we approximate the sensitivity of models (the gap between probability of member and non-member points in reference models). See Figure~1 in \cite{zhang2021counterfactual} for such computation.

We use our existing reference models to compute the rate at which $\Pr(x)$ for any sample $x$ changes between reference models that include $x$ versus the others. We approximate the gap with a linear function, so we obtain ${\Pr(x)_{IN} = a . \Pr(x)_{OUT} + b}$, and finally can obtain ${\Pr(x) = (\Pr(x)_{IN} + \Pr(x)_{OUT})/2}$. Given that both $\Pr(x)_{IN}$ and $\Pr(x)_{OUT}$ fall within the range of 0 to 1, it follows that ${a+b=1}$. Consequently, we can simplify the linear function as ${\Pr(x)_{IN} = a . (\Pr(x)_{OUT} - 1) + 1}$ which results in:
\begin{align}
\label{eq:p_in_approximation}
    \begin{split}
        \Pr(x) \approx
        \frac{1}{2}\big((1+a)\Pr(x)_{OUT} + (1-a)\big)
    \end{split}
\end{align}

In Figure~\ref{fig:auc_vs_offline_params}, we present the AUC obtained by our offline RMIA on CIFAR-10 models, varying the value of $a$ from 0 to 1. As $a$ approaches 1, there is a degradation in AUC (by up to 5.5\%), because we heavily rely on $\Pr(x)_{OUT}$ to approximate $\Pr(x)_{IN}$. Lower values of $a$ result in an enhancement of $\Pr(x)_{OUT}$ to approximate $\Pr(x)_{IN}$, leading to improved performance. The AUC appears to be more robust against lower values of $a$, particularly those below 0.5. Notably, even with $a=0$, where $\Pr(x) = (\Pr(x)_{OUT} + 1)/2$, a considerable improvement in results is observed. In this case, we are alleviating the influence of very low $\Pr(x)_{OUT}$ for atypical/hard samples. 

To determine the best value of $a$ for models trained with each dataset, we use the following procedure. It is executed only once, independent of test queries, without the need to train any new models. We choose two existing models and then, select one as the temporary target model and subject it to attacks from the other model using varying values of $a$. Finally, we select the one that yields the highest AUC as the optimal $a$. In the case of having only one reference model, we simulate an attack against the reference model and use the original target model as the reference model for the simulated attack to obtain the best $a$. Based on the result of our experiments, this optimal $a$ remains roughly consistent across random selections of reference models. In our experiments, we empirically derived the following values for our models: $a=0.3$ for CIFAR-10 and CINIC-10, $a=0.6$ for CIFAR-100, $a=1$ for ImageNet, and $a=0.2$ for Purchase-100 models.

\subsection{Direct Computation of Likelihood Ratio in \eqref{eq:likelihood_ratio}}
\label{app:direct_likelihood_computation}

In Section~\ref{sec:method}, we introduced two separate approaches for computing the fundamental likelihood ratio in~\eqref{eq:likelihood_ratio}:  I) a Bayesian method, as shown by~\eqref{eq:LR_computation}, and II) a direct method, expressed in~\eqref{eq:LR_expansion_direct}. While our empirical results are primarily derived using the Bayesian method, we anticipate that the outcomes of these two approaches will converge closely when a sufficient number of reference models is employed, although their performance may vary with a limited number of models. In this section, we attempt to assess these two methods and possibly invalidate our initial anticipation.

The direct approach involves employing Gaussian modeling over logits. Our attack is then simplified to the estimation of the mean and variance for the logits of samples $x$ and $z$ in two distinct distributions of reference models; One distribution comprises models trained with $x$ in their training set but not $z$ (denoted by $\theta'_{x,\Bar{z}}$), while the other comprises models trained with
$z$ in their training set but not $x$ (denoted by $\theta'_{\Bar{x},z}$). Let $\mu_{x,\Bar{z}}(x)$ and $\sigma_{x,\Bar{z}}(x)$ be the mean and variance of $f_{\theta'}(x)$ in the distribution of $\theta'_{x,\Bar{z}}$ models, respectively (a similar notation can be defined for $\theta'_{\Bar{x},z}$ models). We can approximate the likelihood ratio in~\eqref{eq:LR_expansion_direct} as follows:

\begin{align}
\label{eq:LR_expansion_direct_guassian}
    \LR & \approx \frac{\Pr(f_\theta(x), f_\theta(z) | x)}{\Pr(f_\theta(x), f_\theta(z) | z)}
    \approx \frac{\Pr(f_\theta(x) | \mathcal{N}(\mu_{x,\Bar{z}}(x),\,\sigma_{x,\Bar{z}}^{2}(x)))}{\Pr(f_\theta(x) | \mathcal{N}(\mu_{\Bar{x},z}(x),\,\sigma_{\Bar{x},z}^{2}(x)))} \times \frac{\Pr(f_\theta(z) | \mathcal{N}(\mu_{x,\Bar{z}}(z),\,\sigma_{x,\Bar{z}}^{2}(z)))}{\Pr(f_\theta(z) | \mathcal{N}(\mu_{\Bar{x},z}(z),\,\sigma_{\Bar{x},z}^{2}(z)))}
\end{align}
where $f_{\theta}(x)$ represents the output (logits) of the target model $\theta$ on sample $x$. Although the direct method may appear to be more straightforward and accurate, it comes with a significantly higher computational cost, since we must train online reference models, i.e. $\theta'_{x,\Bar{z}}$, to compute the probabilities in the above relation.

We now demonstrate the performance of our RMIA when the likelihood ratio is computed using~\eqref{eq:LR_computation} (hereafter, called RMIA-Bayes), in comparison to using the aforementioned likelihood ratio in~\eqref{eq:LR_expansion_direct_guassian} (referred to as RMIA-direct). Figure~\ref{fig:roc_32_combined} compares the ROCs achieved by the two methods when 64 reference models are employed to estimate the probabilities. We show the results obtained from models trained with different datasets (we use no augmented queries). In this case, RMIA-direct slightly outperforms in terms of both AUC and TPR at low FPRs, yet RMIA-Bayes closely matches its pace, even at low FPR values. On the other hand, Figure~\ref{fig:roc_2_combined} presents the scenario where 4 reference models are used. When utilizing fewer models, RMIA-Bayes exhibits a better performance across all datasets (e.g., 6.6\% higher AUC in CIFAR-10). It appears that RMIA-direct struggles to accurately estimate the parameters of Gaussian models with only four models available. Given the substantial processing cost of RMIA-direct, associated with training online models relative to sample pairs, RMIA-Bayes emerges as a more reasonable choice due to its ability to operate with a reduced number of models trained in an offline context.

\subsection{Analyzing $\gamma$ and $\beta$ Parameters and their Relation}
\label{app:relation_gamma_beta}

We here investigate the impact of selecting $\gamma$ on the efficacy of our attack. In Figure~\ref{fig:stats_vs_gamma}, we illustrate the sensitivity of our attack, measured in terms of AUC and FPR@TPR, to changes in the value of $\gamma$. We do not show the result for $\gamma < 1$, because it implies that a target sample $x$ is allowed to have a lower chance of being member than reference samples to pass our pairwise likelihood ratio test, causing lots of non-members to be wrongly inferred as member. As it can be seen from the figure, our attack's performance is consistent against changes in the value of $\gamma$; both AUC and FPR@TPR remain relatively stable with increasing $\gamma$ (except for a considerably high value of $\gamma$). As $\gamma$ increases, the need arises to decrease the value of $\beta$ in the hypothesis test equation (\eqref{eq:MIA_test}) to strike a balance between the power and error of the attack. By appropriately adjusting the value of $\beta$, we achieve a roughly same result across different $\gamma$ values. In extreme cases where a very high $\gamma$ is employed, the detection of $\gamma$-dominated reference samples between a limited set of $z$ records becomes exceedingly challenging. We consistently note the same trend in results across all our datasets and with varying numbers of reference models. Throughout our experiments, we set $\gamma$ to 2, as it yields a slightly higher TPR@FPR. 

\setvalue{\tmplabel}{stats_vs_gamma-}
\setcounter{figureNumber}{0}
{\tikzset{external/figure name/.add={}{\tmplabel}}
\begin{figure*}[h]
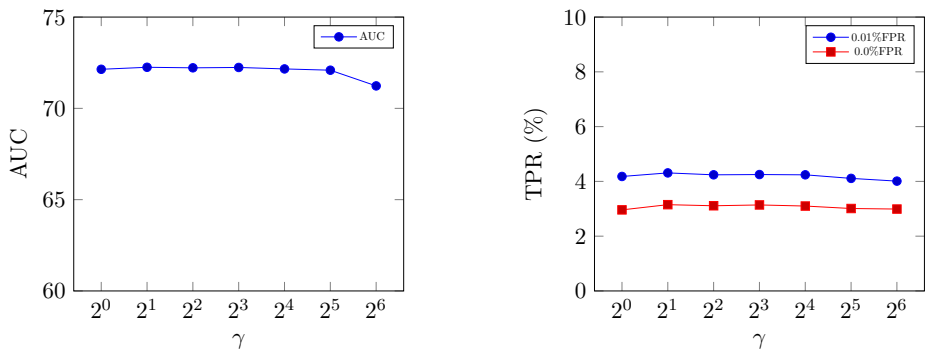

    \centering
    \if\compileFigures1
    \begin{tikzpicture}
    \centering
    \pgfplotstableread[col sep=comma]{data/stats_and_gammas.csv}\table
    \begin{axis}[
        legend pos=north east,
        xmode=log, log basis x={2},
        ymin=60,ymax=75,
        scale=0.8,
        ylabel={AUC},
        xlabel={$\gamma$},
        legend style={nodes={scale=0.5, transform shape}},
        ]
        \addplot table[x=x, y=0] \table;
        \addlegendentry {AUC}
    \end{axis}
\end{tikzpicture}\hspace*{2em}
\begin{tikzpicture}
    \centering
    \pgfplotstableread[col sep=comma]{data/stats_and_gammas.csv}\table
    \begin{axis}[
        legend pos=north east,
        xmode=log, log basis x={2},
        scale=0.8,
        ymin=0,ymax=10,
        ylabel={TPR (\%)},
        xlabel={$\gamma$},
        legend style={nodes={scale=0.5, transform shape}},
        ]
        \addplot table[x=x, y=4] \table;
        \addlegendentry {0.01\%FPR}

        \addplot table[x=x, y=6] \table;
        \addlegendentry {0.0\%FPR}
    \end{axis}
\end{tikzpicture}
    \else
    \scalebox{\SensiscaleFactor}{\includegraphics{fig/figure-\tmplabel\thefigureNumber.pdf}\stepcounter{figureNumber}}\hspace*{4em}
    \scalebox{\SensiscaleFactor}{\includegraphics{fig/figure-\tmplabel\thefigureNumber.pdf}\stepcounter{figureNumber}}
    \fi
    \caption{The performance sensitivity of our attack (RMIA) with respect to $\gamma$ parameter. Here, we use 254 reference models trained on CIFAR-10. The left plot shows the AUC obtained when using different $\gamma$ values, while the right plot demonstrates the TPR at 0.01\% and 0\% FPR values versus~$\gamma$ (corresponding to blue and red lines, respectively). }
    \label{fig:stats_vs_gamma}
\end{figure*}
}

As we discussed in Section~\ref{sec:method}, $\beta$ is a common threshold among all attacks, and will be used to generate the ROC curve. In our method (as opposed to e.g., \citet{Carlini2022Membership}), the exact value of $\beta$ is interpretable. Specifically, when $\gamma$ equals 1, the value of $\beta$ shows a strong correlation with the value of 1 - FPR. To show this, Figure~\ref{fig:gamma_beta} illustrates the impact of selecting $\beta$ from the range [0, 1] on the TPR and FPR of RMIA. We use two distinct $\gamma$ values: $\gamma=1$ (shown in the left plot) and $\gamma=2$ (depicted in the right plot). Both TPR and FPR approach zero, as we increase $\beta$ to 1 (under both $\gamma$ values) and the reason is that it is unlikely that a target sample can dominate most of reference records. However, FPR decreases more rapidly than TPR in two plots, especially with $\gamma=2$, which allows us to always have a higher power gain than error. When $\gamma=1$, FPR decreases in a calibrated manner, proportional to $1-\beta$. 

\setvalue{\tmplabel}{gamma_beta-}
\setcounter{figureNumber}{0}
{\tikzset{external/figure name/.add={}{\tmplabel}}
\begin{figure}[ht]
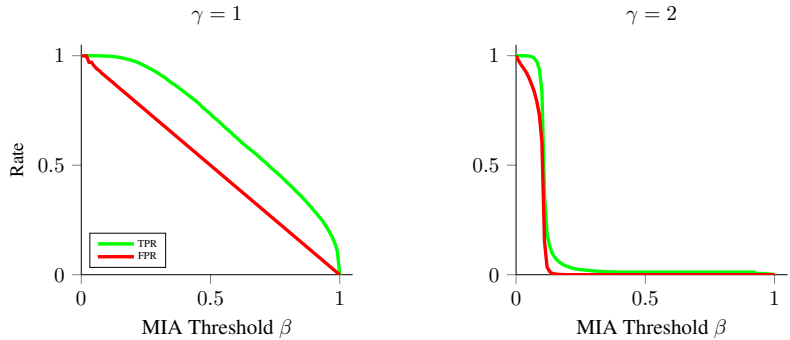

    \centering
    \if\compileFigures1
    \begin{tikzpicture}
        \pgfplotstableread[col sep=comma]{data/beta_fpr_tpr_1.0.csv}\data
        \begin{axis} [
            scale=0.4,width=0.75\linewidth,height=0.65\linewidth,
            ylabel={Rate},
            ymin=0,ymax=1.05,xmin=0,xmax=1.05,
            xlabel={MIA Threshold $\beta$},axis y line*=left,axis x line*=bottom,
            xtick align = outside,
            ytick align = outside,
            ylabel near ticks,ylabel style={align=center, text width=4cm, font=\small},
            legend pos=south west,
            legend columns=1,
            legend style={nodes={scale=0.5, transform shape}},
            title={$\gamma=1$}
            ]
            \addplot +[no marks, green, ultra thick] table[x=betas, y=tpr, col sep=comma] \data;
            \addlegendentry {TPR} %
            \addplot +[no marks, red, ultra thick] table[x=betas, y=fpr, col sep=comma] \data;
            \addlegendentry {FPR}
        \end{axis}
    \end{tikzpicture}\hspace*{4em}
    \begin{tikzpicture}
        \pgfplotstableread[col sep=comma]{data/beta_fpr_tpr_2.0.csv}\data
        \begin{axis} [
            scale=0.4,width=0.75\linewidth,height=0.65\linewidth,
            ymin=0,ymax=1.05,xmin=0,xmax=1.05,
            xlabel={MIA Threshold $\beta$},axis y line*=left,axis x line*=bottom,
            xtick align = outside,
            ytick align = outside,
            ylabel near ticks,ylabel style={align=center, text width=4cm, font=\small},
            legend pos=south east,
            legend columns=1,
            legend style={nodes={scale=0.5, transform shape}},
            title={$\gamma=2$}
            ]
            \addplot +[no marks, green, ultra thick] table[x=betas, y=tpr, col sep=comma] \data;
            \addplot +[no marks, red, ultra thick] table[x=betas, y=fpr, col sep=comma] \data;

        \end{axis}
    \end{tikzpicture}
    \else
    \scalebox{\SensiscaleFactor}{
    \includegraphics{fig/figure-\tmplabel\thefigureNumber.pdf}\stepcounter{figureNumber}}\hspace*{4em}
    \scalebox{\SensiscaleFactor}{\includegraphics{fig/figure-\tmplabel\thefigureNumber.pdf}\stepcounter{figureNumber}}
    \fi
    \caption{TPR and FPR achieved by RMIA for different values of $\beta$. The left and right plots correspond to $\gamma=1$ and $\gamma=2$, respectively. The number of reference models, trained with CIFAR-10, is 254.}
    \label{fig:gamma_beta}
\end{figure}
}

\subsection{Boosting RMIA with Augmented Queries}
\label{app:augmented_queries}
We can further enhance the effectiveness of attacks by augmenting the input query $x$ with multiple data samples that are similar to $x$~\cite{Carlini2022Membership, ChoquetteChoo2021Label}. These data samples can be simple transformations of $x$ (for example, using shift or rotation in case of image data). To consolidate the results in our multi-query setting, we use majority voting on our hypothesis test in \eqref{eq:likelihood_ratio}: $x$ is considered to dominate $z$ if more than half of all generated transformations of~$x$ dominate~$z$.

In \cite{Carlini2022Membership}, the authors discussed how to extend LiRA to support multiple queries. We here compare the performance of online and offline attacks when we increase the number of augmented queries from 1 to 50. Note that Attack-P and Attack-R have no result for multiple queries, as they do not originally support query augmentations. In this experiment, we use 254 reference models (for offline attacks, we only use 127 OUT models). 

We first compare the offline attacks, shown in the Offline column of Table~\ref{tab:all_queries_comparison}. With no augmented queries, RMIA presents a clear advantage over Attack-R (with 6.7\% higher AUC and 116\% more TPR at zero FPR) and the performance gap between two attacks widens with using more queries. As we increase queries, RMIA gets a better result (for example, a 4x improvement in TPR at zero FPR and also about 4.6\% higher AUC as queries go from 1 to 50), while LiRA cannot benefit from the advantage of more queries to improve its AUC. Note that we use the same technique to generate augmentations, as proposed in \cite{Carlini2022Membership}. 

In the Online column of Table~\ref{tab:all_queries_comparison}, we show the result of LiRA and RMIA when all 254 IN and OUT reference models are available to the adversary. Compared to LiRA, RMIA always has a slightly higher AUC and at least 48\% better TPR at zero FPR. Note that in this case, even minor AUC improvements are particularly significant when closely approaching the true leakage of the training algorithm through hundreds of models. Both LiRA and RMIA work better with increasing augmented queries, e.g. around $2\times$ improvement in TPR@FPR when going from 1 query to 50 queries. Unless explicitly stated otherwise, the experimental results in this paper are obtained using 18 augmented queries for both LiRA and RMIA.

\begin{table*}[t]
    \centering
        \pgfplotstableread[col sep=comma]{data/cifar10_augmentation_comparison_new.csv}\tab
        \pgfplotstableread[col sep=comma]{data/cifar10_augmentation_comparison_offline_new.csv}\offlinetab

                \resizebox{0.8\textwidth}{!}{
              \begin{tabular}{lllllllll}
                \toprule
                \multirow{3}{*}{\# Queries} &
                  \multirow{3}{*}{Attack} &
                  \multicolumn{3}{c}{Online} &
                   &
                  \multicolumn{3}{c}{Offline} \\ \cmidrule{3-5} \cmidrule{7-9}
                 &
                   &
                  \multicolumn{1}{c}{\multirow{2}{*}{AUC}} &
                  \multicolumn{2}{c}{\TPRatFPR} &
                   &
                  \multicolumn{1}{c}{\multirow{2}{*}{AUC}} &
                  \multicolumn{2}{c}{\TPRatFPR} \\ \cmidrule{4-5} \cmidrule{8-9}
                 &
                   &
                  \multicolumn{1}{c}{} &
                  \hundredth &
                  \zero &
                   &
                  \multicolumn{1}{c}{} &
                  \hundredth &
                  \zero \\ \midrule
                \multirow{4}{*}{1} &
                  \reference &
                  \multicolumn{1}{c}{-} &
                  \multicolumn{1}{c}{-} &
                  \multicolumn{1}{c}{-} &
                  &
                  \pgfplotstablegetelem{0}{{reference}}\of\tab\pgfplotsretval$\,\pm\,$\pgfplotstablegetelem{1}{{reference}}\of\tab\pgfplotsretval &
                  \pgfplotstablegetelem{4}{{reference}}\of\tab\pgfplotsretval$\,\pm\,$\pgfplotstablegetelem{5}{{reference}}\of\tab\pgfplotsretval &
                  \pgfplotstablegetelem{6}{{reference}}\of\tab\pgfplotsretval$\,\pm\,$\pgfplotstablegetelem{7}{{reference}}\of\tab\pgfplotsretval \\
\bottomrule
                \end{tabular}
                }
    \caption{Performance of attacks when we use different number of augmented queries for \lirans~\citep{Carlini2022Membership} and \relativens. We evaluate attacks in two different settings, shown in Online and Offline columns. In the online setting, we use 254 models where half of them are IN models and half are OUT (for each sample). The offline setting uses only 127 OUT models. Both \population and \referencens~\citep{Ye2022Enhanced} do not originally support multiple augmented queries. The \population does not work with reference models, thus we consider it offline. Models are trained with CIFAR-10. Results are averaged over 10 random target models.}
    \label{tab:all_queries_comparison}
\end{table*}

\clearpage

\section{Supplementary Empirical Results}
\subsection{Robustness of Attacks against Noise and OOD Non-member Queries} 
\label{app:robustness_ood}
To evaluate the performance of attacks against OOD non-member samples, we consider two strategies for generating test queries: incorporating samples from a dataset different from the training dataset and using pure noise. Figure~\ref{fig:ood_noise_roc} illustrates the ROC curves of three offline attacks using models trained on CIFAR-10 (all member queries are from in-distribution). We use 127 (OUT) reference models. The ROCs are presented in normal scale to highlight the gap between attacks. To generate OOD samples, we employ samples from CINIC-10 with the same label as CIFAR-10. When using OOD samples as test queries (depicted in the left plot), we observe a substantial performance gap (at least 21\% higher AUC) between RMIA and other attacks, with LiRA not performing much better than random guessing. In the case of using noise as non-member test queries (depicted in the right plot), RMIA once again outperforms the other two two attacks by at least 61\% in terms of AUC, while LiRA falls significantly below the random guess (i.e., it has a very large false positive rate).

\setvalue{\tmplabel}{rocs_ood_cifar10-}
\setcounter{figureNumber}{0}
{\tikzset{external/figure name/.add={}{\tmplabel}}
\begin{figure}[h]
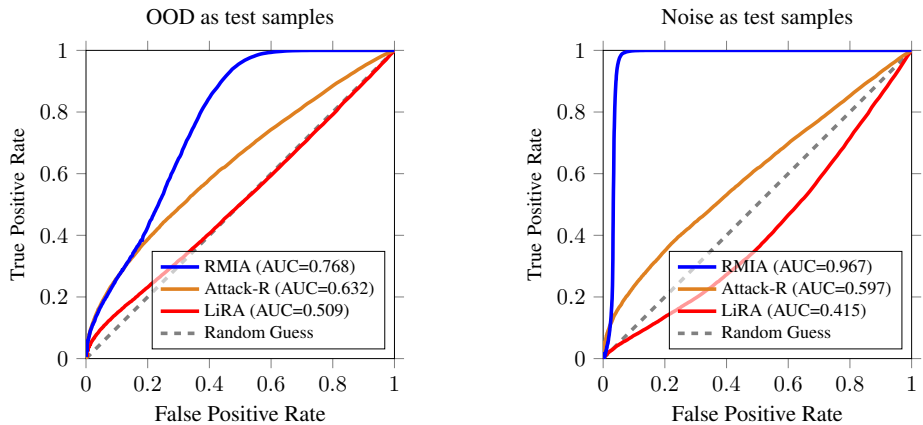

    \centering
    \if\compileFigures1
\scalebox{\ROCscaleFactor}{
    \begin{tikzpicture}
        \pgfplotstableread[col sep=comma]{data/rocs/rocs_ood_roc.csv}\roc
        \pgfplotstableread[col sep=comma]{data/rocs/rocs_ood_auc.csv}\auc
        \begin{axis} [
            title=OOD as test samples,
            scale=0.31,
            width=1.0\linewidth,height=1.0\linewidth,
            xlabel={False Positive Rate},
            ylabel={True Positive Rate},
            ymin=10e-6,ymax=1,xmin=10e-6,xmax=1, 
            xtick align = outside,
            ytick align = outside,
            ylabel near ticks,ylabel style={align=center, text width=4cm, font=\small},
            legend pos=south east,
            legend columns=1,
            legend style={nodes={scale=0.6, transform shape}, fill opacity=0.6, draw opacity=1,text opacity=1},
            legend cell align={left},
            every axis plot/.append style={\ROCthickness},
            reverse legend=true,
            ]
            \addplot [\colorRandom,dashed] table[x=random,y=random, col sep=comma] \roc;
            \addlegendentry {\random}

            \addplot [\colorOnline,no marks,\styleAugOthers] table[x=random,y={LiRA}, col sep=comma] \roc;
            \addlegendentry {\lira (\AUC=\pgfplotstablegetelem{0}{LiRA}\of\auc\pgfplotsretval)}


            \addplot [\colorReference,no marks] table[x=random,y={Attack-R}, col sep=comma] \roc;
            \addlegendentry {\reference (\AUC=\pgfplotstablegetelem{0}{{Attack-R}}\of\auc\pgfplotsretval)}

            \addplot [\colorRelative,no marks,\styleAugOurs] table[x=random,y={RMIA}, col sep=comma] \roc;
            \addlegendentry {\relative (\AUC=\pgfplotstablegetelem{0}{{RMIA}}\of\auc\pgfplotsretval)}
        \end{axis}
    \end{tikzpicture}\hspace*{4em}
    \begin{tikzpicture}
        \pgfplotstableread[col sep=comma]{data/rocs/rocs_noise_roc.csv}\roc
        \pgfplotstableread[col sep=comma]{data/rocs/rocs_noise_auc.csv}\auc
        \begin{axis} [
            title=Noise as test samples,
            scale=0.31,
            width=1.0\linewidth,height=1.0\linewidth,
            xlabel={False Positive Rate},
            ylabel={True Positive Rate},
            ymin=10e-6,ymax=1,xmin=10e-6,xmax=1, 
            xtick align = outside,
            ytick align = outside,
            ylabel near ticks,ylabel style={align=center, text width=4cm, font=\small},
            legend pos=south east,
            legend columns=1,
            legend style={nodes={scale=0.6, transform shape}, fill opacity=0.6, draw opacity=1,text opacity=1},
            legend cell align={left},
            every axis plot/.append style={\ROCthickness},
            reverse legend=true,
            ]
            \addplot [\colorRandom,dashed] table[x=random,y=random, col sep=comma] \roc;
            \addlegendentry {\random}

            \addplot [\colorOnline,no marks,\styleAugOthers] table[x=random,y={LiRA}, col sep=comma] \roc;
            \addlegendentry {\lira (\AUC=\pgfplotstablegetelem{0}{LiRA}\of\auc\pgfplotsretval)}


            \addplot [\colorReference,no marks] table[x=random,y={Attack-R}, col sep=comma] \roc;
            \addlegendentry {\reference (\AUC=\pgfplotstablegetelem{0}{{Attack-R}}\of\auc\pgfplotsretval)}

            \addplot [\colorRelative,no marks,\styleAugOurs] table[x=random,y={RMIA}, col sep=comma] \roc;
            \addlegendentry {\relative (\AUC=\pgfplotstablegetelem{0}{{RMIA}}\of\auc\pgfplotsretval)}

        \end{axis}
    \end{tikzpicture}
}
    \else
    \scalebox{\AUCMainscaleFactor}{
    \includegraphics{fig/figure-\tmplabel\thefigureNumber.pdf}\stepcounter{figureNumber}}\hspace*{4em}
    \scalebox{\AUCMainscaleFactor}{\includegraphics{fig/figure-\tmplabel\thefigureNumber.pdf}\stepcounter{figureNumber}}
    \fi
    \caption{ROC of three offline attacks using models trained on CIFAR-10 when non-member test queries come from a  distribution different from the population data $\pi$. The result is obtained on one random target model. The left plot uses out-of-distribution (OOD) samples from CINIC-10 (with the same label as CIFAR-10) as non-member test queries, while the right plot uses random noise. We here use 127 reference models (OUT). For RMIA, we use $a=0$ and $\gamma=2$.}
    \label{fig:ood_noise_roc}
\end{figure}

Figure~\ref{fig:ood_ratio} illustrates the performance of various attacks when using different ratios of OOD non-member samples as test queries (clearly, all member queries are in-distribution). When the ratio is zero, only in-distribution non-member queries are used, and when it is one, all non-member test queries are OOD. As depicted in the left plot, increasing the ratio of OOD samples results in an increase in the AUC of RMIA and Attack-P, whereas the performance of the other two attacks decreases. This is primarily because the former two attacks are capable of filtering OOD samples by comparing the MIA score of the test query with in-distribution population data. When comparing attacks based on their TPR at zero FPR (the left plot), the population attack yields nearly zero TPR, while RMIA outperforms all other attacks in this regard. We observe a decreasing trend for TPR at zero FPR in all attacks, indicating the presence of OOD cases that are difficult to detect via comparison of their MIA score with population data.

\setvalue{\tmplabel}{ood_ratio-}
\setcounter{figureNumber}{0}
{\tikzset{external/figure name/.add={}{\tmplabel}}
\begin{figure*}[h]
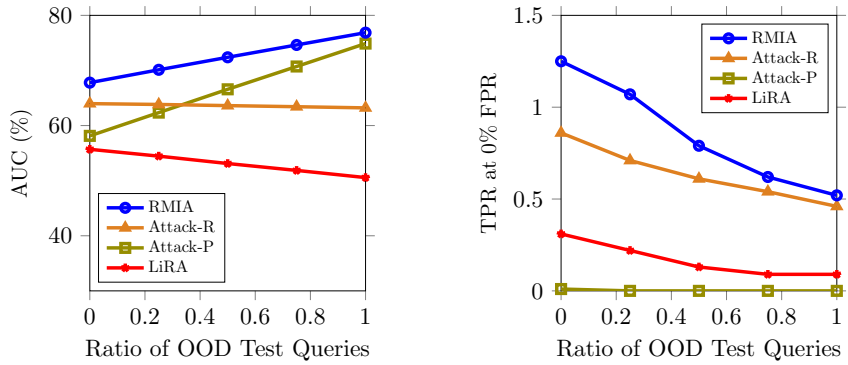

    \centering
    \if\compileFigures1
    \scalebox{\ROCscaleFactor}{
    \begin{tikzpicture}
        \centering
        \pgfplotstableread[col sep=comma]{data/ood_ratio_auc.csv}\table
        \begin{axis}[
            scale=0.31,
            width=1.0\linewidth,height=1.0\linewidth,
            ylabel={AUC (\%)},
            xlabel={Ratio of OOD Test Queries},
            ymin=30,ymax=80,xmin=0,xmax=1, 
            xtick align = outside,
            ytick align = outside,
            ylabel near ticks,ylabel style={align=center, text width=4cm, font=\small},
            legend pos=south west,
            legend columns=1,
            legend style={nodes={scale=0.7, transform shape}, fill opacity=0.6, draw opacity=1,text opacity=1},
            legend cell align={left},
            every axis plot/.append style={\ROCthickness},
            reverse legend=true,
            ]
            \addplot [color=\colorOnline, mark=star] table[x=ratio, y=lira] \table;
            \addlegendentry {LiRA}
            \addplot [color=\colorPopulation, mark=square] table[x=ratio, y=population] \table;
            \addlegendentry {Attack-P}
            \addplot [color=\colorReference, mark=triangle] table[x=ratio, y=reference] \table;
            \addlegendentry {Attack-R}
            \addplot [color=\colorRelative, mark=o] table[x=ratio, y=rmia] \table;
            \addlegendentry {RMIA}

        \end{axis}
    \end{tikzpicture}
    \begin{tikzpicture}
        \centering
        \pgfplotstableread[col sep=comma]{data/ood_ratio_fpr_0.csv}\table
        \begin{axis}[
            scale=0.31,
            width=1.0\linewidth,height=1.0\linewidth,
            ylabel={TPR at 0\% FPR},
            xlabel={Ratio of OOD Test Queries},
            ymin=0,ymax=1.5,xmin=0,xmax=1, 
            xtick align = outside,
            ytick align = outside,
            ylabel near ticks,ylabel style={align=center, text width=4cm, font=\small},
            legend pos=north east,
            legend columns=1,
            legend style={nodes={scale=0.7, transform shape}, fill opacity=0.6, draw opacity=1,text opacity=1},
            legend cell align={left},
            every axis plot/.append style={\ROCthickness},
            reverse legend=true,
            ]
            \addplot [color=\colorOnline, mark=star] table[x=ratio, y=lira] \table;
            \addlegendentry {LiRA}
            \addplot [color=\colorPopulation, mark=square] table[x=ratio, y=population] \table;
            \addlegendentry {Attack-P}
            \addplot [color=\colorReference, mark=triangle] table[x=ratio, y=reference] \table;
            \addlegendentry {Attack-R}
            \addplot [color=\colorRelative, mark=o] table[x=ratio, y=rmia] \table;
            \addlegendentry {RMIA}

        \end{axis}
    \end{tikzpicture}
    }
    \else
    \scalebox{\ROCscaleFactor}{
    \includegraphics{fig/figure-\tmplabel\thefigureNumber.pdf}\stepcounter{figureNumber}\hspace*{4em}
    \includegraphics{fig/figure-\tmplabel\thefigureNumber.pdf}\stepcounter{figureNumber}}
    \fi
    \caption{Performance of attacks, when testing against different ratios of OOD non-member queries (if ratio=0, we use no OOD sample as test query, i.e. all non-members are in-distribution, while for ratio=1, all non-member queries are OOD). The left plot shows the AUC obtained by different attacks, and the right plot depicts the TPR at zero FPR. All attacks are offline and the number of reference models is 127. All models are trained with CIFAR-10 and OOD queries are from CINIC-10.}
    \label{fig:ood_ratio}
\end{figure*}
}

To gain a deeper insight into RMIA's response to OOD MIA queries, Figure~\ref{fig:rmia_id_ood_fpr_comparison} presents a comparison of the FPRs when using two distinct sets of non-members: out-of-distribution (OOD) versus in-distribution (ID) samples. The left plot illustrates the TPR and FPR achieved for various MIA thresholds~$\beta$, separately for ID and OOD samples. Notably, as~$\beta$ increases from zero, the FPR for OOD samples diminishes significantly faster in contrast to the FPR and TPR for ID samples. This observation underscores RMIA's efficacy in discerning a majority of OOD queries as non-members. Furthermore, as illustrated in the right plot, when comparing the FPRs obtained from OOD and ID non-member queries, a striking discrepancy emerges. While the FPR for ID samples can exceed 0.8, the FPR for OOD samples (at the same $\beta$ value) is almost zero. This discrepancy implies that RMIA finds it easier to detect OOD non-members compared to ID non-members.

\setvalue{\tmplabel}{rmia_id_ood_fpr-}
\setcounter{figureNumber}{0}
{\tikzset{external/figure name/.add={}{\tmplabel}}
\begin{figure*}[h]
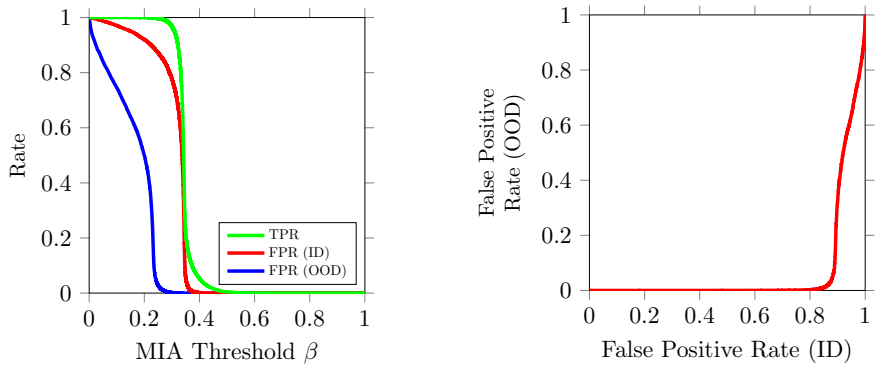

    \centering
    \if\compileFigures1
    \scalebox{\ROCscaleFactor}{
    \begin{tikzpicture}
        \centering
        \pgfplotstableread[col sep=comma]{data/non_ood.csv}\table
        \begin{axis}[
            scale=0.31,
            width=1.0\linewidth,height=1.0\linewidth,
            ylabel={Rate},
            xlabel={MIA Threshold $\beta$},
            ymin=10e-6,ymax=1,xmin=10e-6,xmax=1, 
            xtick align = outside,
            ytick align = outside,
            ylabel near ticks,ylabel style={align=center, text width=4cm, font=\small},
            legend pos=south east,
            legend columns=1,
            legend style={nodes={scale=0.6, transform shape}, fill opacity=0.6, draw opacity=1,text opacity=1},
            legend cell align={left},
            every axis plot/.append style={\ROCthickness},
            reverse legend=true,
            ]
            \addplot [color=blue, no marks] table[x=threshold, y={fpr_ood}] \table;
            \addlegendentry {FPR (OOD)}
            \addplot [color=red, no marks] table[x=threshold, y={fpr_id}] \table;
            \addlegendentry {FPR (ID)}
            \addplot [color=green, no marks] table[x=threshold, y={tpr}] \table;
            \addlegendentry {TPR}
        \end{axis}
    \end{tikzpicture}
    \begin{tikzpicture}
        \centering
        \pgfplotstableread[col sep=comma]{data/non_ood.csv}\table
        \begin{axis}[
            scale=0.31,
            width=1.0\linewidth,height=1.0\linewidth,
            ylabel={False Positive Rate (OOD)},
            xlabel={False Positive Rate (ID)},
            ymin=10e-6,ymax=1,xmin=10e-6,xmax=1, 
            xtick align = outside,
            ytick align = outside,
            ylabel near ticks,ylabel style={align=center, text width=4cm, font=\small},
            legend pos=north west,
            legend columns=1,
            legend style={nodes={scale=0.6, transform shape}, fill opacity=0.6, draw opacity=1,text opacity=1},
            legend cell align={left},
            every axis plot/.append style={\ROCthickness},
            reverse legend=true,
            ]
            \addplot [color=red, no marks] table[x={fpr_id}, y={fpr_ood}] \table;
        \end{axis}
    \end{tikzpicture}
    }
    \else
    \scalebox{\ROCscaleFactor}{
    \includegraphics{fig/figure-\tmplabel\thefigureNumber.pdf}\stepcounter{figureNumber}\hspace*{4em}
    \includegraphics{fig/figure-\tmplabel\thefigureNumber.pdf}\stepcounter{figureNumber}}
    \fi
    \caption{Comparing FPR of RMIA (Offline) when two separate sets of non-member test queries, i.e. in-distribution (ID) and out-of-distribution (OOD) samples, are used. The number of reference models is 127 and $\gamma=2$. All models are trained with CIFAR-10 and OOD queries are from CINIC-10. The left plot shows TPR and FPR obtained for each MIA threshold $\beta$, while the right plot compares obtained FPRs when OOD vs ID samples are used as non-member test queries.}
    \label{fig:rmia_id_ood_fpr_comparison}
\end{figure*}
}

\subsection{Data Distribution Shift} 
\label{app:data_shift}
Table~\ref{tab:cifar_cinic_models_2_4_new} compares the result of attacks when the target models are trained on different datasets than the reference models. More specifically, the target models are trained on CIFAR-10, while the reference models are trained on CINIC-10 (all test queries are from CIFAR-10). We use images with common class labels between two datasets. We here concentrate on offline attacks, as the reference models are trained on a completely different dataset than the target model. We report the results obtained with different number of reference models (1, 2 and 4). The shift in distribution of training data between the target model and the reference models affects the performance of all attacks. Compared with other two attacks, RMIA always obtains a higher AUC (e.g. by up to 25\% in comparison with LiRA) and a better TPR at low FPRs.
\begin{table}[h]
    \centering
    \pgfplotstableread[col sep=comma]{data/cifar_target_cinic_reference_comparison_low_num_ref_aug_18.csv}\tab
        \resizebox{0.65\textwidth}{!}{
          \begin{tabular}{lllll}
            \toprule
            \multirow{2}{*}{\# Ref Models} &
              \multirow{2}{*}{Attack} &
              \multicolumn{1}{c}{\multirow{2}{*}{AUC}} &
              \multicolumn{2}{c}{\TPRatFPR} \\ \cmidrule{4-5}
             &
               &
              \multicolumn{1}{c}{} &
              \hundredth &
              \zero \\ \midrule
            \multirow{3}{*}{1} &
              \referenceWithAuthors &
              \pgfplotstablegetelem{0}{{reference 2}}\of\tab\pgfplotsretval$\,\pm\,$\pgfplotstablegetelem{1}{{reference 2}}\of\tab\pgfplotsretval &
              \pgfplotstablegetelem{4}{{reference 2}}\of\tab\pgfplotsretval$\,\pm\,$\pgfplotstablegetelem{5}{{reference 2}}\of\tab\pgfplotsretval &
              \pgfplotstablegetelem{6}{{reference 2}}\of\tab\pgfplotsretval$\,\pm\,$\pgfplotstablegetelem{7}{{reference 2}}\of\tab\pgfplotsretval \\
             &
              \liraOfflineWithAuthors &
              \pgfplotstablegetelem{0}{{liraoffline 2}}\of\tab\pgfplotsretval$\,\pm\,$\pgfplotstablegetelem{1}{{liraoffline 2}}\of\tab\pgfplotsretval &
              \pgfplotstablegetelem{4}{{liraoffline 2}}\of\tab\pgfplotsretval$\,\pm\,$\pgfplotstablegetelem{5}{{liraoffline 2}}\of\tab\pgfplotsretval &
              \pgfplotstablegetelem{6}{{liraoffline 2}}\of\tab\pgfplotsretval$\,\pm\,$\pgfplotstablegetelem{7}{{liraoffline 2}}\of\tab\pgfplotsretval \\
\bottomrule
            \end{tabular}
            }
    \caption{Performance of offline attacks when different datasets are used for training target and reference models. Specifically, the target models are trained on \textbf{CIFAR-10}, while the reference models are trained on \textbf{CINIC-10}. We use different numbers of reference models (1, 2 and 4 OUT models). Results are averaged over 10 random target models. }
    \label{tab:cifar_cinic_models_2_4_new}
\end{table}

\subsection{Variations in Neural Network Architectures}
\label{app:architecture_variation}

Figure~\ref{fig:roc_multi_architectures} illustrates the performance of attacks when models are trained with different architectures, including CNN and Wide ResNet (WRN) of various sizes, on CIFAR-10. In this scenario, both target and reference models share the same architecture. RMIA consistently outperforms other attacks across all architectures (e.g. 7.5\%-16.8\% higher AUC compared with LiRA). Using network architectures with more parameters can lead to increased leakage, as shown by \citet{Carlini2019Secret}.

\setvalue{\tmplabel}{roc_multi_architectures-}
\setcounter{figureNumber}{0}
{\tikzset{external/figure name/.add={}{\tmplabel}}
\begin{figure*}[h!]
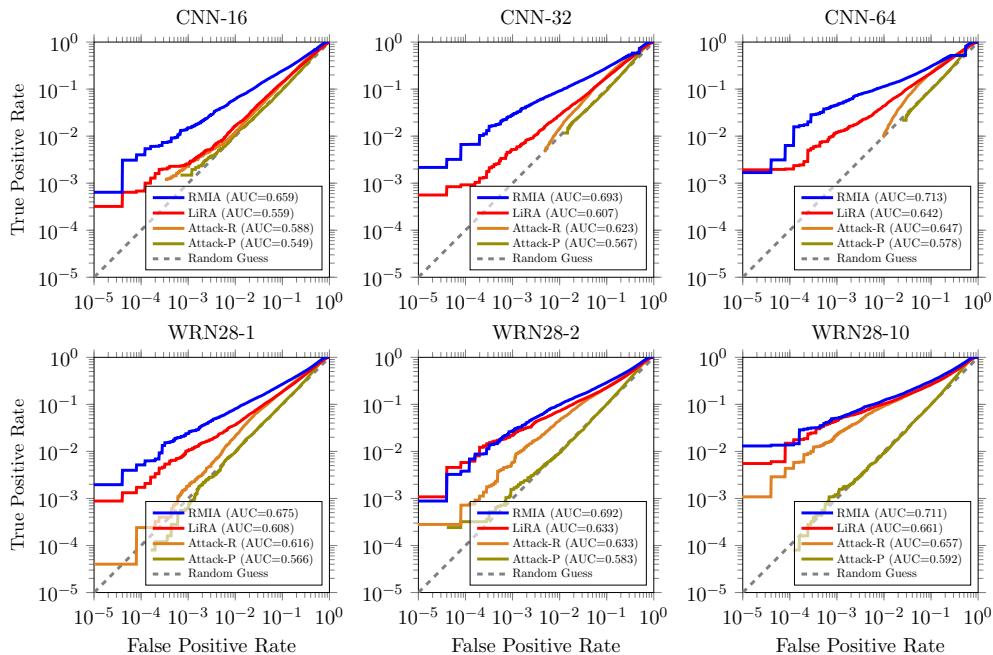

    \centering
    \if\compileFigures1
    \input{figure_scripts/multi_architecture.tex}
    \else
    \scalebox{\ROCMultiscaleFactor}{
    \includegraphics{fig/figure-\tmplabel\thefigureNumber.pdf}\stepcounter{figureNumber}
    \includegraphics{fig/figure-\tmplabel\thefigureNumber.pdf}\stepcounter{figureNumber}
    \includegraphics{fig/figure-\tmplabel\thefigureNumber.pdf}\stepcounter{figureNumber}
    } \\
    \scalebox{\ROCMultiscaleFactor}{
    \includegraphics{fig/figure-\tmplabel\thefigureNumber.pdf}\stepcounter{figureNumber}
    \includegraphics{fig/figure-\tmplabel\thefigureNumber.pdf}\stepcounter{figureNumber}
    \includegraphics{fig/figure-\tmplabel\thefigureNumber.pdf}\stepcounter{figureNumber}
    }
    \fi
    \caption{ROC of attacks using different neural network architectures for training models on CIFAR-10. Here, both target and reference models share the same architecture. We use 2 reference models (1 IN, 1 OUT ). }
    \label{fig:roc_multi_architectures}
\end{figure*}
}

Figure~\ref{fig:roc_multi_architectures_same_target} presents the performance of attacks when different architectures are used to train reference models, while keeping the structure of the target model fixed as WRN28-2. So, the target and reference models may have different architectures. The optimal performance for attacks is observed when both target and reference models share similar architectures. However, RMIA outperforms other attacks again, and notably, the performance gap widens under architecture shifts.

\setvalue{\tmplabel}{roc_multi_architectures_same_target-}
\setcounter{figureNumber}{0}
{\tikzset{external/figure name/.add={}{\tmplabel}}
\begin{figure*}[h!]
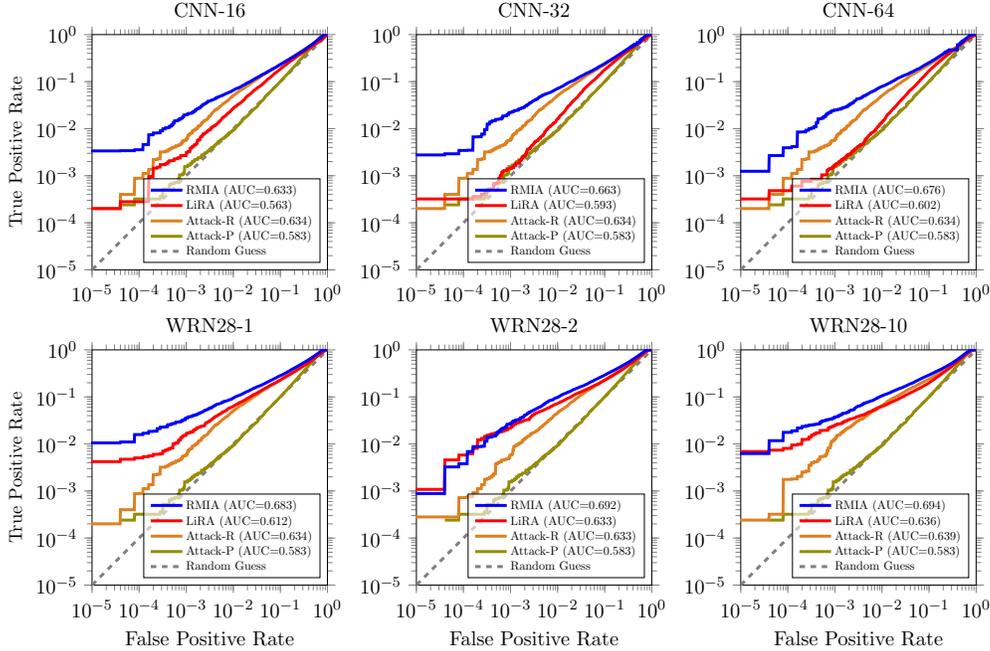

    \centering
    \if\compileFigures1
    \input{figure_scripts/multi_architecture_same_target.tex}
    \else
    \scalebox{\ROCMultiscaleFactor}{
    \includegraphics{fig/figure-\tmplabel\thefigureNumber.pdf}\stepcounter{figureNumber}
    \includegraphics{fig/figure-\tmplabel\thefigureNumber.pdf}\stepcounter{figureNumber}
    \includegraphics{fig/figure-\tmplabel\thefigureNumber.pdf}\stepcounter{figureNumber}
    } \\
    \scalebox{\ROCMultiscaleFactor}{
    \includegraphics{fig/figure-\tmplabel\thefigureNumber.pdf}\stepcounter{figureNumber}
    \includegraphics{fig/figure-\tmplabel\thefigureNumber.pdf}\stepcounter{figureNumber}
    \includegraphics{fig/figure-\tmplabel\thefigureNumber.pdf}\stepcounter{figureNumber}
    }
    \fi
    \caption{ROC of attacks using different neural network architectures for training reference models on CIFAR-10. The target model is always trained using WRN28-2. We here use 2 reference models (1 IN, 1 OUT). }
    \label{fig:roc_multi_architectures_same_target}
\end{figure*}
}

\subsection{Performance of Attacks on Models Trained with a Larger Subset of CINIC-10}
\label{app:larger_datasets}

Analyzing the effectiveness of attacks against models with larger training set is crucial, as they might experience performance deterioration, as highlighted by Bertran et al. (2023). We here evaluate attacks when tested against models trained with an expanded training set from CINIC-10. Specifically, we use a random subset consisting of 90k samples which is larger than the size of training sets in our previous experiments (i.e. 25k). We train a Wide ResNet network for 200 epochs on half of the randomly chosen dataset (with batch size, learning rate, and weight decay set to 64, 0.1, and 0.0005, respectively). The test accuracy of CINIC-10 models is 79\%. For our offline attack, we use the original Softmax with temperature of 1 as the confidence function. Moreover,the best value of $a$ in \eqref{eq:p_in_approximation} is obtained 1. Figure~\ref{fig:roc_larger_cinic10_dataset_1_model} illustrates the ROC curves of various attacks conducted on CINIC-10 when using only 1 (OUT) reference model. Additionally, it presents the outcome of the Quantile Regression attack by Bertran et al. (2023), which trains regression-based attack models instead of reference models. Our attack demonstrates superior performance compared to other attacks in terms of both AUC and TPR at low FPRs. For instance, it achieves approximately 15\% higher AUC than LiRA and an order of magnitude higher TPR at 0.01\% FPR than the Quantile Regression attack. Although the AUC obtained by the Quantile Regression attack is notably higher than that of LiRA (with only 1 reference model), it was unable to achieve any true positives at zero FPR.

\setvalue{\tmplabel}{roc_1_model_quantile-}
\setcounter{figureNumber}{0}
{\tikzset{external/figure name/.add={}{\tmplabel}}
\begin{figure*}[!t]
    \centering
    \if\compileFigures1
    \scalebox{\ROCscaleFactor}{
    \begin{tikzpicture}
        \pgfplotstableread[col sep=comma]{data/rocs/q_cinic10_rocs_roc.csv}\roc
        \pgfplotstableread[col sep=comma]{data/rocs/q_cinic10_rocs_auc.csv}\auc
        \begin{axis} [
            scale=0.31,
            width=1.0\linewidth,height=1.0\linewidth,
            ylabel={True Positive Rate},
            ymin=10e-6,ymax=1,xmin=10e-6,xmax=1, 
            xmode=log,ymode=log,
            xtick align = outside,
            ytick align = outside,
            ylabel near ticks,ylabel style={align=center, text width=4cm, font=\small},
            legend pos=south east,
            legend columns=1,
            legend style={nodes={scale=0.6, transform shape}, fill opacity=0.6, draw opacity=1,text opacity=1},
            legend cell align={left},
            every axis plot/.append style={\ROCthickness},
            reverse legend=true,
            ]
            \addplot [\colorRandom,dashed] table[x=random,y=random, col sep=comma] \roc;
            \addlegendentry {\random}

            \addplot [\colorPopulation,no marks] table[x=random,y={Attack-P}, col sep=comma] \roc;
            \addlegendentry {\population (\AUC=\pgfplotstablegetelem{0}{{Attack-P}}\of\auc\pgfplotsretval)}

            \addplot [pink,no marks,\styleAugOurs] table[x=q-fpr,y={q-tpr}, col sep=comma] \roc;
            \addlegendentry {Quantile-Reg. \AUC=\pgfplotstablegetelem{0}{{Quantile}}\of\auc\pgfplotsretval)}

            \addplot [\colorReference,no marks] table[x=random,y={Attack-R}, col sep=comma] \roc;
            \addlegendentry {\reference (\AUC=\pgfplotstablegetelem{0}{{Attack-R}}\of\auc\pgfplotsretval)}

            \addplot [\colorOnline,no marks,\styleAugOthers] table[x=random,y={LiRA}, col sep=comma] \roc;
            \addlegendentry {\lira (\AUC=\pgfplotstablegetelem{0}{LiRA}\of\auc\pgfplotsretval)}

            \addplot [\colorRelative,no marks,\styleAugOurs] table[x=random,y={RMIA}, col sep=comma] \roc;
            \addlegendentry {\relative (\AUC=\pgfplotstablegetelem{0}{{RMIA}}\of\auc\pgfplotsretval)}


        \end{axis}
    \end{tikzpicture}\hspace*{2em}
    \begin{tikzpicture}
        \pgfplotstableread[col sep=comma]{data/rocs/cinic10_num_ref_comparison_4_report_18_aug_roc.csv}\roc
        \pgfplotstableread[col sep=comma]{data/rocs/cinic10_num_ref_comparison_4_report_18_aug_auc.csv}\auc
        \begin{axis} [
            scale=0.31,
            width=1.0\linewidth,height=1.0\linewidth,
            ymin=10e-6,ymax=1,xmin=10e-6,xmax=1, 
            xtick align = outside,
            ytick align = outside,
            ylabel near ticks,ylabel style={align=center, text width=4cm, font=\small},
            legend pos=south east,
            legend columns=1,
            legend style={nodes={scale=0.6, transform shape}, fill opacity=0.6, draw opacity=1,text opacity=1},
            legend cell align={left},
            every axis plot/.append style={\ROCthickness},
            reverse legend=true,
            ]
            \addplot [\colorRandom,dashed] table[x=random,y=random, col sep=comma] \roc;
            \addlegendentry {\random}

            \addplot [\colorPopulation,no marks] table[x=random,y={Attack-P}, col sep=comma] \roc;
            \addlegendentry {\population (\AUC=\pgfplotstablegetelem{0}{{Attack-P}}\of\auc\pgfplotsretval)}

            \addplot [pink,no marks,\styleAugOurs] table[x=q-fpr,y={q-tpr}, col sep=comma] \roc;
            \addlegendentry {Quantile-Reg. \AUC=\pgfplotstablegetelem{0}{{Quantile}}\of\auc\pgfplotsretval)}

            \addplot [\colorReference,no marks] table[x=random,y={Attack-R}, col sep=comma] \roc;
            \addlegendentry {\reference (\AUC=\pgfplotstablegetelem{0}{{Attack-R}}\of\auc\pgfplotsretval)}

            \addplot [\colorOnline,no marks,\styleAugOthers] table[x=random,y={LiRA}, col sep=comma] \roc;
            \addlegendentry {\lira (\AUC=\pgfplotstablegetelem{0}{LiRA}\of\auc\pgfplotsretval)}

            \addplot [\colorRelative,no marks,\styleAugOurs] table[x=random,y={RMIA}, col sep=comma] \roc;
            \addlegendentry {\relative (\AUC=\pgfplotstablegetelem{0}{{RMIA}}\of\auc\pgfplotsretval)}

        \end{axis}
    \end{tikzpicture}
    }
    \else
    \includegraphics{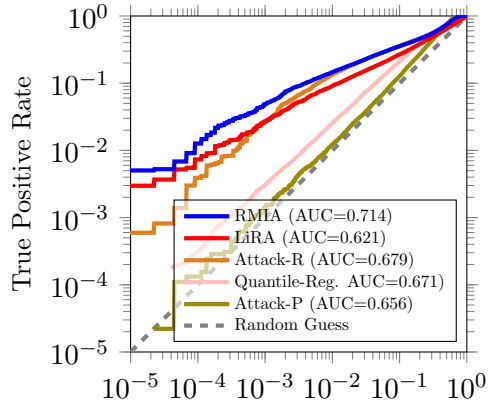}\stepcounter{figureNumber}
    \fi
    \caption{ROC of attacks using models trained with a larger subset of CINIC-10 (i.e. 90k samples). The result is obtained on one random target model. We here use \textbf{1 reference model} (OUT). We do not use augmented queries for any attacks. }
    \label{fig:roc_larger_cinic10_dataset_1_model}
\end{figure*}
}

\subsection{Performance of Attacks using Fewer Reference Models} 
\label{app:fewer_reference_models}
In Table~\ref{tab:cifar_10_100_cinic_models_1_2_4}, we show the average result of attacks with their standard deviations obtained on three different datasets, i.e. CIFAR-10, CIFAR-100 and CINIC-10, under low computation budget where we use a few reference models.
    \begin{table*}[t!]
    \centering
    \pgfplotstableread[col sep=comma]{data/cifar10_aug_18_num_ref_comparison_4_true.csv}\tab
    \pgfplotstableread[col sep=comma]{data/cifar100_aug_18_num_ref_comparison_4_true.csv}\cifartab
    \pgfplotstableread[col sep=comma]{data/cinic10_num_ref_comparison_4_report_18_aug.csv}\cinictab
    \pgfplotstableread[col sep=comma]{data/cinic10_aug_18_comparison.csv}\cinicpoptab

        \resizebox{0.96\textwidth}{!}{
          \begin{tabular}{lllllllllllll}
            \toprule
            \multirow{3}{*}{\# Ref} &
              \multirow{3}{*}{Attack} &
              \multicolumn{3}{c}{CIFAR-10} &
               &
              \multicolumn{3}{c}{CIFAR-100} &
               &
              \multicolumn{3}{c}{CINIC-10} \\ \cmidrule{3-5} \cmidrule{7-9} \cmidrule{11-13}
             &
               &
              \multicolumn{1}{c}{\multirow{2}{*}{AUC}} &
              \multicolumn{2}{c}{\TPRatFPR} &
               &
              \multicolumn{1}{c}{\multirow{2}{*}{AUC}} &
              \multicolumn{2}{c}{\TPRatFPR} &
               &
              \multicolumn{1}{c}{\multirow{2}{*}{AUC}} &
              \multicolumn{2}{c}{\TPRatFPR} \\ \cmidrule{4-5} \cmidrule{8-9} \cmidrule{12-13}
             &
               &
              \multicolumn{1}{c}{} &
              \hundredth &
              \zero &
               &
              \multicolumn{1}{c}{} &
              \hundredth &
              \zero &
               &
              \multicolumn{1}{c}{} &
              \hundredth &
              \zero \\
\bottomrule
         \end{tabular}
    }
    \caption{Performance of attacks where a \textbf{few reference models} are used. All attacks, except LiRA (Online), are \textbf{low-cost and offline} (do not need new reference models to be trained per query). Results are averaged over 10 random target models.}
    \label{tab:cifar_10_100_cinic_models_1_2_4}
\end{table*}

\subsection{Performance of Attacks on Models Trained with other ML Algorithms}
\label{app:gradient_boosting_decision_tree}

While the majority of contemporary machine learning (ML) models rely on neural networks, understanding how attacks generalize in the presence of other machine learning algorithms is intriguing. Although it is beyond the scope of this paper to comprehensively analyze attacks across a wide range of ML algorithms on various datasets, we conduct a simple experiment to study their impact by training models with a Gradient Boosting Decision Tree (GBDT) algorithm. Figure~\ref{fig:gradient_boosted_decision_trees} illustrates the ROC of attacks when GBDT (with three different max depths) is employed to train models on our non-image dataset, i.e. Purchase-100. The hyper-parameters of GBDT are set as n\_estimators=250, lr=0.1 and subsample=0.2, yielding a test accuracy of around 53\%. For this experiment, we use two reference models (1 IN, 1 OUT). Since the output of GBDT is the prediction probability (not logit), we use this probability as the input signal for all attacks. The TPR obtained by our attack consistently outperforms all other attacks, particularly by an order of magnitude at zero FPR.

\setvalue{\tmplabel}{gradient_boosted_decision_trees-}
\setcounter{figureNumber}{0}
{\tikzset{external/figure name/.add={}{\tmplabel}}
\begin{figure*}[h]
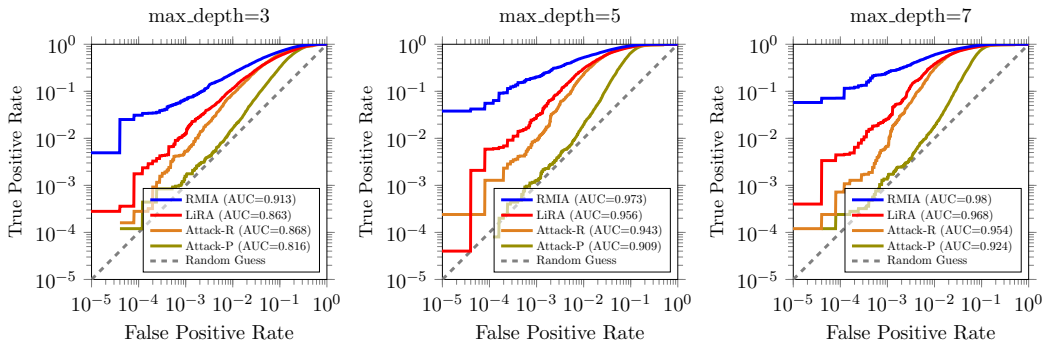

    \centering
    \if\compileFigures1
    \begin{tikzpicture}
    \pgfplotstableread[col sep=comma]{data/multi/purchase100_gbdt_roc.csv}\roc
    \pgfplotstableread[col sep=comma]{data/multi/purchase100_gbdt_auc.csv}\auc
    \begin{axis} [
        scale=0.31,
        width=1.0\linewidth,height=1.0\linewidth,
        ylabel={True Positive Rate},
        ymin=10e-6,ymax=1,xmin=10e-6,xmax=1, 
        xmode=log,ymode=log,
        xtick align = outside,
        ytick align = outside,
        ylabel near ticks,ylabel style={align=center, text width=4cm, font=\small},
        legend pos=south east,
        legend columns=1,
        legend style={nodes={scale=0.6, transform shape}, fill opacity=0.6, draw opacity=1,text opacity=1},
        legend cell align={left},
        every axis plot/.append style={\ROCthickness},
        reverse legend=true,
        ]
        \addplot [\colorRandom,dashed] table[x=random,y=random, col sep=comma] \roc;
        \addlegendentry {\random}

        \addplot [\colorPopulation,no marks] table[x=random,y={population}, col sep=comma] \roc;
        \addlegendentry {\population (\AUC=\pgfplotstablegetelem{0}{{population}}\of\auc\pgfplotsretval)}

        \addplot [\colorReference,no marks] table[x=random,y={reference 1}, col sep=comma] \roc;
        \addlegendentry {\reference (\AUC=\pgfplotstablegetelem{0}{{reference 1}}\of\auc\pgfplotsretval)}

        \addplot [\colorOnline,no marks,\styleAugOthers] table[x=random,y={lira 2}, col sep=comma] \roc;
        \addlegendentry {\lira (\AUC=\pgfplotstablegetelem{0}{lira 2}\of\auc\pgfplotsretval)}

        \addplot [\colorRelative,no marks,\styleAugOurs] table[x=random,y={relative 2}, col sep=comma] \roc;
        \addlegendentry {\relative (\AUC=\pgfplotstablegetelem{0}{{relative 2}}\of\auc\pgfplotsretval)}

    \end{axis}
\end{tikzpicture}
    \else
    \scalebox{\ROCMultiscaleFactor}{\includegraphics{fig/figure-\tmplabel\thefigureNumber.pdf}\stepcounter{figureNumber}
    \includegraphics{fig/figure-\tmplabel\thefigureNumber.pdf}\stepcounter{figureNumber}
    \includegraphics{fig/figure-\tmplabel\thefigureNumber.pdf}\stepcounter{figureNumber}}
    \fi
    \caption{ROC of attacks on models trained with Gradient Boosted Decision Tree (GBDT) on the Purchase-100 dataset. Three different values of max depth parameter are used in GBDT. The test accuracy of models is around 53\%. We use 2 reference models (1 IN, 1 OUT). In LiRA \cite{Carlini2022Membership}, we use the output probability $p$ to compute the rescaled-logit signal as $\log(\frac{p}{1-p})$. In other attacks, we use the output probability as the input signal of attacks.}
    \label{fig:gradient_boosted_decision_trees}
\end{figure*}
}

\subsection{Performance of Attacks Obtained on Different Datasets}
\label{app:roc_datasets}
We here compare the ROC of attacks using reference models trained on four different datasets, i.e. CIFAR-10, CIFAR-100, CINIC-10 and Purchase-100. We evaluate attacks when different number of reference models is used. We provide a depiction of the ROC curves in both log and normal scales. In Figure~\ref{fig:roc_four_datasets_1_model}, we show the ROC of attacks obtained with using 1 reference model. Since we only have 1 (OUT) model, the result of offline attacks is reported here. RMIA works remarkably better than other three attacks across all datasets. Although LiRA has a rather comparable TPR at low FPR for CIFAR-10 and Purchase-100 models, but it yields a much lower AUC (e.g. 22\% lower AUC in CIFAR-10), when compared with RMIA. In other two datasets, RMIA results in around 3 times better TPR at zero FPR, than its closest rival.

Figure~\ref{fig:roc_four_datasets_2_models} displays the ROC of attacks resulted from employing 2 reference model (1 IN, 1 OUT). Again, RMIA works much better than other three attacks (in terms of both AUC and TPR@FPR) across all datasets. For example, it has a 10\% higher AUC in CIFAR-10 models and at least 3 times better TPR at zero FPR in CIFAR-100 and CINIC-10, than other attacks.
Figure~\ref{fig:roc_four_datasets_127_models} presents the ROC of offline attacks when using 127 OUT models.  RMIA works much better than other three attacks across all datasets. For example, it leads to at least 20\% higher AUC than LiRA.
Figure~\ref{fig:roc_four_datasets_254_models} illustrates the ROC of attacks obtained when using all 254 models (127 IN, 127 OUT models). With the help of training hundreds of IN and OUT models, LiRA can work close to our attack in terms of AUC, but the TPR at zero FPR of RMIA is considerably higher, e.g. by at least 50\% in CIFAR-10, CINIC-10 and Purchase-100 models, as compared with LiRA.

For a better comparison between attack performances concerning the number of reference models, Figure~\ref{fig:aucs_line_graph} presents the AUC results for both offline and online attacks across varying reference model counts (ranging from 1 to 254). The left plots in this figure showcase the outcomes of offline attacks, while the right plots highlight the performance of online attacks. A consistent trend emerges, revealing that an increase in the number of reference models yields an improvement in AUC across all attacks. Notably, in both offline and online scenarios and across all datasets, RMIA consistently outperforms other attacks, particularly when employing a limited number of models. 

\subsection{Performance of Attacks on Purchase-100 Models}
\label{app:purchase-100}

Table~\ref{tab:purchase_models_2_4_new} presents the attack results when evaluating models trained on the Purchase-100 dataset. The observed superiority aligns with findings from other datasets: employing more reference models enhances the performance of all attacks. However, across all scenarios, RMIA consistently outperforms other attacks in terms of both AUC and TPR at low FPR metrics. Notably, RMIA with just 1 (OUT) reference model achieves over 30\% higher AUC than LiRA. The offline RMIA with 2 or 4 reference models can surpass online LiRA (using the same number of reference models). RMIA can maintain its superiority even in the presence of hundreds of reference models.
\begin{table}[h]
  \centering
  \pgfplotstableread[col sep=comma]{data/purchase100_num_ref_comparison.csv}\tab
  \pgfplotstableread[col sep=comma]{data/purchase_no_aug_comparison.csv}\purchasetab
      \resizebox{0.5\textwidth}{!}{
        \begin{tabular}{lllll}
          \toprule
          \multirow{2}{*}{\# Ref} &
            \multirow{2}{*}{Attack} &
            \multicolumn{1}{c}{\multirow{2}{*}{AUC}} &
            \multicolumn{2}{c}{\TPRatFPR} \\ \cmidrule{4-5}
           &
             &
            \multicolumn{1}{c}{} &
            \hundredth &
            \zero \\ \midrule
            \multirow{1}{*}{0} &
              \population &
              \pgfplotstablegetelem{0}{{population}}\of\tab\pgfplotsretval$\,\pm\,$\pgfplotstablegetelem{1}{{population}}\of\tab\pgfplotsretval &
              \pgfplotstablegetelem{4}{{population}}\of\tab\pgfplotsretval$\,\pm\,$\pgfplotstablegetelem{5}{{population}}\of\tab\pgfplotsretval &
              \pgfplotstablegetelem{6}{{population}}\of\tab\pgfplotsretval$\,\pm\,$\pgfplotstablegetelem{7}{{population}}\of\tab\pgfplotsretval \\
              \midrule
          \multirow{3}{*}{1} &
            \reference &
            \pgfplotstablegetelem{0}{{reference 1}}\of\tab\pgfplotsretval$\,\pm\,$\pgfplotstablegetelem{1}{{reference 1}}\of\tab\pgfplotsretval &
            \pgfplotstablegetelem{4}{{reference 1}}\of\tab\pgfplotsretval$\,\pm\,$\pgfplotstablegetelem{5}{{reference 1}}\of\tab\pgfplotsretval &
            \pgfplotstablegetelem{6}{{reference 1}}\of\tab\pgfplotsretval$\,\pm\,$\pgfplotstablegetelem{7}{{reference 1}}\of\tab\pgfplotsretval \\
           &
            \lira &
            \pgfplotstablegetelem{0}{{liraoffline 1}}\of\tab\pgfplotsretval$\,\pm\,$\pgfplotstablegetelem{1}{{liraoffline 1}}\of\tab\pgfplotsretval &
            \pgfplotstablegetelem{4}{{liraoffline 1}}\of\tab\pgfplotsretval$\,\pm\,$\pgfplotstablegetelem{5}{{liraoffline 1}}\of\tab\pgfplotsretval &
            \pgfplotstablegetelem{6}{{liraoffline 1}}\of\tab\pgfplotsretval$\,\pm\,$\pgfplotstablegetelem{7}{{liraoffline 1}}\of\tab\pgfplotsretval \\
\bottomrule
          \end{tabular}
    }
  \caption{Performance of attacks obtained for attacking models trained on \textbf{Purchase-100} using varying number of reference models. All attacks are \textbf{low-cost and offline} by default (do not need new reference models to be trained per query). However, we also provide the result of online versions of LiRA and RMIA to showcase their highest performance. We do not use augmented queries. Results are averaged over 10 random target models.}
  \label{tab:purchase_models_2_4_new}
\end{table}

\subsection{Performance of RMIA with Different Number of $z$ Samples}
\label{app:z_samples}
Table~\ref{tab:z_fraction_1_ref} presents the performance of RMIA obtained with 1 (OUT) reference model, as we change the number of reference samples. We here observe roughly the same trend of results as the one we saw in Table~\ref{tab:z_fraction} where we employed 254 reference models. Specifically, the AUC increases when we use more reference samples, but using 2500 population samples, equivalent to 10\% of the size of the models’ training
set, brings results comparable to those obtained with a 10
fold larger population set. 
\begin{table}[h]
\centering
\pgfplotstableread[col sep=comma]{data/relative_ratio_comparison_table_updated_1_ref.csv}\relativetab
    \resizebox{0.5\textwidth}{!}{
    \begin{tabular}{llllll}
    \toprule
    \multirow{2}{*}{\# population samples $z$} & \multirow{2}{*}{AUC}   & \multicolumn{2}{c}{\TPRatFPR}  \\ \cmidrule{3-4}
                                &                        & \hundredth & \zero             \\ \midrule
    25 samples                                                             & \pgfplotstablegetelem{0}{{z 25}}\of\relativetab\pgfplotsretval$\,\pm\,$\pgfplotstablegetelem{1}{{z 25}}\of\relativetab\pgfplotsretval                                      & \pgfplotstablegetelem{4}{{z 25}}\of\relativetab\pgfplotsretval$\,\pm\,$\pgfplotstablegetelem{5}{{z 25}}\of\relativetab\pgfplotsretval                                        & \pgfplotstablegetelem{6}{{z 25}}\of\relativetab\pgfplotsretval$\,\pm\,$\pgfplotstablegetelem{7}{{z 25}}\of\relativetab\pgfplotsretval                                      \\
    250 samples                                                             & \pgfplotstablegetelem{0}{{z 250}}\of\relativetab\pgfplotsretval$\,\pm\,$\pgfplotstablegetelem{1}{{z 250}}\of\relativetab\pgfplotsretval                                      & \pgfplotstablegetelem{4}{{z 250}}\of\relativetab\pgfplotsretval$\,\pm\,$\pgfplotstablegetelem{5}{{z 250}}\of\relativetab\pgfplotsretval                                        & \pgfplotstablegetelem{6}{{z 250}}\of\relativetab\pgfplotsretval$\,\pm\,$\pgfplotstablegetelem{7}{{z 250}}\of\relativetab\pgfplotsretval                                      \\
\bottomrule
    \end{tabular}
  }
\caption{Performance of \relative\offline using different number of random $z$ samples. We only use \textbf{1 (OUT) reference model}, trained on CIFAR-10. Here, we do not use augmented queries. Results are averaged over 10 random target models.}
\label{tab:z_fraction_1_ref}
\end{table}

\subsection{MIA Score Comparison between Attacks}
\label{app:mia_test_score_comparison}

To better understand the difference between the performance of our attack with others', we compare the MIA score of member and non-member samples obtained in RMIA and other attacks. Figure~\ref{fig:score_comparison_lira} displays RMIA scores versus LiRA scores in two scenarios: one using only 1 reference model (shown on the left) and another using 4 reference models (shown on the right). With just 1 reference model, RMIA provides clearer differentiation between numerous member and non-member samples, as it assigns distinct MIA scores in the ${[0, 1]}$ range, separating many members on the right side and non-members on the left side. In contrast, LiRA scores are more concentrated towards the upper side of the plot, lacking a distinct separation between member and non-member scores. When employing more reference models, we observe a higher degree of correlation between the scores of the two attacks.

\begin{figure*}[b!]
  \begin{subfigure}{0.45\textwidth}
    \includegraphics[width=\linewidth]{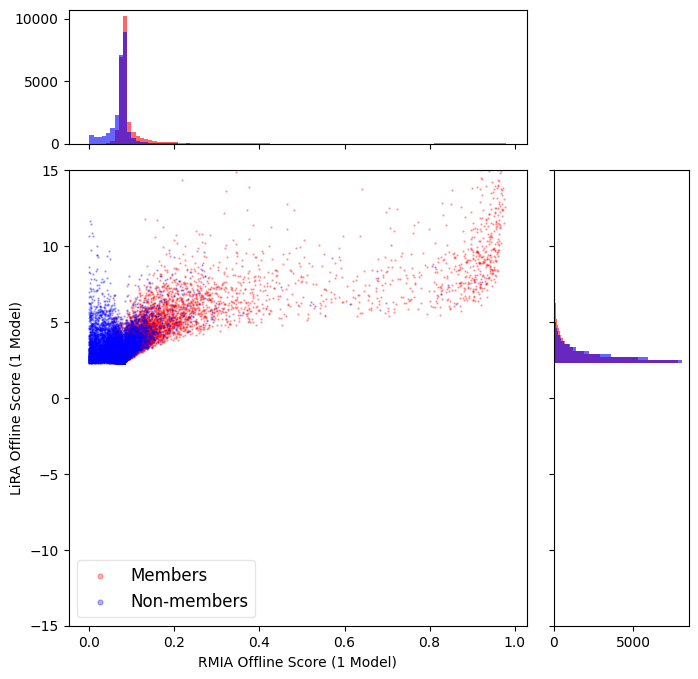}
    \caption{\lira against \relative with 1 model}
    \label{fig:score_lira_1}
  \end{subfigure}%
  \hspace*{\fill}   
  \begin{subfigure}{0.45\textwidth}
    \includegraphics[width=\linewidth]{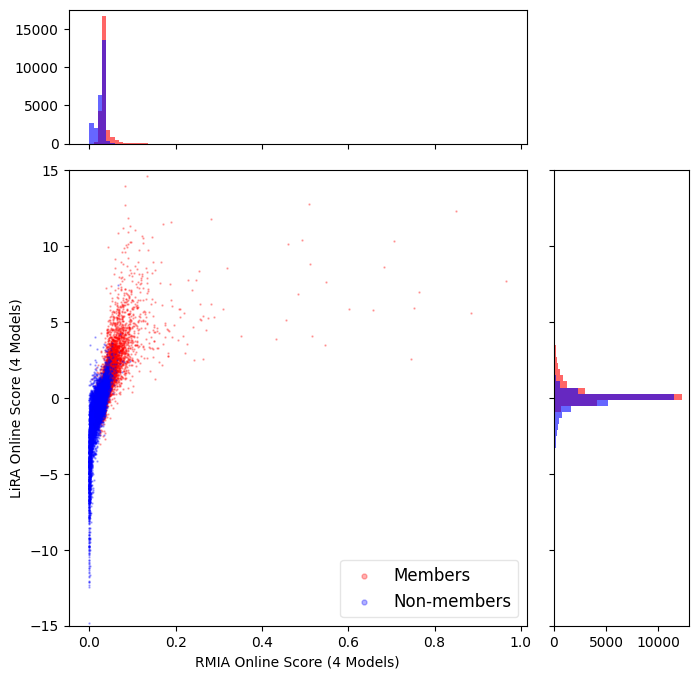}
    \caption{Against \lira against \relative with 4 models}
    \label{fig:score_lira_4}
  \end{subfigure}
    \caption{MIA score comparison between RMIA and \lirans~\citep{Carlini2022Membership} over all member and non-member samples of a random target model. The left plot is obtained when only 1 (OUT) model is used, while the right plot uses 4 reference models (2 IN and 2 OUT). In both plots, the x-axis and y-axis represent \relative and \lira scores, respectively.}    \label{fig:score_comparison_lira}
\end{figure*}

Similarly, Figure~\ref{fig:score_comparison_reference} shows RMIA scores compared to Attack-R scores in the same two scenarios. RMIA can better separate members from non-members via assigning distinct MIA scores to them (member scores apparently tend to be larger than non-member scores for lots of samples). In contrast, there is no such clear distinction in Attack-R. 

We also show RMIA scores versus Attack-P scores in Figure~\ref{fig:score_comparison_population}. In this experiment, we only use 1 reference model (OUT) for RMIA, because Attack-P does not work with reference models. As opposed to RMIA, Attack-P clearly fails to provide a good separation between member and non-member scores.

\begin{figure*}[h!]
  \begin{subfigure}{0.45\textwidth}
    \includegraphics[width=\linewidth]{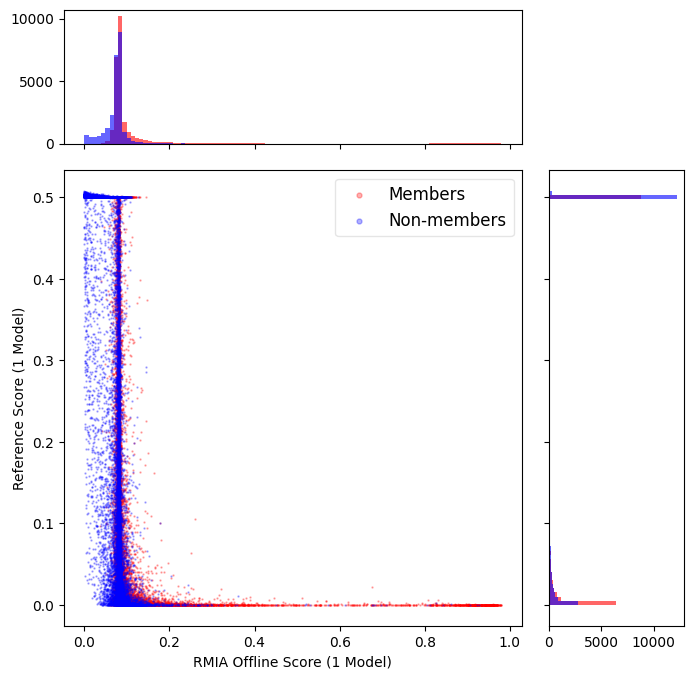}
    \caption{\reference against \relative with 1 model}
    \label{fig:score_reference_1}
  \end{subfigure}%
  \hspace*{\fill}
   \begin{subfigure}{0.45\textwidth}
    \includegraphics[width=\linewidth]{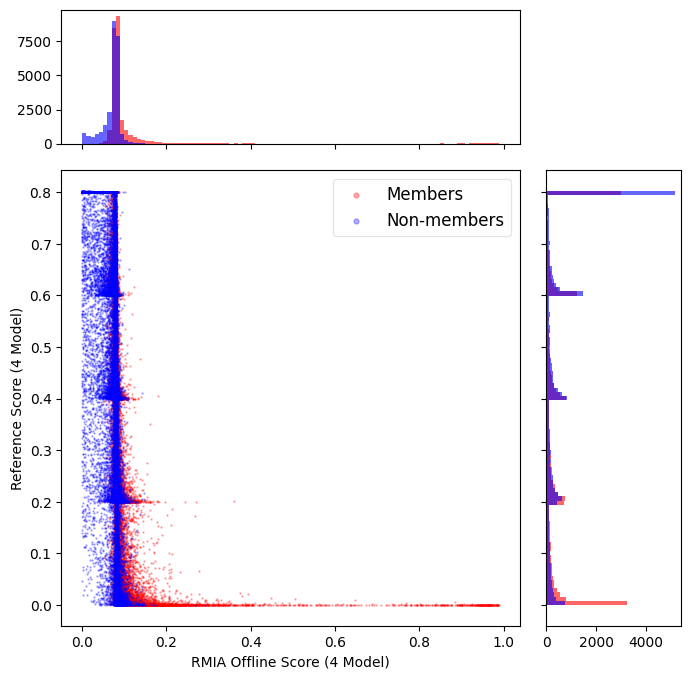}
    \caption{\reference against \relative with 4 models}
    \label{fig:score_reference_4}
  \end{subfigure}%
    \caption{MIA score comparison between RMIA and \referencens~\citep{Ye2022Enhanced}. The left plot is obtained when only 1 reference model (OUT) is used, while the right plot uses 4 models (OUT). In both plots, the x-axis and y-axis represent \relative and \reference scores, respectively.}
    \label{fig:score_comparison_reference}
\end{figure*}

\begin{figure*}[h!]
    \centering
      \includegraphics[width=0.45\linewidth]{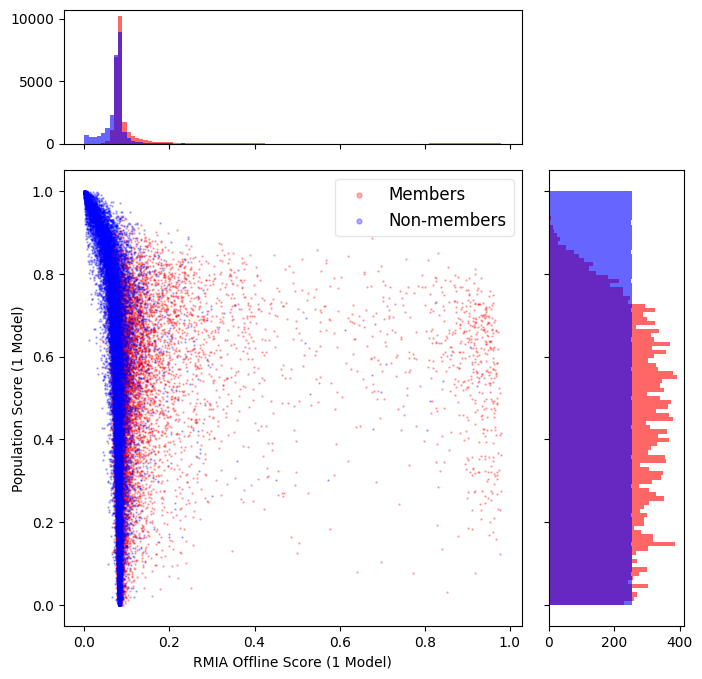}
      \caption{MIA score comparison between RMIA and \populationns~\citep{Ye2022Enhanced} obtained for a random target model. The x-axis represents \relative scores, while the y-axis depicts \population scores. For \relativens, we use only 1 OUT model. }
      \label{fig:score_comparison_population}
  \end{figure*}

Strong inference attacks compute $\ScoreMIA$ in such a way that it accurately reflects the distinguishability between models that are trained on a target data point and the ones that are not. For a better understanding of the distinctions between attacks, Figure~\ref{fig:mia_scores_all_samples} illustrates the distribution of MIA scores obtained from various attacks for a set of test samples. We compute $\ScoreMIA$ on 254 target models, with the sample being a member to half of them and a non-member to the other half. The attacks are conducted using 127 OUT models. The MIA score produced by \relative contributes to a more apparent separation between members and non-members (member scores tend to concentrate in the right part of the plot, as opposed to other attacks). 

\setvalue{\tmplabel}{roc_32_combined-}
\setcounter{figureNumber}{0}
{\tikzset{external/figure name/.add={}{\tmplabel}}
\begin{figure*}[th!]
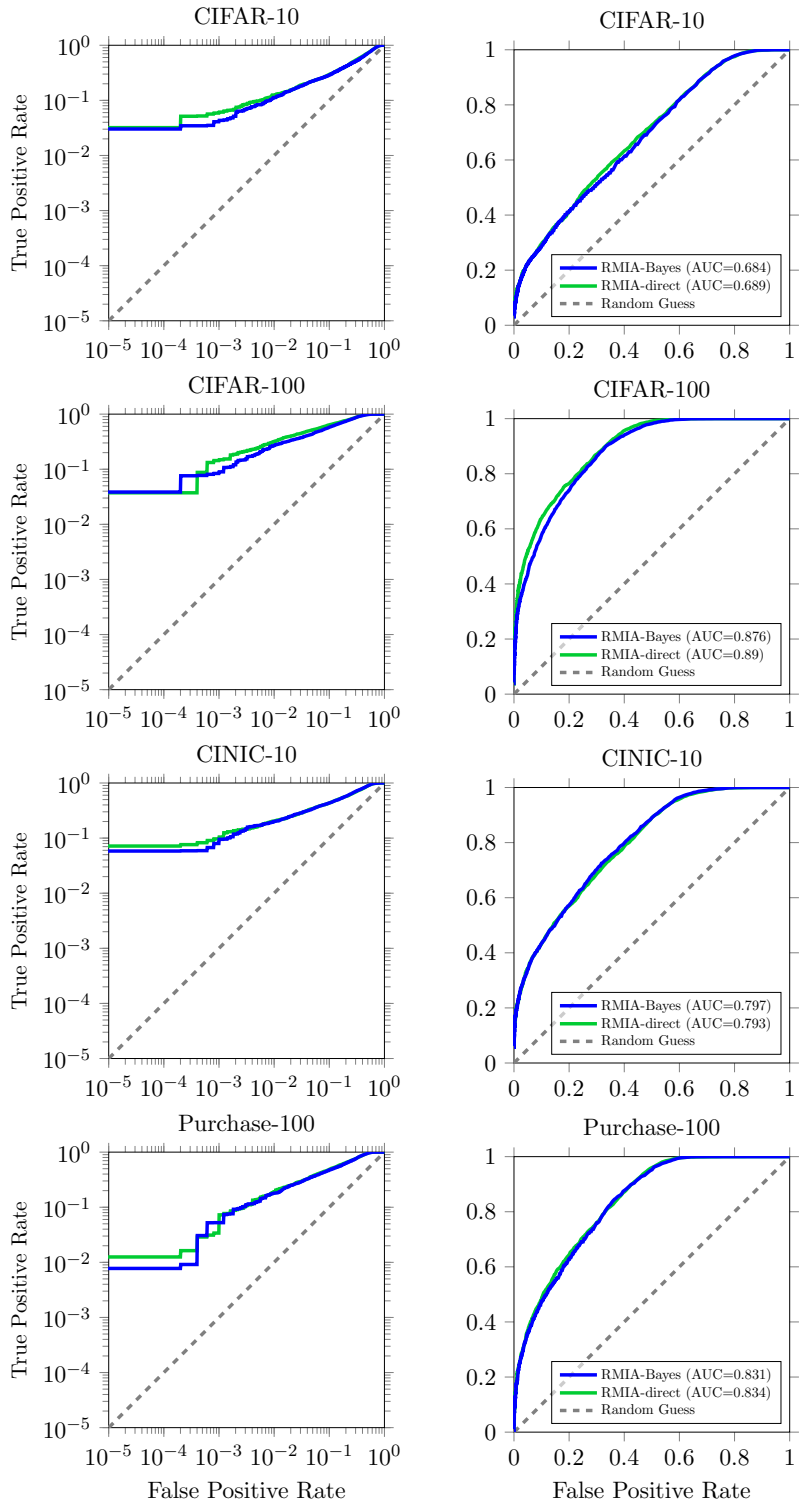

    \centering
    \if\compileFigures1
    \input{figure_scripts/all_rocs_combined_attack_32_script.tex}
    \else
    \scalebox{\ROCscaleFactor}{
    \includegraphics{fig/figure-\tmplabel\thefigureNumber.pdf}\stepcounter{figureNumber}\hspace*{2em}
    \includegraphics{fig/figure-\tmplabel\thefigureNumber.pdf}\stepcounter{figureNumber}}
    \scalebox{\ROCscaleFactor}{
    \includegraphics{fig/figure-\tmplabel\thefigureNumber.pdf}\stepcounter{figureNumber}\hspace*{2em}
    \includegraphics{fig/figure-\tmplabel\thefigureNumber.pdf}\stepcounter{figureNumber}}
    \scalebox{\ROCscaleFactor}{
    \includegraphics{fig/figure-\tmplabel\thefigureNumber.pdf}\stepcounter{figureNumber}\hspace*{2em}
    \includegraphics{fig/figure-\tmplabel\thefigureNumber.pdf}\stepcounter{figureNumber}}
    \scalebox{\ROCscaleFactor}{
    \includegraphics{fig/figure-\tmplabel\thefigureNumber.pdf}\stepcounter{figureNumber}\hspace*{2em}
    \includegraphics{fig/figure-\tmplabel\thefigureNumber.pdf}\stepcounter{figureNumber}}
    \fi
    \caption{ROC of RMIA when two different methods, i.e. RMIA-direct (as formulated in \eqref{eq:LR_expansion_direct_guassian}) and RMIA-Bayes (based on \eqref{eq:LR_computation}), are used to approximate the likelihood ratio. ROCs are shown in both log and normal scales. Here, we use \textbf{64 reference models}.  }
    \label{fig:roc_32_combined}
\end{figure*}
}

\setvalue{\tmplabel}{roc_2_combined-}
\setcounter{figureNumber}{0}
{\tikzset{external/figure name/.add={}{\tmplabel}}
\begin{figure*}[th!]
    \centering
    \if\compileFigures1
    \input{figure_scripts/all_rocs_combined_attack_script.tex}
    \else
    \scalebox{\ROCscaleFactor}{
    \includegraphics{fig/figure-\tmplabel\thefigureNumber.pdf}\stepcounter{figureNumber}\hspace*{2em}
    \includegraphics{fig/figure-\tmplabel\thefigureNumber.pdf}\stepcounter{figureNumber}}
    \scalebox{\ROCscaleFactor}{
    \includegraphics{fig/figure-\tmplabel\thefigureNumber.pdf}\stepcounter{figureNumber}\hspace*{2em}
    \includegraphics{fig/figure-\tmplabel\thefigureNumber.pdf}\stepcounter{figureNumber}}
    \scalebox{\ROCscaleFactor}{
    \includegraphics{fig/figure-\tmplabel\thefigureNumber.pdf}\stepcounter{figureNumber}\hspace*{2em}
    \includegraphics{fig/figure-\tmplabel\thefigureNumber.pdf}\stepcounter{figureNumber}}
    \scalebox{\ROCscaleFactor}{
    \includegraphics{fig/figure-\tmplabel\thefigureNumber.pdf}\stepcounter{figureNumber}\hspace*{2em}
    \includegraphics{fig/figure-\tmplabel\thefigureNumber.pdf}\stepcounter{figureNumber}}
    \fi
    \caption{ROC of RMIA when two different methods, i.e. RMIA-direct (as formulated in \eqref{eq:LR_expansion_direct_guassian}) and RMIA-Bayes (based on \eqref{eq:LR_computation}), are used to approximate the likelihood ratio. ROCs are shown in both log and normal scales. Here, we use \textbf{4 reference models}.}
    \label{fig:roc_2_combined}
\end{figure*}
}

\setvalue{\tmplabel}{roc_four_datasets_1_model-}
\setcounter{figureNumber}{0}
{\tikzset{external/figure name/.add={}{\tmplabel}}
\begin{figure*}[h]
    \centering
    \if\compileFigures1
    \input{figure_scripts/all_rocs_1_script.tex}
    \else
    \scalebox{\ROCscaleFactor}{
    \includegraphics{fig/figure-\tmplabel\thefigureNumber.pdf}\stepcounter{figureNumber}\hspace*{2em}
    \includegraphics{fig/figure-\tmplabel\thefigureNumber.pdf}\stepcounter{figureNumber}}
    \scalebox{\ROCscaleFactor}{
    \includegraphics{fig/figure-\tmplabel\thefigureNumber.pdf}\stepcounter{figureNumber}\hspace*{2em}
    \includegraphics{fig/figure-\tmplabel\thefigureNumber.pdf}\stepcounter{figureNumber}}
    \scalebox{\ROCscaleFactor}{
    \includegraphics{fig/figure-\tmplabel\thefigureNumber.pdf}\stepcounter{figureNumber}\hspace*{2em}
    \includegraphics{fig/figure-\tmplabel\thefigureNumber.pdf}\stepcounter{figureNumber}}
    \scalebox{\ROCscaleFactor}{
    \includegraphics{fig/figure-\tmplabel\thefigureNumber.pdf}\stepcounter{figureNumber}\hspace*{2em}
    \includegraphics{fig/figure-\tmplabel\thefigureNumber.pdf}\stepcounter{figureNumber}}
    \fi
    \caption{ROC of attacks using models trained on different datasets (ROCs are shown in both log and normal scales). The result is obtained on one random target model. We here use \textbf{1 reference model} (OUT). }
    \label{fig:roc_four_datasets_1_model}
\end{figure*}
}

\setvalue{\tmplabel}{roc_four_datasets_2_models-}
\setcounter{figureNumber}{0}
{\tikzset{external/figure name/.add={}{\tmplabel}}
\begin{figure*}[h]
    \centering
    \if\compileFigures1
    \input{figure_scripts/all_rocs_2_script.tex}
    \else
    \scalebox{\ROCscaleFactor}{
    \includegraphics{fig/figure-\tmplabel\thefigureNumber.pdf}\stepcounter{figureNumber}\hspace*{2em}
    \includegraphics{fig/figure-\tmplabel\thefigureNumber.pdf}\stepcounter{figureNumber}}
    \scalebox{\ROCscaleFactor}{
    \includegraphics{fig/figure-\tmplabel\thefigureNumber.pdf}\stepcounter{figureNumber}\hspace*{2em}
    \includegraphics{fig/figure-\tmplabel\thefigureNumber.pdf}\stepcounter{figureNumber}}
    \scalebox{\ROCscaleFactor}{
    \includegraphics{fig/figure-\tmplabel\thefigureNumber.pdf}\stepcounter{figureNumber}\hspace*{2em}
    \includegraphics{fig/figure-\tmplabel\thefigureNumber.pdf}\stepcounter{figureNumber}}
    \scalebox{\ROCscaleFactor}{
    \includegraphics{fig/figure-\tmplabel\thefigureNumber.pdf}\stepcounter{figureNumber}\hspace*{2em}
    \includegraphics{fig/figure-\tmplabel\thefigureNumber.pdf}\stepcounter{figureNumber}}
    \fi
    \caption{ROC of attacks using models trained on different datasets (ROCs are shown in both log and normal scales). The result is obtained on one random target model. We here use \textbf{2 reference models} (1 IN, 1 OUT). }
    \label{fig:roc_four_datasets_2_models}
\end{figure*}
}

\setvalue{\tmplabel}{roc_four_datasets_127_models-}
\setcounter{figureNumber}{0}
{\tikzset{external/figure name/.add={}{\tmplabel}}
\begin{figure*}[h]
    \centering
    \if\compileFigures1
    \input{figure_scripts/all_rocs_full_script_offline.tex}
    \else
    \scalebox{\ROCscaleFactor}{
    \includegraphics{fig/figure-\tmplabel\thefigureNumber.pdf}\stepcounter{figureNumber}\hspace*{2em}
    \includegraphics{fig/figure-\tmplabel\thefigureNumber.pdf}\stepcounter{figureNumber}}
    \scalebox{\ROCscaleFactor}{
    \includegraphics{fig/figure-\tmplabel\thefigureNumber.pdf}\stepcounter{figureNumber}\hspace*{2em}
    \includegraphics{fig/figure-\tmplabel\thefigureNumber.pdf}\stepcounter{figureNumber}}
    \scalebox{\ROCscaleFactor}{
    \includegraphics{fig/figure-\tmplabel\thefigureNumber.pdf}\stepcounter{figureNumber}\hspace*{2em}
    \includegraphics{fig/figure-\tmplabel\thefigureNumber.pdf}\stepcounter{figureNumber}}
    \scalebox{\ROCscaleFactor}{
    \includegraphics{fig/figure-\tmplabel\thefigureNumber.pdf}\stepcounter{figureNumber}\hspace*{2em}
    \includegraphics{fig/figure-\tmplabel\thefigureNumber.pdf}\stepcounter{figureNumber}}
    \fi
    \caption{ROC of offline attacks using models trained on different datasets (ROCs are shown in both log and normal scales). The result is obtained on one random target model. We use \textbf{127 reference models} (OUT). }
    \label{fig:roc_four_datasets_127_models}
\end{figure*}
}

\setvalue{\tmplabel}{roc_four_datasets_254_models-}
\setcounter{figureNumber}{0}
{\tikzset{external/figure name/.add={}{\tmplabel}}
\begin{figure*}[h]
    \centering
    \if\compileFigures1
    \input{figure_scripts/all_rocs_full_script.tex}
    \else
    \scalebox{\ROCscaleFactor}{
    \includegraphics{fig/figure-\tmplabel\thefigureNumber.pdf}\stepcounter{figureNumber}\hspace*{2em}
    \includegraphics{fig/figure-\tmplabel\thefigureNumber.pdf}\stepcounter{figureNumber}}
    \scalebox{\ROCscaleFactor}{
    \includegraphics{fig/figure-\tmplabel\thefigureNumber.pdf}\stepcounter{figureNumber}\hspace*{2em}
    \includegraphics{fig/figure-\tmplabel\thefigureNumber.pdf}\stepcounter{figureNumber}}
    \scalebox{\ROCscaleFactor}{
    \includegraphics{fig/figure-\tmplabel\thefigureNumber.pdf}\stepcounter{figureNumber}\hspace*{2em}
    \includegraphics{fig/figure-\tmplabel\thefigureNumber.pdf}\stepcounter{figureNumber}}
    \scalebox{\ROCscaleFactor}{
    \includegraphics{fig/figure-\tmplabel\thefigureNumber.pdf}\stepcounter{figureNumber}\hspace*{2em}
    \includegraphics{fig/figure-\tmplabel\thefigureNumber.pdf}\stepcounter{figureNumber}}
    \fi
    \caption{ROC of attacks using models trained on different datasets (ROCs are shown in both log and normal scales). The result is obtained on one random target model. We here use \textbf{254 reference models} (127 IN, 127 OUT). }
    \label{fig:roc_four_datasets_254_models}
\end{figure*}
}

\setvalue{\tmplabel}{aucs_line_graph-}
\setcounter{figureNumber}{0}
{\tikzset{external/figure name/.add={}{\tmplabel}}
\begin{figure*}[h]
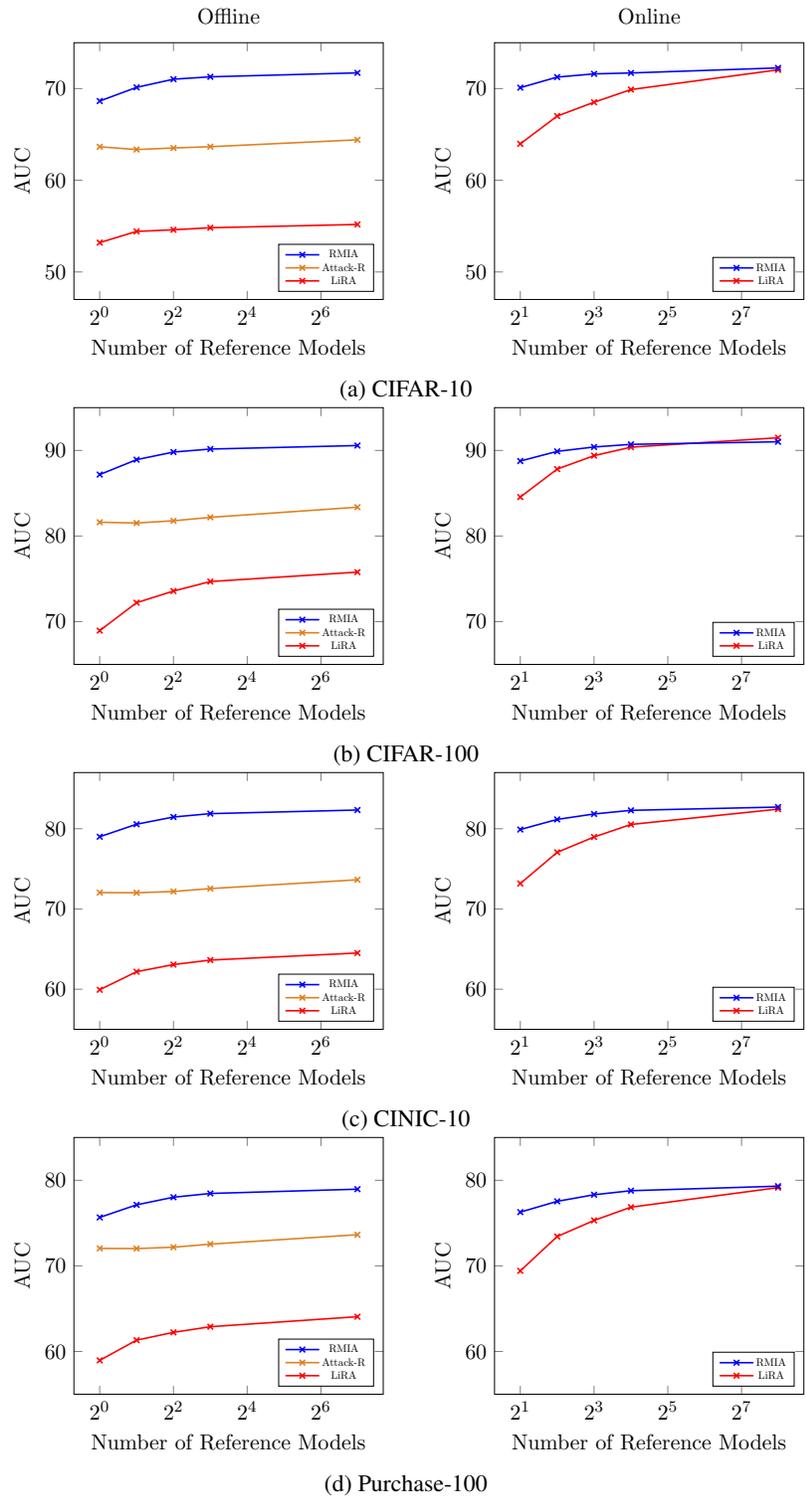

    \centering
    \if\compileFigures1
    \input{figure_scripts/aucs_line_graph.tex}
    \else
    \begin{subfigure}{\linewidth}
        \centering
        \scalebox{\AUCscaleFactor}{
        \includegraphics{fig/figure-\tmplabel\thefigureNumber.pdf}\stepcounter{figureNumber}\hspace*{2em}
        \includegraphics{fig/figure-\tmplabel\thefigureNumber.pdf}\stepcounter{figureNumber}
        }
        \caption{CIFAR-10}
    \end{subfigure}

    \begin{subfigure}{\linewidth}
        \centering
        \scalebox{\AUCscaleFactor}{
        \includegraphics{fig/figure-\tmplabel\thefigureNumber.pdf}\stepcounter{figureNumber}\hspace*{2em}
        \includegraphics{fig/figure-\tmplabel\thefigureNumber.pdf}\stepcounter{figureNumber}
        }
        \caption{CIFAR-100}
    \end{subfigure}

    \begin{subfigure}{\linewidth}
        \centering
        \scalebox{\AUCscaleFactor}{
        \includegraphics{fig/figure-\tmplabel\thefigureNumber.pdf}\stepcounter{figureNumber}\hspace*{2em}
        \includegraphics{fig/figure-\tmplabel\thefigureNumber.pdf}\stepcounter{figureNumber}
        }
        \caption{CINIC-10}
    \end{subfigure}

    \begin{subfigure}{\linewidth}
        \centering
        \scalebox{\AUCscaleFactor}{
        \includegraphics{fig/figure-\tmplabel\thefigureNumber.pdf}\stepcounter{figureNumber}\hspace*{2em}
        \includegraphics{fig/figure-\tmplabel\thefigureNumber.pdf}\stepcounter{figureNumber}
    }
        \caption{Purchase-100}
    \end{subfigure}

    \fi
    \caption{AUC of various attacks obtained with using different number of reference models. The left plots illustrate the results of offline attacks, while the right ones depict the AUC scores obtained by online attacks. For online attacks, half of reference models are OUT and half are IN. }
    \label{fig:aucs_line_graph}
\end{figure*}
}

\begin{figure*}[t!]
    \centering
    \scalebox{0.32}{    \includegraphics{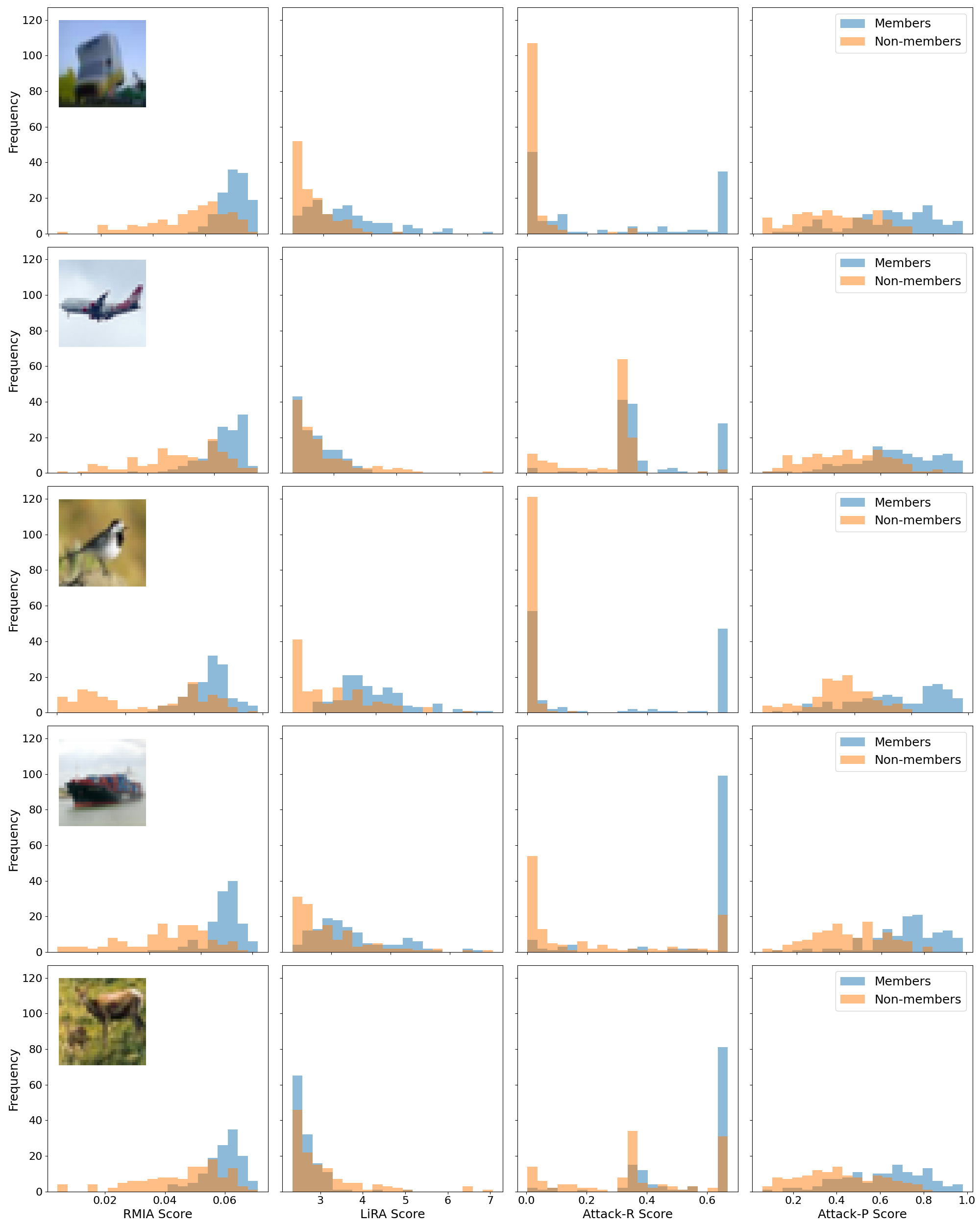}}
    \caption{Distribution of MIA scores obtained from different attacks for some test sample. We calculate the MIA score of each sample across 254 target models where half of them are IN models, and the remaining half are OUT models. To perform attacks, we use 127 reference models (OUT), trained on CIFAR-10.}
    \label{fig:mia_scores_all_samples}
\end{figure*}

\clearpage

\clearpage
\section{Related Work}

Neural networks, particularly when trained with privacy-sensitive datasets, have been proven to be susceptible to leaking information about their training data. A variety of attacks have been designed to gauge the degree of leakage and the subsequent privacy risk associated with training data. For instance, \emph{data extraction attacks} attempt to recreate individual samples used in training the model \citep{Carlini2021Extracting}, whereas \emph{model inversion attacks} focus on extracting aggregate information about specific sub-classes instead of individual samples \citep{Fredrikson2015Model}. In contrast, \emph{property inference attacks} derive non-trivial properties of samples in a target model's training dataset \citep{Ganju2018Property}. This paper, however, is concerned with \emph{membership inference attacks (MIAs)}, which predict whether a particular sample was used in the training process of the model. MIAs, due to their simplicity, are commonly utilized as auditing tools to quantify data leakage in trained models.

MIAs first found their use in the realm of genome data to identify the presence of an individual's genome in a mixed batch of genomes \citep{Homer2008Resolving}. \citep{Backes2016Membership} went on to formalize the risk analysis for identifying an individual's genome from aggregate statistics on independent attributes, with this analysis later extended to include data with dependent attributes \citep{Murakonda2021Quantifying}.

Algorithms featuring differential privacy (DP) are designed to limit the success rate of privacy attacks when distinguishing between two neighboring datasets \citep{Nasr2021Adversary}. Some researches, such as \cite{Thudi2022Bounding}, provide upper limits on the average success of MIAs on general targets. Other studies evaluate the effectiveness of MIAs on machine learning models trained with DP algorithms \citep{Rahman2018Membership}.

\cite{Shokri2017Membership} introduced membership inference attacks for machine learning algorithms. The work demonstrated the efficacy of membership inference attacks against machine learning models in a setting where the adversary has query access to the target model. This approach was based on the training of reference models, also known as shadow models, with a dataset drawn from the same distribution as the training data. Subsequent works extended the idea of shadow models to different scenarios, including white-box \citep{Leino2020Stolen, Nasr2019Comprehensive, Sablayrolles2019White} and black-box settings \citep{Song2021Systematic, Hisamoto2020Membership, Chen2021When}, label-only access \citep{ChoquetteChoo2021Label, Li2021Membership}, and diverse datasets \citep{Salem2019ML}. However, such methods often require the training of a substantial number of models upon receiving an input query, making them unfeasible due to processing and storage costs, high response times, and the sheer amount of data required to train such a number of models. MIA has been also applied in other machine learning scenarios, such as federated learning \citep{Nasr2019Comprehensive, Melis2019Exploiting, Truex2019Demystifying} and multi-exit networks \citep{Li2022Auditing}. 

Various mechanisms have been proposed to defend against MIAs, although many defense strategies have proven less effective than initially reported \citep{Song2021Systematic}. Since the over-fitting issue is an important factor affecting membership leakage, several regularization techniques have been used to defend against membership inference attacks, such as L2 regularization, dropout and label smoothing \citep{Shokri2017Membership, Salem2019ML, Liu2022Doctor}. Some recent works try to mitigate membership inference attacks by reducing the target model’s generalization gap \citep{Li2021Defense, Chen2022Relax} or self-distilling the training dataset \citep{Tang2022Mitigating}. \cite{Abadi2016Differential} proposed DP-SGD method which adds differential privacy \citep{Dwork2006Differential} to the stochastic gradient descent algorithm. Subsequently, some works concentrated on reducing the privacy cost of DP-SGD through adaptive clipping or adaptive learning rate \citep{Yu2019Differentially, Xu2020Adaptive}. In addition, there are defense mechanisms, such as AdvReg \citep{Nasr2018Adversarial} and MemGuard \citep{Jia2019MemGuard}, that have been designed to hide the distinctions between the output posteriors of members and non-members. 

Recent research has emphasized evaluating attacks by calculating their true positive rate (TPR) at a significantly low false positive rate (FPR) \citep{Carlini2022Membership, Ye2022Enhanced, Liu2022Membership, Long2020Pragmatic, Watson2022Difficulty}. For example, \cite{Carlini2022Membership} found that many previous attacks perform poorly under this evaluation paradigm. They then created an effective attack based on a likelihood ratio test between the distribution of models that use the target sample for training (IN models) and models that do not use it (OUT models). Despite the effectiveness of their attack, especially at low FPRs, it necessitates the training of many reference models to achieve high performance. \cite{Watson2022Difficulty} constructed a membership inference attack incorporating sample hardness and using each sample’s hardness threshold to calibrate the loss from the target model. \cite{Ye2022Enhanced} proposed a template for defining various MIA games and a comprehensive hypothesis testing framework to devise potent attacks that utilize reference models to significantly improve the TPR for any given FPR. \cite{Liu2022Membership} presented an attack which utilizes trajectory-based signals generated during the training of distilled models to effectively enhance the differentiation between members and non-members. The recent paper \citep{Wen2023Canary} has improved the performance of likelihood test-driven attacks by estimating a variant of the target sample through minimally perturbing the original sample, which minimizes the fitting loss of IN and OUT shadow models. Lately, 
\cite{Leemann2023Gaussian} have introduced a novel privacy notion called $f$-MIP, which allows for bounding the trade-off between the power and error of attacks using a function $f$. This is especially applicable when models are trained using gradient updates. The authors demonstrated the use of DP-SGD to achieve this $f$-MIP bound. Additionally, they proposed a cost-effective attack for auditing the privacy leakage of ML models, with the assumption that the adversary has white-box access to gradients of the target model.

A number of related work tune the threshold~$\beta$ in MIA based on the data point. \citet{chang2021privacy} improve the attack by determining the threshold for a data point, based on its attributes (e.g., data points corresponding to the same gender are used as the reference population to determine the threshold corresponding to a given FPR.) When it is not very easy to identify the reference population, \citet{Bertran2023Scalable} train an attack model for each FPR threshold to determine the threshold associated with the reference population.

Some recent studies have focused on low-cost auditing of differentially private deep learning algorithms. This is a different setting in terms of the problem statement, yet it is aligned with the spirit of our work that aims at reducing the auditing cost. \citet{steinke2023privacy} outline an efficient method for auditing differentially private machine learning algorithms. Their approach achieves this with a single training run, leveraging parallelism in the addition or removal of multiple training examples independently. 

In alignment with MIA research~\citep{Sankararaman2009Genomic, Murakonda2021Quantifying, Ye2022Enhanced, Carlini2022Membership}, we highlight the power (TPR) of MIAs at extremely low errors (FPR). We propose a different test that allows us to devise a new attack that reaps the benefits of various potent attacks. The attack proposed in this paper demonstrates superior overall performance and a higher TPR at zero FPR than the attacks introduced by \citep{Carlini2022Membership, Ye2022Enhanced}, especially when only a limited number of reference models are trained. Additionally, it maintains high performance in an offline setting where none of the reference models are trained with the target sample. This characteristic renders our attack suitable for practical scenarios where resources, time, and data are limited.

\end{document}